\documentclass[10pt]{article}

\usepackage{amsfonts}
\usepackage{amsmath}
\usepackage{amsthm}
\usepackage{amssymb}
\usepackage{bm}
\usepackage{color}
\usepackage{graphicx}
\usepackage{url}
\usepackage{subcaption}
\usepackage{algorithm}
\usepackage{algorithmic}
\usepackage{authblk}
\usepackage{natbib}

\usepackage{mcr}

\title{Valid and Exact Statistical Inference for Multi-dimensional Multiple Change-Points by Selective Inference}

\date{\today}

\makeatletter
\def\@fnsymbol#1{\ensuremath{\ifcase#1\or
{*}\or 
{}\or 
{}\or 
\else\@ctrerr\fi}}
\makeatother

\author[1]{Ryota Sugiyama}
\author[1]{Hiroki Toda }
\author[1,2]{Vo Nguyen Le Duy}
\author[1]{Yu Inatsu}
\author[1,2]{Ichiro Takeuchi \thanks{Corresponding to: takeuchi.ichiro@nitech.ac.jp}}
\affil[1]{Nagoya Institute of Technology}
\affil[2]{RIKEN}

\begin{document}

\maketitle

\thispagestyle{empty}

\begin{abstract}
 \noindent
In this paper, we study statistical inference of change-points (CPs) in multi-dimensional sequence.
In CP detection from a multi-dimensional sequence, it is often desirable not only to detect the location, but also to identify the subset of the components in which the change occurs.
Several algorithms have been proposed for such problems, but no valid exact inference method has been established to evaluate the statistical reliability of the detected locations and components.
In this study, we propose a method that can guarantee the statistical reliability of both the location and the components of the detected changes.
We demonstrate the effectiveness of the proposed method by applying it to the problems of genomic abnormality identification and human behavior analysis.
\end{abstract}

\section{Introduction}
\label{sec:introduction}
In this paper, we consider the multi-dimensional multiple change-point (CP) detection problem~\citep{truong2018review,sharma2016trend,enikeeva2019high,enikeeva2019high,cho2016change,horvath2012change,jirak2015uniform,cho2015multiple,zhang2010detecting,vert2010fast,levy2009detection,bleakley2011group}.
We denote a multi-dimensional sequence dataset by a $D \times N$ matrix $X \in \RR^{D \times N}$, where each column is called \emph{location} and each row is called \emph{component}.
Each location is represented by a $D$-dimensional vector, whereas each component is represented by one-dimensional sequence with the length $N$. 
We consider the case where changes exist in multiple locations and multiple components.
Multi-dimensional multiple CP detection is an important problem in various fields such as computational biology~\citep{takeuchi2009potential} and signal processing~\citep{hammerla2016deep}.

We focus on the problem of detecting both location and component when there are changes in multiple components at each of the multiple locations. 
For example, in the field of computational biology, the identification of genome copy-number abnormalities is formulated as a CP detection problem, where it is important to find changes that are commonly observed in multiple patients.
This problem can be interpreted as the problem of detecting both the location (genome position of the genome abnormality) and the component (the patients with the genomic abnormality) of multiple CPs.

\begin{figure}[t]
 \centering
 \includegraphics[width=.9\linewidth]{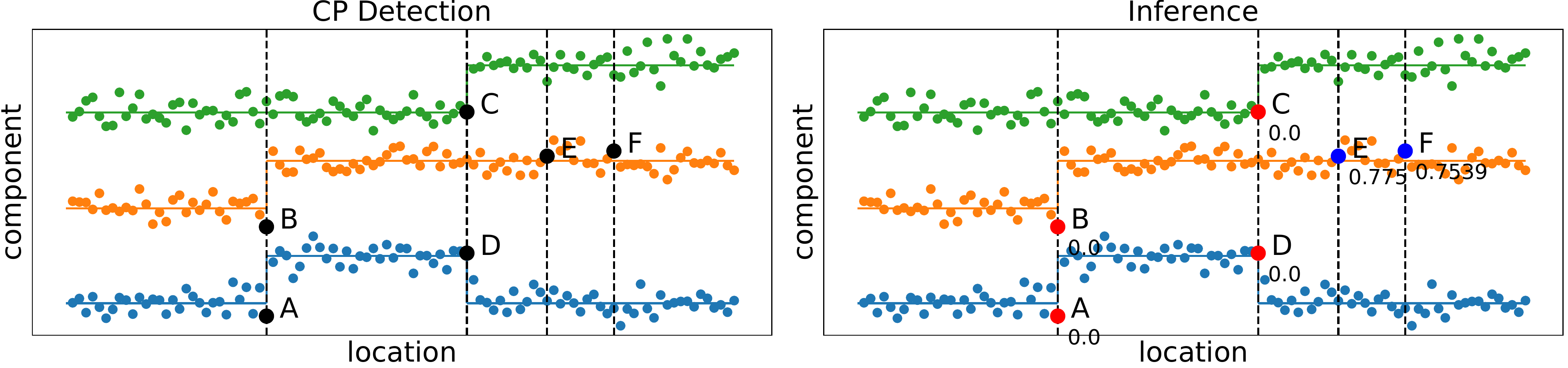}
 \caption{
 An example of the problem considered in this paper.
 The left and the right plots indicate the results of CP detection and CP inference, respectively.  
 In this example, six CPs (A-F) are detected by a CP detection method but the CP inference method that we propose in this paper suggests that the changes are statistically significant ($p~{\rm value} < 0.05$) at A, B, C, and D but not at E and F ($p~{\rm value} \ge 0.05$). 
 }
 \label{fig:illustration}
\end{figure}

Our main contribution in this paper is NOT to propose an algorithm for multi-dimensional multiple CP detection, but to develop a method for evaluating the reliability of the detected CPs. 
Figure~\ref{fig:illustration} shows an example of the problem we consider in this paper.
In this paper, we develop a novel method to quantify the statistical reliability of multiple CP locations and components in the form of $p$-values when they are detected by a certain CP detection algorithm.
The importance of evaluating the reliability of AI and machine learning (ML) results has been actively discussed in the context of \emph{reliable AI/ML} --- this study can be interpreted as a contribution toward reliable AI/ML.

Reliability evaluation of the detected CPs is inherently difficult.
This is because, in most CP detection problems, only a single dataset is given, and the CP detection and evaluation must be done on the same dataset.
Therefore, locations and components of the CPs are \emph{selected} to fit the observed dataset, which suggests that the \emph{selection bias} must be properly considered. 
Unlike supervised learning such as regression and classification, it is not possible to use data splitting or cross-validation.

In traditional statistics, reliability evaluation of CPs has been based on asymptotic theory, i.e., the case where $N \to \infty$~\citep{enikeeva2019high,cho2016change,horvath2012change,zhang2010detecting,jirak2015uniform,harchaoui2009regularized,korostelev2008multi}.
The asymptotic theory of CPs is built on various regularity assumptions, such as sufficiently weak sequential correlations, and it has been pointed out that unless these regularity assumptions are met, the evaluation will be inaccurate.
Furthermore, the asymptotic theory of CPs is developed in relatively simple problem settings, such as a single CP or one-dimensional sequence, and it cannot be easily extended to a multi-dimensional multiple CP detection problem. 

In this paper, we propose a method for valid and exact (non-asymptotic) statistical inference on CPs by utilizing the framework of conditional selective inference (SI), which has recently received much attention in the context of statistical inference on feature selection~\citep{lee2016exact,taylor2015statistical,fithian2014optimal,tibshirani2016exact,loftus2014significance,loftus2015selective,suzumura2017selective,terada2019selective,yamada2018post_a,yamada2018post_b,le2021parametric,sugiyama2021more}.
In the past few years, there have been attempts to extend the conditional SI framework into various problems~\citep{lee2015evaluating,chen2019valid,shimodaira2019selective,tanizaki2019computing,duy2020quantifying,tsukurimichi2021conditional,duy2021exact}, and several researchers have started to utilize conditional SI framework to evaluate the reliability of CPs~\citep{umezu2017selective,hyun2018post,jewell2020estimation,duy2020computing}.
However, to the best of our knowledge, there is no such study for multi-dimensional multiple CPs. 
Furthermore, if we simply extends the methods of these existing studies to multi-dimensional multiple CP detection problems, it turns out that the power of the inference would be low. 
Therefore, we propose a new approach based on conditional SI, which is more powerful than the naive extensions of existing approaches.

\section{Problem Statement}
\paragraph{Dataset and statistical model}
Consider a dataset in the form of a multi-dimensional sequence with the length $N$ and the dimension $D$. 
We denote the dataset as
$X \in \RR^{D \times N}$ 
where
each row is called a
\emph{component}
and 
each column is called a 
\emph{location}. 
The 
$i^{\rm th}$
component,
denoted as
$X_{i,:} \in \RR^N$
for $i \in [D]$, 
is represented as one-dimensional sequence with the length
$N$.
The 
$j^{\rm th}$
location,
denoted as
$X_{:,j} \in \RR^D$
for $j \in [N]$,
is represented as a vector with the length 
$D$. 
Furthermore,
we denote the $ND$-dimensional vector obtained by
concatenating rows of $X$ as 
${\rm vec}(X) = [X_{1,:}^\top, \ldots, X_{D,:}^\top]^\top \in \RR^{ND}$ 
where
${\rm vec}(\cdot)$
is an operator to transform a matrix into a vector with concatenated rows. 

We interpret that the dataset $X$ is a random sample from a statistical model
\begin{align}
 \label{eq:statistical_model}
 {\rm vec}(X) \sim N({\rm vec}(M), \Xi \otimes \Sigma), 
\end{align}
where
$M \in \RR^{D \times N}$
is the matrix containing the mean values of each element of 
$X$,
$\Xi \in \RR^{D \times D}$
is the covariance matrix for representing correlations between components
and 
$\Sigma \in \RR^{N \times N}$
is the covariance matrix for representing correlations between locations. 
Here, we do not assume any specific structure for the mean matrix
$M$ 
and assume it to be unknown.
On the other hand, 
the covariance matrices 
$\Xi$ 
and 
$\Sigma$
are assumed to be estimable from an independent dataset
where
it is known that no change points exist.

\paragraph{Detection of locations and components of multiple CPs}
We consider
the problem of the identifying
both the locations and the components
of multiple CPs
in the dataset
$X$
when there exist changes in a subset of the
$D$
components
at each of the multiple CP locations. 
Let
$K$
be the number of detected CPs.
The locations of the $K$ detected CPs are denoted as 
$1 \le \tau_1, \ldots, \tau_K < N$
and 
the subsets of the components where changes present are denoted as
$\bm \theta^1, \ldots, \bm \theta^K$ 
where
$\bm \theta^k$,
$k \in [K]$,
is the set of the components
in which changes exist
at the location
$\tau_k$.
Elements of 
$\bm \theta^k$
is denoted as
$\theta^k_1, \ldots, \theta^k_{|\bm \theta^k|}$. 
For notational simplicity,
we also define
$\tau_0 = 0$
,
$\tau_{K+1} = N$
and
$\bm{\theta}^0 = \emptyset$, 
$\bm{\theta}^{K+1} = \emptyset$.
Given a multi-dimensional sequence dataset $X$, 
the task is to identify both the locations 
$\cT = (\tau_1, \ldots, \tau_K)$
and
the sets of the components 
$\Theta = (\bm \theta^1, \ldots, \bm \theta^K)$. 
of the detected multiple CPs (see Fig.~\ref{fig:illustration}).

As a CP detection method
for both locations and components, 
we employ a method combining
dynamic-programming (DP)-based approach~\citep{bellman1966dynamic,guthery1974partition,bai2003computation,lavielle2005using,rigaill2010pruned}
and
scan-statistic (SS)-based approach~\citep{enikeeva2019high}.
This CP detection method is formulated as the following optimization problem:
\begin{align}
 \label{eq:scan}
 (\hat{\cT}, \hat{\Theta})
 =
 \argmax_{
 \substack{
 \cT = (\tau_1, \ldots, \tau_K), \\
 \Theta = (\bm \theta^1, \ldots, \bm \theta^K)
 }
 }
 \sum_{k \in [K]}
 C_{\tau_k-L+1:\tau_k+L}^{(\bm \theta^k)}(\tau_k, W)
\end{align}
where
\begin{subequations}
\begin{align}
 \label{eq:scan_a}
 C_{s:e}^{(\bm \theta)}(j, W)
 &
 =
 \frac{
 \|
 {\rm diag}(\bm 1_{D, \bm \theta})
 Z_{s:e}^\prime(j, W)
 \|_2^2
 -
 |\bm \theta|
 }{
 \sqrt{
 2 |\bm \theta|
 }
 },
 \\
 \label{eq:scan_b}
 Z_{s:e}^\prime(j, W)
 &
 =
 \sum_{\Delta = -W}^W
 Z_{s, e}
 (j + \Delta),
 \\
 \label{eq:scan_c}
 Z_{s:e}(j)
 &
 =
 V_{s,j,e}
 \left(
 \frac{
 \sum_{j^\prime=s}^j
 X_{:,j^\prime}
 }{
 j - s + 1
 }
 -
 \frac{
 \sum_{j^\prime=j+1}^e
 X_{:,j^\prime}
 }{
 e-j
 }
 \right),
 \\
 \label{eq:scan_d}
 V_{s,j,e}
 &
 =
 \sqrt{
 \frac{
 (j-s+1)(e-j)
 }{
 e-s+1
 }
 }.
\end{align}
\end{subequations}
Here, 
$L$ 
and 
$W$
are hyperparameters
which are explained below, 
$\bm 1_{D, \bm \theta}$
is $D$-dimensional binary vector 
in which the $i^{\rm th}$ element is 
$1$ if $i \in \bm \theta$ 
and $0$ otherwise, 
and
${\rm diag}(\bm 1_{D, \bm \theta}) \in \mathbb{R}^{D\times D}$
is a matrix in which vector $\bm 1_{D, \bm \theta}$ 
is arranged diagonally.
The CP detection method
in \eq{eq:scan} 
is motivated as follows. 
%
%
We employ 
the
\emph{scan statistic}
proposed in 
\citep{enikeeva2019high}
as the building block,
which was
introduced 
for
the problem of
detecting a single CP
from a multi-dimensional sequence\footnote{
Double CUMSUM~\citep{cho2016change} is known as another CP detection method that enables us to select not only the location but also the subset of the components of the change.
We focus on the scan-statistic in this study, but conjecture that our SI framework can be applied also to double CUMSUM. 
}.
As a criterion
for a single CP
at the location 
$\tau \in [N]$
and
the components
$\bm \theta \subseteq [D]$,
the scan statistic is defined as 
\begin{align*}
\frac{
\|
{\rm diag}(\bm 1_{D, \bm \theta}) Z_{s:e}(\tau)
\|^2
-
 |\bm \theta|
}{
\sqrt{2 |\bm \theta|}
}. 
\end{align*}
We extend the scan statistic in various directions
in \eq{eq:scan}. 
First,
we extended the scan statistic
to handle multiple CPs,
i.e.,
multiple
locations 
$\tau_1, \ldots, \tau_K$
and the corresponding sets of the components
$\bm \theta^1, \cdots, \bm \theta^K$. 
Next,
in each component
$\theta^k_h \in \bm \theta^k, h \in [|\bm \theta^k|]$
we consider the mean signal difference 
between 
the 
$L$
points before and after the CP location 
$\tau_k$, 
i.e.,
the mean difference between
$X_{h, \tau_k - L+1: \tau_k}$
and 
$X_{h, \tau_k+1: \tau_k + L}$.
This is to avoid the influence of neighboring CPs 
when evaluating the statistical reliability of
each location 
$\tau_k$
and
the components
$\theta^k_h$. 
Here,
$L$
is considered as a hyperparameter 
%
Furthermore, 
the reason why we use
$Z_{(s,e)}^\prime(\tau)$
instead of
$Z_{(s,e)}(\tau)$
in
\eq{eq:scan_a}
is that
we consider changes to be common among components 
if it is within the range of
$\pm W$
rather than
defining common changes as the changes 
at exactly the same location.
Here, 
$W$
is also considered as a hyperparameter.

The optimal solution of 
\eq{eq:scan}
can be efficiently obtained
by using 
DP 
and the algorithm for efficient scan-statistic computation
described in
\citet{enikeeva2019high}.
The pseudo-code for solving the optimization problem 
\eq{eq:scan}
is presented in Algorithms~\ref{alg:cp_detection}
(SortComp and UpdateDP functions are presented in Appendix~\ref{app:A}).
%
%

\begin{algorithm}
 \caption{CP detection algorithm for \eq{eq:scan}}
 \label{alg:cp_detection}
 \begin{algorithmic}
  \REQUIRE dataset $X \in \mathbb{R}^{D\times N}$ and parameters $K, L, W$
  \STATE Set $F_{(0, j)} = 0, \mathcal{T}_{(0, j)} = (0), \Theta_{(0, j)} = \phi$ for $j \in [N - LK]$
  \STATE Set $\mathcal{S}_{:, m} = $SortComp(m) for $m \in [L, N-L]$
  \FOR{$k = 1$ to $K-1$}
  \FOR{$j = L(k+1)$ to $N - L(K - k)$}
  \STATE $F_{(k, j)}, \hat{\tau}_k^{(k, j)}, \hat{\bm{\theta}}_k^{(k, j)}$
  \STATE $\qquad = \text{UpdateDP}(j, k, F_{(k-1, Lk:j-L)}, \mathcal{S})$
  \STATE $\mathcal{T}_{(k, j)} = {\rm concat}(\mathcal{T}_{(k-1, \hat{\tau}_k^{(k, j)})}, \hat{\tau}_k^{(k, j)})$
  \STATE $\Theta_{(k, j)} = {\rm concat}(\Theta_{(k-1, \hat{\tau}_k^{(k, j)})}, \hat{\bm{\theta}}_k^{(n, k)})$
  \ENDFOR
  \ENDFOR
  \STATE $F_{(K, N)}, \hat{\tau}_K^{(K, N)}, \hat{\bm{\theta}}_K^{(K, N)}$
  \STATE $\qquad =  \text{UpdateDP}(N, K, F_{(K-1, LK:N-L)}, \mathcal{S})$
  \STATE $\mathcal{T}_{(K, N)} = (\mathcal{T}_{(K-1, \hat{\tau}_K^{(K, N)})}, \hat{\tau}_K^{(K, N)}, N)$
  \STATE $\Theta_{(K, N)} = {\rm concat}(\Theta_{(K-1, \hat{\tau}_K^{(K, N)})}, \hat{\bm{\theta}}_K^{(K, N)}, \phi)$
  \ENSURE $\{\tau_k\}_{k \in [K]} = \mathcal{T}_{(K, N)}, \{\bm{\theta}^{k}\}_{k \in [K]} = \Theta_{(K, N)}$
 \end{algorithmic}
\end{algorithm}



\paragraph{Statistical Inference on Detected Locations and Components}
The main goal of this paper is
to develop a valid and exact (non-asymptotic) statistical inference method
for
detected locations and components
of multiple CPs.
For each of the detected location
$\tau_k$
and
component
$\theta^k_h$,
$k \in [K], h \in [|\bm \theta^k|]$, 
we consider the statistical test
with the following
null and alternative hypotheses:
\begin{subequations} 
 \label{eq:test}
\begin{align}
 \label{eq:test_null}
 &
 {\rm H}_0^{(k, h)}:
 \frac{1}{L-W}
 \sum_{j=\tau_k-L+1}^{\tau_k-W}
 M_{\theta_h^k, j}
 =
 \frac{1}{L-W}
 \sum_{j=\tau_k+W+1}^{\tau_k+L}
 M_{\theta_h^k, j}
 \\
 \label{eq:test_alternative}
 &
 {\rm H}_1^{(k, h)}:
 \frac{1}{L-W}
 \sum_{j=\tau_k-L+1}^{\tau_k-W}
 M_{\theta_h^k, j}
 \neq
 \frac{1}{L-W}
 \sum_{j=\tau_k+W+1}^{\tau_k+L}
 M_{\theta_h^k, j}
\end{align}
\end{subequations}
This test simply concerns
whether the averages of
$L$
points before and after the CP location 
$\tau_k \pm W$
is equal or not
in the component
$\theta_h^k$.
Note that,
throughout the paper,
we do NOT assume that 
the mean of each component
$M_{:,i}, i \in [D]$
is piecewise-constant function,
i.e.,
the proposed statistical inference method in 
\S\ref{sec:sec3}
is valid
in the sense that
the type I error is properly controlled
at the specified significance level
$\alpha$ (e.g., $\alpha = 0.05$).
While
existing statistical inference methods for CPs
often require the assumption
on the true structure 
(such as piecewise constantness, piecewise linearity, etc.),
the conditional SI framework
in this paper
enables us to be free from 
assumptions about
the true structure on
$M_{i,j}, i \in [D], j \in [N]$
although the power of the method depend on the unknown structure. 

For the statistical test in
\eq{eq:test},
a reasonable test statistic is the observed mean signal difference
\begin{align*}
 \psi_{k, h}
 =
 \frac{1}{L-W}
 \left(
 \sum_{j = \tau_k - L + 1}^{\tau_k - W}
 X_{\theta^k_h, j}
 - 
 \sum_{j = \tau_k + W + 1}^{\tau_k + L}
 X_{\theta^k_h, j}
 \right).
\end{align*}
This test statistic can be written as
\begin{align*}
 \psi_{k, h} = \bm \eta_{k, h}^\top {\rm vec}(X), 
\end{align*}
by defining an $ND$-dimensional vector
\begin{align*}
 \bm \eta_{k, h}
 = 
 \frac{1}{L-W}
 \bm 1_{\theta^k_h} 
 \otimes
 \left(
 \bm 1_{\tau_k - L + 1 : \tau_k - W}
 - 
 \bm 1_{\tau_k + W + 1 : \tau_k + L}
 \right),
\end{align*}
where
$\bm 1_{\theta^k_h}$
is $D$-dimensional binary vector
in which the
$\theta_h^{k}$th
element is 1 
and
0 otherwise,
and
$\bm \one_{s:e}$
is $N$-dimensional binary vector
in which the
$j^{\rm th}$
element is 1 
if 
$j \in [s:e]$
and
0 otherwise.
In case 
both the CP location
$\tau_k$
and
the CP component 
$\theta^k_h$ 
are known before looking at data, 
since 
${\rm vec}(X)$
follows the multivariate normal distribution
in \eq{eq:statistical_model}, 
the null distribution of the test statistic 
$\bm \eta_{k, h}^\top {\rm vec}(X)$
also follows a multivariate normal distribution 
under the null hypothesis
\eq{eq:test_null}. 
Unfortunately,
however,
because 
the CP location
$\tau_k$ 
and
the CP component
$\theta^k_h$
are actually 
\emph{selected}
by looking at the observed dataset $X^{\text{obs}}$,
it is highly difficult to derive the sampling distribution of 
the test statistic.
In the literature of traditional statistical CP detection,
in order to mitigate this difficulty,
only simple problem settings, e.g., with a single location or/and a single component, 
have been studied
by relying on asymptotic approximation with 
$N \to \infty$
or/and 
$D \to \infty$.
However,
these asymptotic approaches are not available for complex problem settings
such as \eq{eq:scan},
and
are only applicable when certain regularity assumptions on sequential correlation are satisfied.

\paragraph{Conditional SI}
To conduct exact (non-asymptotic) statistical inference
on the locations and the components of the detected CPs, 
we employ
conditional SI.
The basic idea of conditional SI is to make inferences
based on the sampling distribution
conditional on the \emph{selected} hypothesis~\citep{lee2016exact,fithian2014optimal}.
Let us denote
the set of pairs of location and component
of the detected CPs 
when the dataset
$X$
is applied to Algorithm~\ref{alg:cp_detection} 
as 
\begin{align*}
 \cM(X) = \{(\tau_k, \theta_h^k)\}_{k \in [K], h \in [|\bm \theta^k|]}. 
\end{align*}
To apply conditional SI framework into our problem, 
we consider 
the following sampling distribution
of the test statistic
$\bm \eta_{k,h}^\top {\rm vec}(X)$
conditional on the selection event that
a change is detected at location
$\tau_k$
in the component
$\theta^k_h$,
i.e., 
\begin{align*}
 \bm \eta_{k, h}^\top {\rm vec}(X)
 \mid
 (\tau_k, \theta^k_h) \in \cM(X).
\end{align*}
Then,
for testing the statistical significance of the change,
we introduce so-called
\emph{selective $p$-value} 
$p_{k, h}$, 
which satisfies the following sampling property 
\begin{align}
 \label{eq:pval_property}
 \PP_{{\rm H_0}^{(k, h)}}
 \left(
 p_{k, h}
 \le
 \alpha
 \mid 
 (\tau_k, \theta^k_h) \in \cM(X)
 \right)
 =
 \alpha,
 ~
 \forall
 \alpha \in (0, 1), 
\end{align}
where
$\alpha \in (0, 1)$
is the significance level of the statistical test
(e.g., $\alpha = 0.05$)\footnote{
Note that 
a valid $p$-value in a statistical test should satisfies
$\PP(p \le \alpha) = \alpha ~ \forall \alpha \in (0, 1)$
under the null hypothesis
so that the probability of the type I error (false positive) 
coincides with the significance level 
$\alpha$. 
}.
To define the selective $p$-value,  
with a slight abuse of notation,
we make distinction between the random variables 
$X$
and the observed dataset
$X^{\rm obs}$. 
The selective $p$-value is defined as
\begin{align}
 \label{eq:selective_p}
 p_{k, h}
 =
 \PP_{{\rm H}_0^{(k, h)}}
 \left(
 |
 \bm \eta_{k, h}^\top
 {\rm vec}(X)
 |
 \ge
 |
 \bm \eta_{k, h}^\top
 {\rm vec}(X^{\rm obs})
 |
 ~\Bigg|~
 \begin{array}{l}
 (\tau_k, \theta_h^k)
 \in
 \cM(X),
  \\
 \bm q_{k, h}(X)
 =
 \bm q_{k, h}(X^{\rm obs})
 \end{array}
 \right),
\end{align}
where
$\bm q_{k, h}(X)$
is the nuisance parameter defined as
\begin{align*}
 \bm q_{k, h}(X)
 =
 I_{ND}
 -
 \bm c
 \bm \eta_{k, h}^\top
 {\rm vec}(X)
 ~
 \text{ with }
 ~
 \bm c
 = 
 \frac{
 \Xi \otimes \Sigma
 \bm \eta_{k, h}
 }{
 \bm \eta_{k, h}^\top 
 \Xi \otimes \Sigma
 \bm \eta_{k, h}
 }.
\end{align*}
Note that
the selective $p$-values
depend on the nuisance component
$\bm q_{k, h}(X)$,
but the sampling property in
\eq{eq:pval_property}
is kept satisfied
without conditioning on the nuisance event
$\bm q_{k, h}(X) = \bm q_{k, h}(X^{\rm obs})$
because
$\bm q_{k, h}(X)$
is independent of the test statistic
$\bm \eta_{k, h}^\top {\rm vec}(X)$. 
The selective $(1-\alpha)$ confidence interval (CI)
for any $\alpha \in (0, 1)$ 
which satisfies the $(1-\alpha)$ coverage property
conditional on the selection event 
$(\tau_k, \theta_h^k) \in \cM(X)$ 
can be also defined similarly. 
See
\citet{lee2016exact} and \citet{fithian2014optimal} 
for a more detailed discussion 
and
the proof
that
the selective $p$-value
\eq{eq:selective_p}
satisfies
the sampling property
\eq{eq:pval_property} 
and how to define selective CIs. 

The discussion so far indicates that, 
if we can compute the conditional probability in
\eq{eq:selective_p},
we can conduct a valid exact test for each of the detected change 
at the location 
$\tau_k$ 
in the component
$\theta^k_h$
for
$k \in [K], h \in [|\bm{\theta}^k|]$.
Most of the existing conditional SI studies
consider the case
where the hypothesis selection event is represented 
as
an intersection of a set of linear or quadratic inequalities.
For example,
in the seminal work of conditional SI in \citet{lee2016exact}, 
the selective $p$-value and CI computations are conducted
by exploiting the fact that 
the event that some features (and their signs) are selected by Lasso
is represented as an intersection of a set of linear inequalities,
i.e.,
the selection event is represented as a polyhedron of sample (data) space. 
Unfortunately, however, the selection event in \eq{eq:selective_p} is highly complicated and cannot be simply represented an intersection of linear nor quadratic inequalities. 
We resolve this computational challenge in the next section. 
Our basic idea is to consider a family of \emph{over-conditioning} problems such that the computation of the conditional probability of each of the over-conditioning problems can be tractable and then sum them up to obtain the selective $p$-value in \eq{eq:selective_p}. 
This idea is motivated by \citet{liu2018more} and \citet{le2021parametric} in which the authors proposed methods to increase the power of conditional SI for Lasso in \citet{lee2016exact}.

\section{Proposed Method}
\label{sec:sec3}
In this section,
we propose a method to compute
the selective $p$-value
in
\eq{eq:selective_p}.
Because 
we already know that the test statistic 
$\bm \eta_{k, h}^\top {\rm vec}(X)$
follows a normal distribution
with the mean 
$\bm \eta_{k, h}^\top {\rm vec}(M)$
and the variance 
$\bm \eta_{k, h}^\top \Xi \otimes \Sigma \bm \eta_{k, h}$
without conditioning, 
the remaining task is to identify the subspace defined as 
\begin{align}
 \label{eq:inverse_problem}
 \cX = 
 \left\{
 \text{vec}(X) \in \RR^{DN}
 ~ \Bigg| ~
 \begin{array}{l}
  (\tau_k, \theta^k_h)
  \in \cM(X),
  \\
  \bm q_{k, h}(X) = \bm q_{k, h}(X^{\rm obs})
 \end{array}
 \right\}.
\end{align}
Unfortunately,
there is no simple way to identify the subspace
$\cX$
because it is an inverse problem of the optimization problem in \eq{eq:scan}.
Therefore, we adopt the following strategy: 
\begin{enumerate}
 \item
      Reformulate the inverse problem \eq{eq:inverse_problem} as an inverse problem on a line in $\RR^{D \times N}$;
 \item
      Consider a family of \emph{over-conditioning} problems by adding additional conditions.
 \item
      Show that the inverse problem for each of the over-conditioning problems considered in step 2 can be solved analytically;
 \item 
      Solve a sequence of inverse problems for a sequence of over-conditioning problems along the line considered in step 1.
\end{enumerate}

\subsection{Conditional data space on a line}
\label{subsec:3.1}
It is easy to see that the condition 
$\bm q_{k, h}(X) = \bm q_{k, h}(X^{\rm obs})$ 
indicates that
$\{X \in \RR^{D \times N} | \bm q_{k, h}(X) = \bm q_{k, h}(X^{\rm obs})\}$
is represented as a line in
$\RR^{D \times N}$
as formally stated in the following lemma.
\begin{lemm}
 Let
 \begin{align*}
  \bm a
  &
  =
  \bm q_{k, h}(X^{\rm obs}),
  \\
  \bm b
  &
  =
  \frac{
  \Xi \otimes \Sigma \bm \eta_{k, h}
  }{
  \bm \eta_{k, h}^{\top} \Xi \otimes \Sigma \bm \eta_{k, h}
  }.
 \end{align*}
 Then, 
 \begin{align*}
  &
  \left\{
 {\rm vec}(X) \in \RR^{DN}
 ~ \Bigg| ~
 \begin{array}{l}
 (\tau_k, \theta^k_h)
 \in \cM(X),
 \\
 \bm q_{k, h}(X) = \bm q_{k, h}(X^{\rm obs})
 \end{array}
 \right\}
  \\
  =
&
  \left\{
 {\rm vec}(X)  = \bm a + \bm b z
 ~ \Bigg| ~
\begin{array}{l}
 (\tau_k, \theta^k_h)
  \in \cM(X),
  \\
 z \in \RR
\end{array}
 \right\}
 \end{align*}
\end{lemm}
\noindent
This lemma indicates that
we do not have to consider the inverse problem for 
$DN$-dimensional data space
but only to consider an inverse problem for 
one-dimensional projected data space
\begin{align*}
 \{z \in \RR | (\tau_k, \theta^k_h) \in \cM(\bm a + \bm b z)\}.
\end{align*}
This fact has been already implicitly exploited
in the seminal conditional SI work in
\citet{lee2016exact}, 
but explicitly discussed first in
\citet{liu2018more}.
Here, with a slight abuse of notation, $\cM(X)$ is equivalent to $\cM({\rm vec} (X))$.

\subsection{Over-conditioning (OC)}
\label{subsec:3.2}
Consider all possible sets of triplets of location and component 
which can be obtained by applying 
Algorithm~\ref{alg:cp_detection}
to 
any
$D \times N$
multivariate sequence dataset
$X$.
Since the number of such sets are finite,
we can represent each of such sets
by using an index
$u = 1, 2, \cdots$
as
$\{\cM_u\}_{u=1, 2, \cdots}$.
Let us consider
\begin{align*}
 \cU^{(k, h)}
 =
 \left\{
 u \mid (\tau_k, \theta^k_h) \in \cM_u
 \right\}. 
\end{align*}
Then,
the conditioning event
$\{(\tau_k, \theta^k_h) \in \cM(X)\}$
in
\eq{eq:selective_p}
is written as 
\begin{align*}
\{(\tau_k, \theta^k_h) \in \cM(X)\}
=
\bigcup_{u \in \cU^{(k, h)}}\{\cM_u = \cM(X)\}.
\end{align*}

Next, 
let us delve into
Algorithm~\ref{alg:cp_detection}
and interpret this algorithm has the following two steps:
\begin{enumerate}
 \item Given $X \in \RR^{D \times N}$, compute $\cA(X) = (\cT, \Theta, \cS)$, 
 \item Given $\cA(X)$, compute $\cM(X)$, 
\end{enumerate}
where the notation 
$\cA(X) = (\cT, \Theta, \cS)$
indicates that
$\cT$ , $\Theta$ and $\cS$ in 
Algorithm~\ref{alg:cp_detection}
depend on the dataset $X$. 
Consider all possible pairs of tables 
$\cA = (\cT, \Theta, \cS)$ 
which can be obtained by applying 
the above 1st step of Algorithm~\ref{alg:cp_detection} 
to 
any
$D \times N$
multivariate sequence dataset
$X$.
Since the number of such pairs is finite,
we can represent each of such pairs 
by using an index
$v = 1, 2, \cdots$
as
$\{\cA_v\}_{v=1, 2, \cdots}$.
Let us now write the above step 2 of 
Algorithm~\ref{alg:cp_detection}
as
$\cM = \phi(\cA)$ 
and consider 
\begin{align*}
 \cV_u^{(k, h)}
 =
 \{v \mid \cM_u = \phi(\cA_v) \},
 u \in \cU^{(k, h)}.
\end{align*}
Then,
the conditioning event
$\{(\tau_k, \theta^k_h) \in \cM(X)\}$
in
\eq{eq:selective_p}
is written as 
\begin{align*}
 \{(\tau_k, \theta^k_h) \in \cM(X)\}
 =
 \bigcup_{u \in \cU^{(k, h)}}
 \left\{
 \cM_u = \cM(X)
 \right\}
 =
 \bigcup_{u \in \cU^{(k, h)}}
 \bigcup_{v \in \cV_u^{(k, h)}}
 \left\{
 \cA_v = \cA(X)
 \right\}
\end{align*}

\subsection{Inverse problem for over-conditioning problem}
\label{subsec:3.3}
Based on
\S\ref{subsec:3.1}
and
\S\ref{subsec:3.2},
we consider
the subset of one-dimensional projected dataset on a line
for each of the over-conditioning problems:
\begin{align*}
 \cZ_{v}
 =
 \left\{
 z \in \RR
 ~|~
 \cA_v = \cA(\bm a + \bm b z)
 \right\}, 
 v = 1, 2, \cdots.
\end{align*}
\begin{lemm}
 \label{lemm:quadratic}
 Let us define the following two sets represented as intersections of quadratic inequalities of $z$:
 \begin{align*}
  \cE_{\text{sort}}^{(m)} 
  &= 
  \bigcap_{d=1}^{D-1}
  \{z
  ~|~ 
  (\bm{b}^{\top}A_{(m, d)}\bm{b})z^2 
  + 2(\bm{b}^{\top}A_{(m, d)}\bm{a})z 
  + \bm{a}^{\top}A_{(m, d)}\bm{a} 
  \leq 
  0
  \} \\
  \cE_{\text{table}}^{(k, j)} 
  &= 
  \bigcap_{m=Lk}^{j - L}\bigcap_{d=1}^D
  \{z 
  ~|~ 
  (\bm{b}^{\top}B_{(k, j, m, d)}\bm{b}) z^2 
  + 2(\bm{b}^{\top}B_{(k, j, m, d)}\bm{a})z 
  + \bm{a}^{\top}B_{(k, j, m, d)}\bm{a}  
  + c_{(k, j, m, d)}
  \leq 
  0
  \}
 \end{align*}
 for $A_{(m, d)}, B_{(k, j, m, d)} \in \mathbb{R}^{ND\times ND}$ and $c_{(k, j, m, d)} \in \mathbb{R}$.
 Then, the subset 
 $\mathcal{Z}_v$
 is characterized as
 \begin{align*}
 \mathcal{Z}_v = \bigcap_{m=L}^{L-N}\{z~|~z \in \cE_{\text{sort}}^{(m)}\} 
 \cap \bigcap_{k=1}^{K-1}\bigcap_{j=L(k+1)}^{N-L(K-k)}\{z~|~z\in \cE_{\text{table}}^{(k, j)}\} 
 \cap \{z~|~z \in \cE_{\text{table}}^{(K, N)}\}
 \end{align*}
\end{lemm}
\noindent
Here,
matrices
$A, B$
and
a scalar 
$c$
require very complex and lengthy definitions
(see Appendix~\ref{app:B} for their complete definitions).
Briefly speaking,
$A_{(m, d)}$ is a matrix for the sorting operation required to select $\tilde{\bm{\theta}}^{(m)}$, 
while 
$B_{(k, j, m, d)}$
and
$c_{(k, j, m, d)}$
are a matrix
and
a scalar for updating the DP tables which are required to select
$(\hat{\tau}_k^{(k, j)}, \hat{\bm{\theta}}_k^{(k, j)})$ in $(k, j)$.
The proof of
Lemma~\ref{lemm:quadratic}
is presented in Appendix~\ref{app:B}. 
Lemma~\ref{lemm:quadratic}
indicates that the subset
$\cZ_v$
for each
$v$
can be characterized by an intersection of quadratic inequalities with respect to $Z$
and
can be obtained analytically. 

\subsection{Line search}
\label{subsec:3.4}

The results in
\S\ref{subsec:3.1},
\S\ref{subsec:3.2}
and 
\S\ref{subsec:3.3}
are summarized as 
\begin{align*}
 \cX
 &
 =
 \left\{
 {\rm vec}(X) \in \RR^{DN}
 ~\Bigg|~
 \begin{array}{l}
  (\tau_k, \theta^k_h) \in \cM(X), 
  \\
  \bm q(X) = \bm q(X^{\rm obs})
 \end{array}
 \right\}
 \\
 &
 =
 \left\{
 {\rm vec}(X) = \bm a + \bm b z
 ~\Bigg|~
 z
 \in
 \bigcup_{u \in \cU^{(k, h)}}
 \bigcup_{v \in \cV_u^{(k, h)}}
 \cZ_v
 \right\}.
\end{align*}
Our strategy is to identify 
$
\cZ
=
\bigcup_{u \in \cU^{(k, h)}}
\bigcup_{v \in \cV_u^{(k, h)}}
\cZ_v
$
by repeatedly applying 
Algorithm~\ref{alg:cp_detection}
to a sequence of datasets 
${\rm vec}(X) = \bm a + \bm b z$
within sufficiently wide range of
$z\in [z_{\rm min}, z_{\rm max}]$\footnote{
In \S\ref{sec:sec4},
we set
$z_{\rm min} = -10^6$
and 
$z_{\rm max} = 10^6$ 
because the probability mass outside the range is negligibly small.
}.
Because
$\cZ_v \in \RR$
is characterized by an intersection of quadratic inequalities, 
it is represented either as
an empty set,
an interval
or
two intervals
in
$\RR$. 
For simplicity\footnote{
The case where $\cZ_v$ is empty set does not appear in the following discussion, and the extension to the case where $\cZ_v$ is represented as two intervals is straightforward.
}, 
we consider the case
where 
$\cZ_v$
is represented by an interval and denoted as 
$\cZ_v = [L_v, U_v]$. 

Our method works as follows.
Let
$z^{(1)} = z_{\rm min}$. 
By applying
Algorithm~\ref{alg:cp_detection}
to the dataset 
${\rm vec}(X) = \bm a + \bm b z^{(1)}$, 
we obtain 
$v^{(1)}$
such that 
$\cA_{v^{(1)}} = \cA(\bm a + \bm b z^{(1)})$. 
It means that 
$\cA_{v^{(1)}} = \cA(\bm a + \bm b z)$
for
$z \in [L_{v^{(1)}}, U_{v^{(1)}}]$. 
Therefore,
we next consider
$z^{(2)} = U_{v^{(1)}} + \delta$
where
$\delta$
is a very small value
(e.g., $\delta = 10^{-6}$).
Then,
we apply 
Algorithm~\ref{alg:cp_detection}
to the dataset 
${\rm vec}(X) = \bm a + \bm b z^{(2)}$,
and obtain 
$v^{(2)}$
such that 
$\cA_{v^{(2)}} = \cA(\bm a + \bm b z^{(2)})$. 
We simply repeat this process up to $z = z_{\rm max}$. 
Finally, 
we enumerate intervals
$[L_{v^{(t)}}, U_{v^{(t)}}]$
such that
$v^{(t)} \in \cV_u^{(k, h)}, u \in \cU^{(k, h)}$,
which enables us to obtain
$\cZ = \bigcup_{u \in \cU^{(k, h)}} \bigcup_{v \in \cV_u^{(k, h)}} \cZ_v$.

Once $\cZ$ is computed,
remembering that the test-statistic is normally distributed, 
the selective $p$-value
$p_{k, h}$
in
\eq{eq:selective_p}
is computed based on a truncated normal distribution with the truncation region
$\cZ$.
Specifically, 
\begin{align*}
 p_{k, h}
 &
 =
 2 \min
 \left\{
 \pi_{k, h}, 1 - \pi_{k, h}
 \right\},
 \\
 \pi_{k, h}
 &
 =
 1 - F_{0, \bm \eta_{k, h}^\top \Xi \otimes \Sigma \bm \eta_{k, h}}^{\cZ}(\bm \eta_{k, h}^\top {\rm vec}(X^{\rm obs}))
\end{align*}
where
$F_{\mu, \sigma^2}^{\cZ}$
is the cumulative distribution function of the truncated normal distribution
with the mean $\mu$, the variance $\sigma^2$ and the truncation region $\cZ$. 

\subsection{Discussion}
\label{subsec:3.5}
In this section, we discuss the power of conditional SI. 
It is important to note that a conditional SI method is still valid (i.e., the type I error is properly controlled at the significance level $\alpha$) in the case of over-conditioning~\citep{fithian2014optimal}.
In fact, in the early papers on conditional SI (including the seminal paper by \citet{lee2016exact}), the computations of selective $p$-values and CIs were made tractable by considering over-conditioning cases.
In the past few years, for improving the power of conditional SI, a few methods were proposed to resolve the over-conditioning issue~\citep{liu2018more,le2021parametric}.
In \citet{liu2018more}, the authors pointed out that the power can be improved by conditioning only on individual features, rather than on the entire subset of features selected by Lasso.
In \citet{le2021parametric}, the authors proposed how to avoid over-conditioning on Lasso conditional SI by interpreting the problem as a search problem on a line in the data space.
In this study, we introduce these two ideas into the task of quantifying the statistical reliability of the locations and components of multiple CPs.
Specifically, by only considering the locations of $L$ points before and after each CP, the statistical reliability of each location can be evaluated independently of other locations, which results in improved power.
Furthermore, the approach of searching for matching conditions on a straight line in the data space is inspired by the method proposed in \citet{le2021parametric}.
In the experiments in \S\ref{sec:sec4}, the SI conditional on the over-condition $\cA(X) = \cA(X^{\rm obs})$ is denoted as ``proposed-OC'', and it is compared with the minimum (i.e., optimal) condition case denoted as ``proposed-MC''.

Optimization approaches to solving families of optimization problems parameterized by scalar parameter are called \emph{parametric programming} or \emph{homotopy methods}.
Parametric programming and homotopy methods are used in statistics and machine learning in the context of \emph{regularization path}.
A regularization path refers to the path of the (optimal) models when the regularization parameter is continuously varied.
Regularization paths for various problems and various models have been studied in the literature~\citep{osborne2000new,HasRosTibZhu04,RosZhu07,BacHecHor06,RosZhu07,Tsuda07,Lee07,Takeuchi09a,takeuchi2011target,Karasuyama11,hocking11a,Karasuyama12a,ogawa2013infinitesimal,takeuchi2013parametric}.
The algorithm described in this section can be interpreted as a parametric optimization problem with respect to $z$ (see Lemma 1).
In the context of Conditional SI, the parametric optimization problem was first used in the work by \citet{le2021parametric}, and this work can be interpreted as an application of their ideas to the multi-dimensional change-point detection problem.

\section{Experiments}
\label{sec:sec4}
We demonstrate the effectiveness of the proposed method through numerical experiments. 
In all experiments, we set the significance level at $\alpha = 0.05$.

\subsection{Synthetic Data Experiments}
We compared the following four methods:
\begin{itemize}
 \item Proposed-MC: the proposed method with minimal (optimal) conditioning;
 \item Proposed-OC: the proposed method with over-conditioning;
 \item Data splitting (DS): an approach that divides the dataset into odd- and even-numbered locations, and uses one for CP detection and the other for inference;
 \item Naive: naive inference without considering selection bias;
\end{itemize}
The goodness of each method was evaluated using false positive rate (FPR), conditional power.

\begin{figure}[tb]
	\begin{minipage}{.5\textwidth}
		\centering
		\includegraphics[width=.9\textwidth]{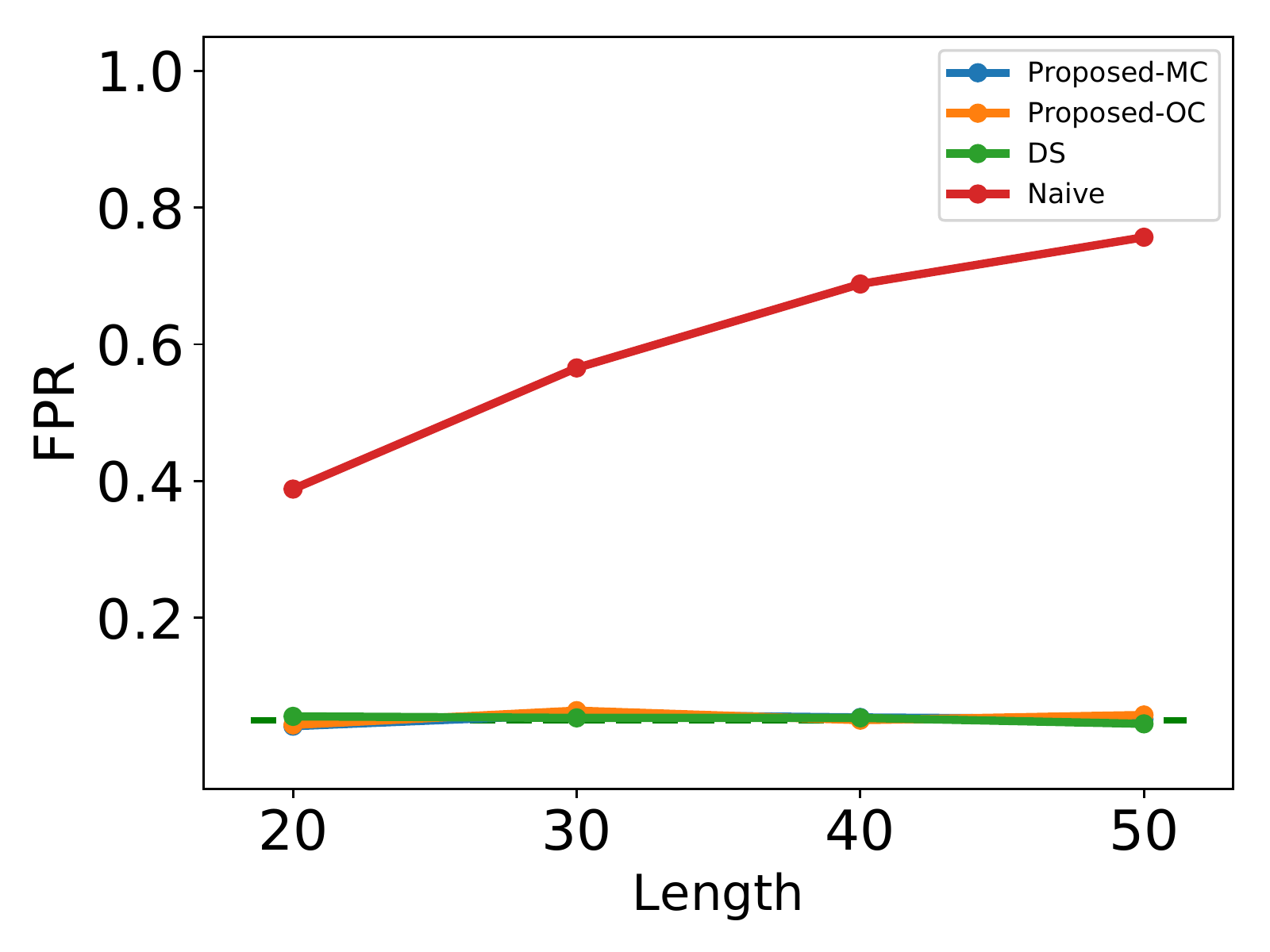}
		\subcaption{Independence}
	\end{minipage}
	\begin{minipage}{.5\textwidth}
		\centering
		\includegraphics[width=.9\textwidth]{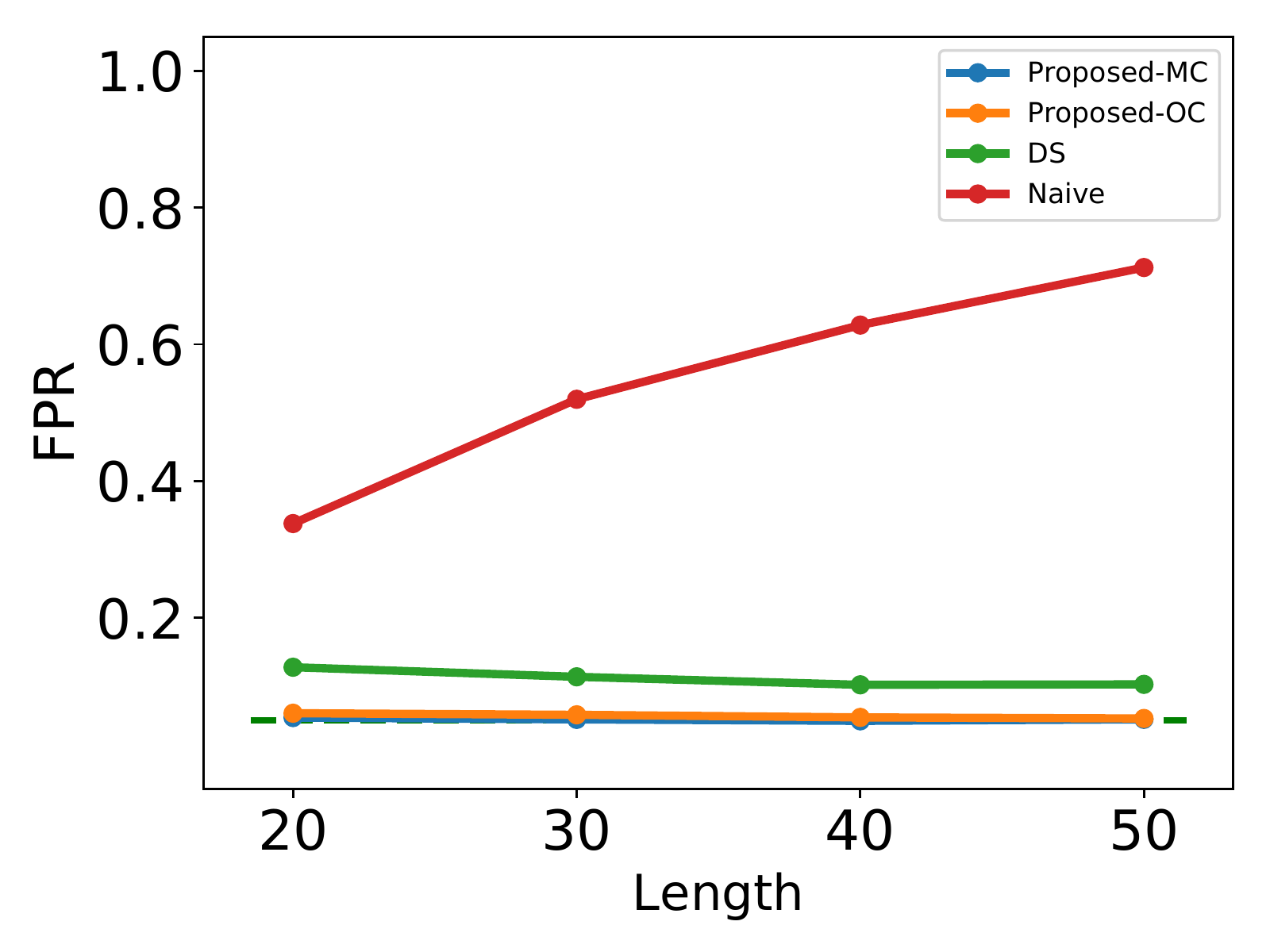}
		\subcaption{AR(1)}
	\end{minipage}
	\caption{FPR of synthetic data experiments with $N \in \{20, 30, 40, 50\}$.}
	\label{fig:FPR_len}
\end{figure}

\begin{figure}[tb]
	\begin{minipage}{.5\textwidth}
		\centering
		\includegraphics[width=0.9\textwidth]{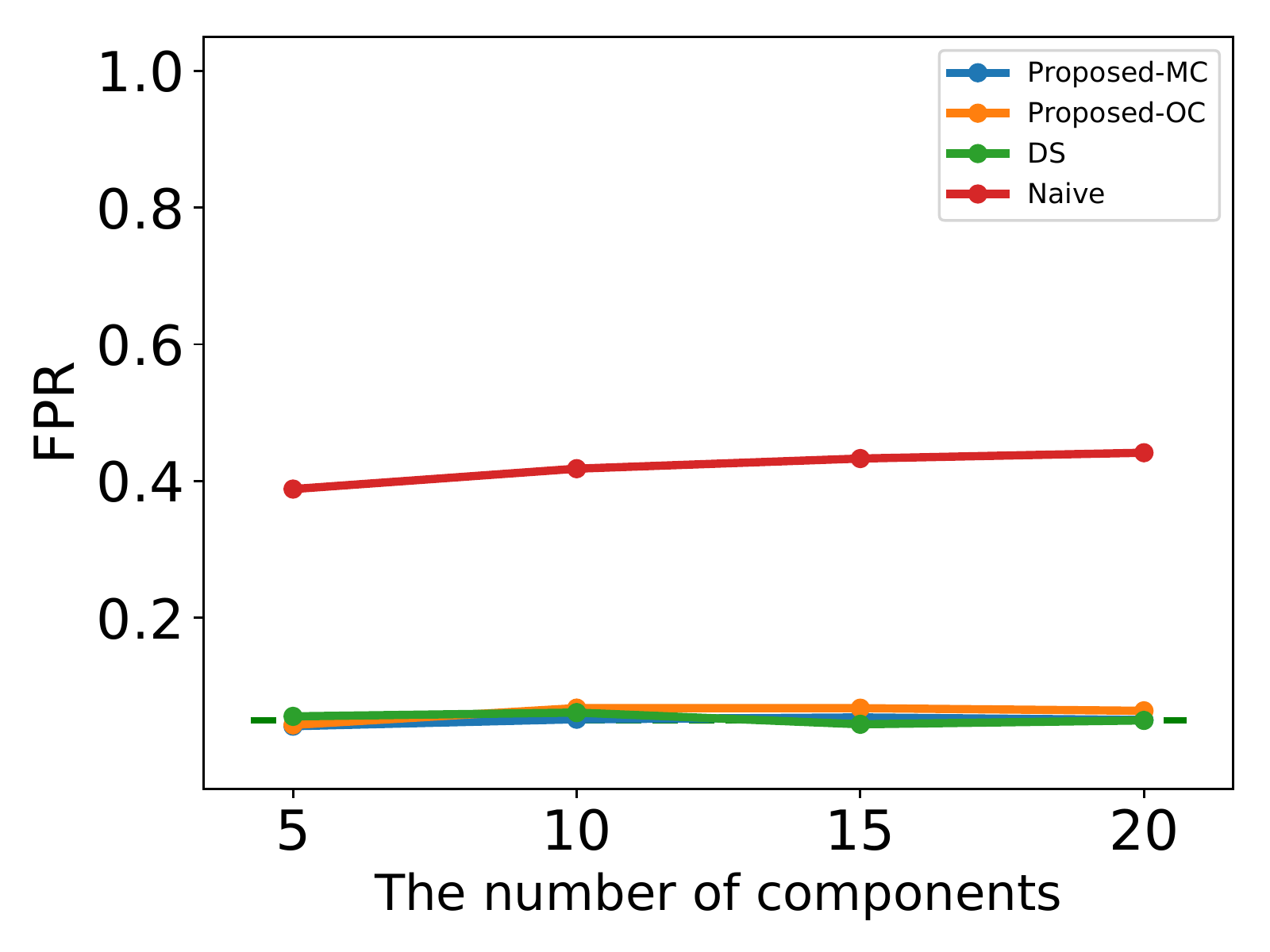}
		\subcaption{Independence}
	\end{minipage}
	\begin{minipage}{.5\textwidth}
		\centering
		\includegraphics[width=0.9\textwidth]{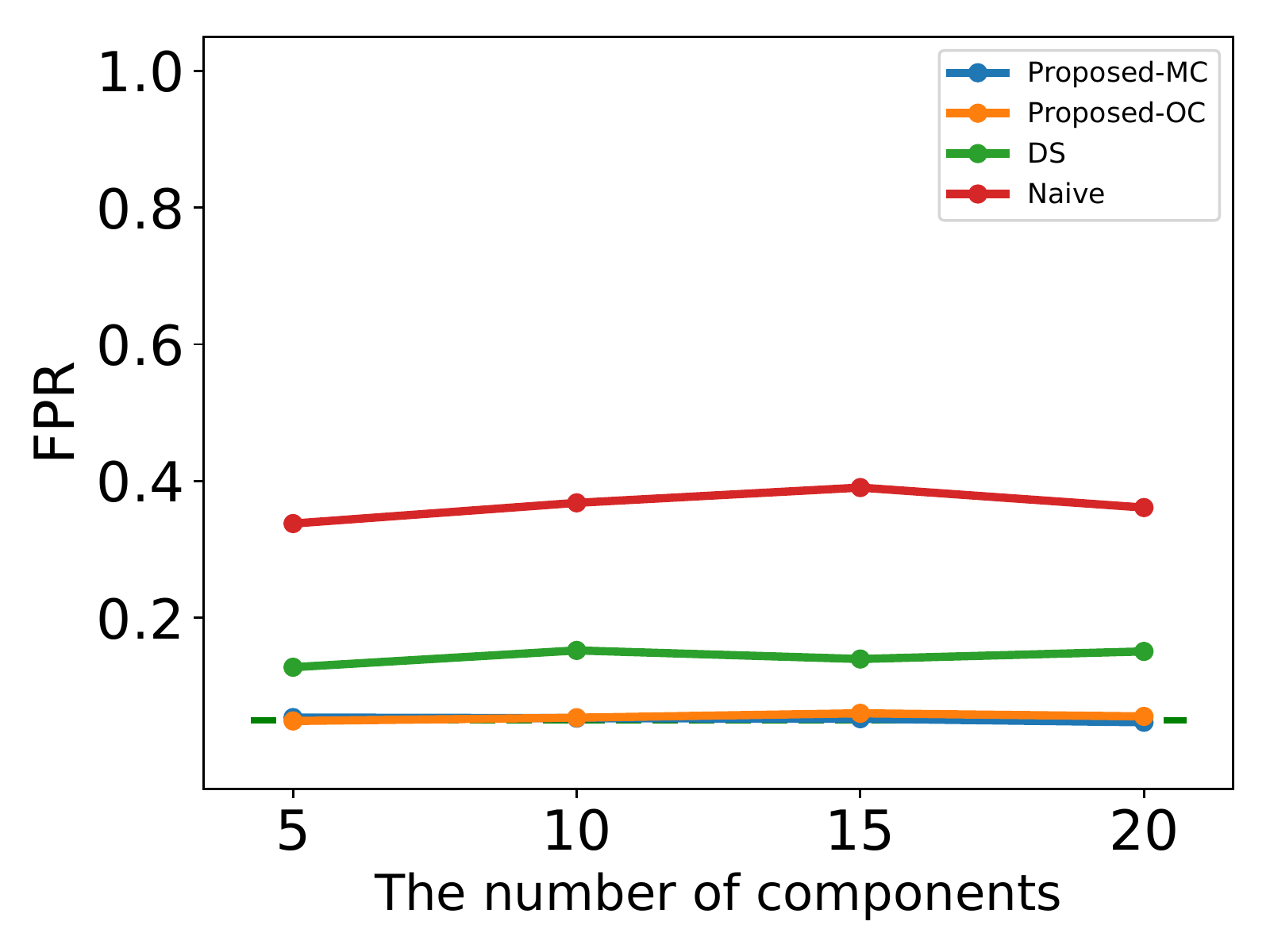}
		\subcaption{AR(1)}
	\end{minipage}
	\caption{FPR of synthetic data experiments with $D \in \{2, 3, 4, 5\}$.}
	\label{fig:FPR_seq}
\end{figure}

\paragraph{False Positive Rate}
FPR is the probability that the null hypothesis will be rejected when it is actually true, i.e., the probability of finding a wrong CPs.
In FPR experiments, synthetic data was generated from the normal distribution 
$\text{vec}(X) \sim N(\bm{0}, I_D\otimes\Sigma)$.
We considered the following two covariance matrices for $\Sigma$:
\begin{itemize}
 \item Independence: $\Sigma = \sigma^2\mathrm{I}_N$
 \item AR(1): $\Sigma = \sigma^2(\rho^{|i-j|})$
\end{itemize}
Here, we set $\sigma^2 = 1$ and $\rho = 0.5$.
We considered a setting in which the number of components is fixed with $D=5$ and the sequence length is changed among $N =20,30,40,50$. 
Similarly, we considered a setting where the sequence length is fixed with $N=30$ and the number of components is changed among $D = 5,10,15,20$.
We generated 1000 samples from $N(\bm{0}, I_D\otimes\Sigma)$, and detected CPs with 
hyperparameters $K = 2, L = 5$ and $W = 3$.
Then, we chose one detected location randomly and tested the changes at all the detected components. 
Figures \ref{fig:FPR_len} and \ref{fig:FPR_seq} 
show the results of FPR experiments. 
Naive method could not control FPR at the significance level. 
Similarly, since DS is a method that is valid only when each locations are independent each other, FPR could not be controlled in the case of AR(1).
On the other hand, the proposed methods (both proposed-MC and proposed-OC) could control FPR at the significance level. 

\begin{figure}[tb]
	\begin{minipage}{.5\textwidth}
		\centering
		\includegraphics[width=0.9\textwidth]{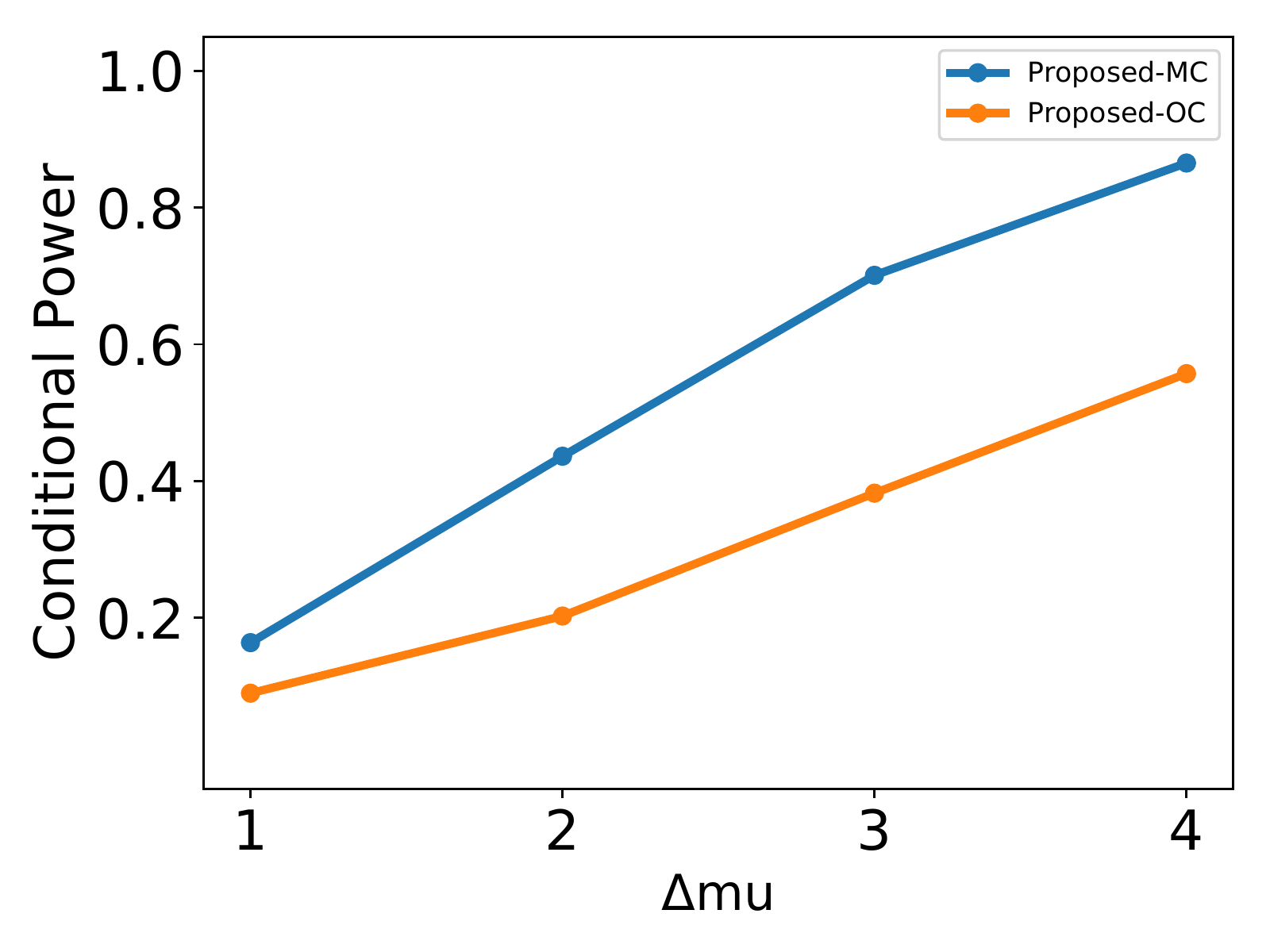}
		\subcaption{Independence}
	\end{minipage}
	\begin{minipage}{.5\textwidth}
		\centering
		\includegraphics[width=0.9\textwidth]{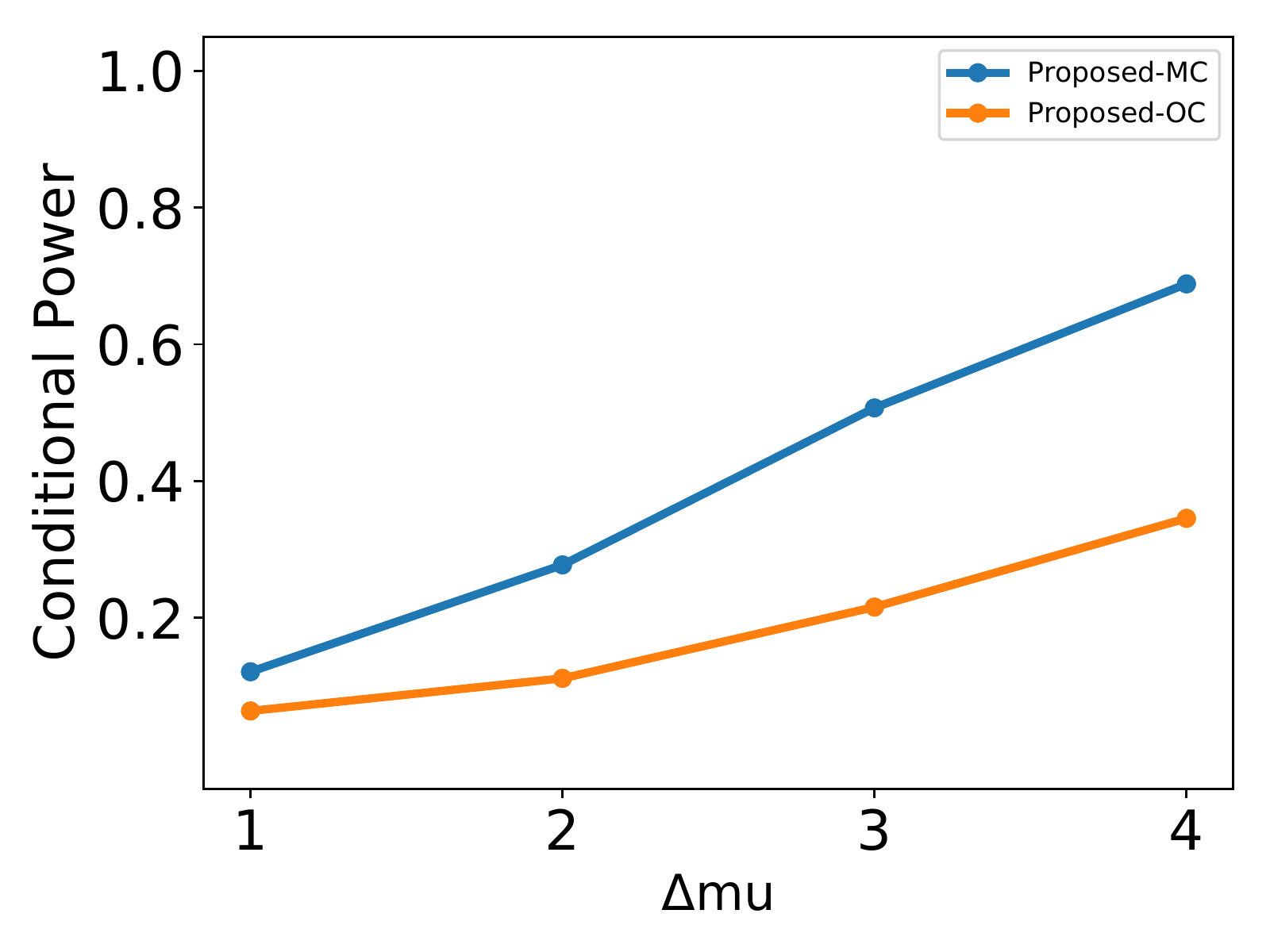}
		\subcaption{AR(1)}
	\end{minipage}
	\caption{Conditional power of synthetic data experiments.}
	\label{fig:CP}
\end{figure}

\paragraph{Conditional Power}
The power of conditional SI was evaluated by the proportion of correctly identified CP locations and components. 
We set the mean matrix
$M$
with 
$N = 30, D = 5$ 
as follows: 
\begin{align*}
M_{1, j} &= \begin{cases}
0 & (1\leq j \leq 10, 21 \leq j \leq 30) \\
\Delta\mu & (11 \leq j \leq 20)\end{cases},  \\
M_{2, j}, M_{3, j} &= \begin{cases}
0 & (1\leq j \leq 10) \\
\Delta\mu & (11 \leq j \leq 30)\end{cases},  \\
M_{4, j}, M_{5, j} &= \begin{cases}
\Delta\mu & (1\leq j \leq 20) \\
0 & (21 \leq j \leq 30)\end{cases}. \\
\end{align*}
Synthetic dataset was generated 1000 times from
$N({\rm vec}(M), I_D\otimes\Sigma)$, 
where 
$\Delta\mu \in \{1, 2, 3, 4\}$
and
the same covariance structures as in the FPR experiments were considered.
In this setting,
we detected locations with 
$K = 2, L = 5, W = 1$,
and  tested all the detected components. 
Here, 
proposed-MC and proposed-OC
were evaluated by using the following performance measure~\citep{hyun2018post}:
\begin{align*}
\text{Conditional Power} = \frac{\text{\#correctly detected \& rejected}}{\text{\#correctly detected}}.
\end{align*}
In this experiment, we defined that detection is considered to be correct if detected points are within $\pm 2$ of the true CP locations.
Figure \ref{fig:CP} shows the result of conditional power experiments.
The power of proposed-MC is much greater than that of proposed-OC, indicating that it is crucially important to resolve over-conditioning issues in conditional SI.



\subsection{Real data}
Here, we compared the performances of the proposed-MC and the naive method on two types of real datasets.
Due to the space limitation, we only show parts of the results: the complete results are presented in Appendix~\ref{app:C}.

\begin{figure}[p]
	\begin{minipage}{.5\textwidth}
		\centering
		\includegraphics[width=.99\linewidth]{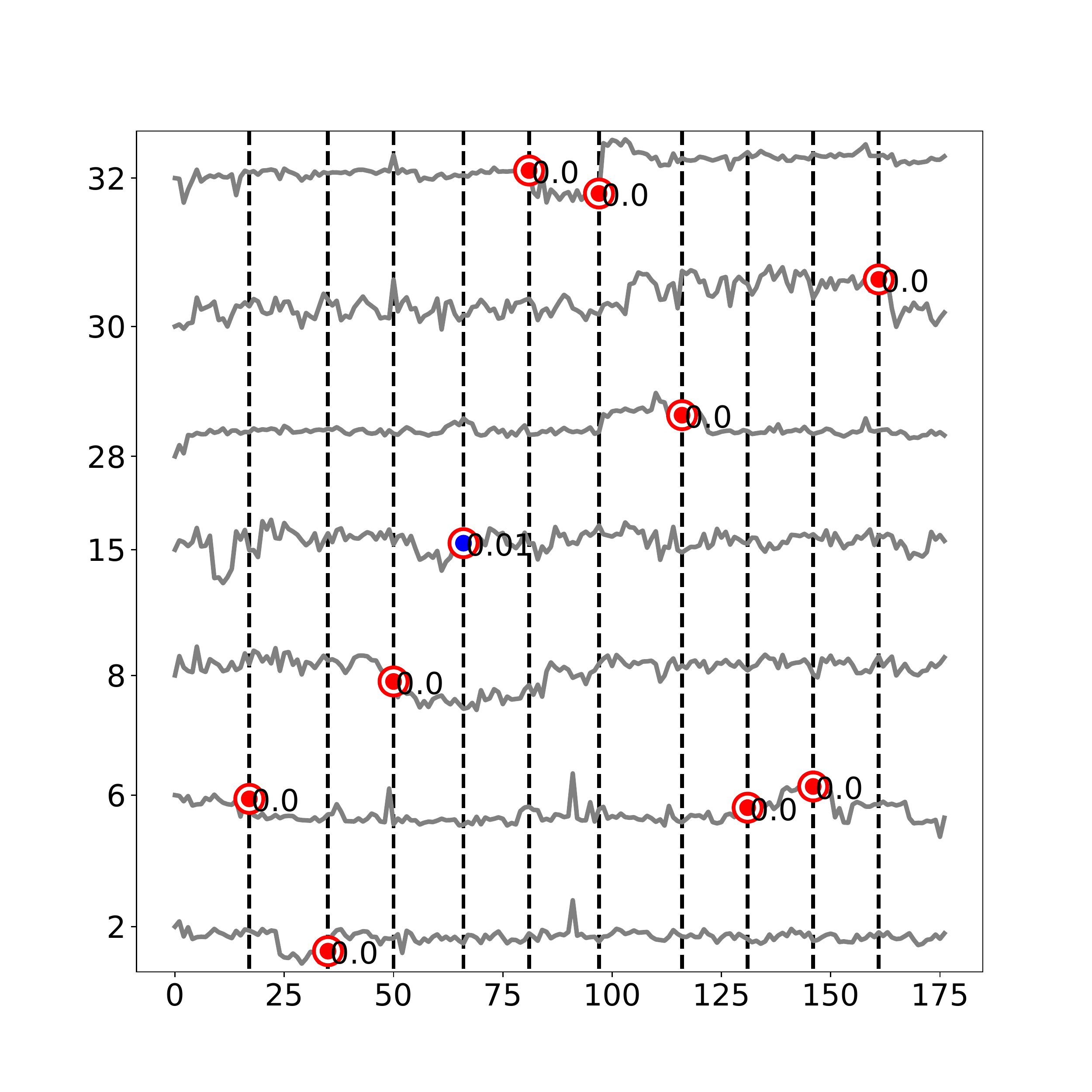}
		\subcaption{Independence, $W=0$}
	\end{minipage}
	\begin{minipage}{.5\textwidth}
		\centering
		\includegraphics[width=.99\linewidth]{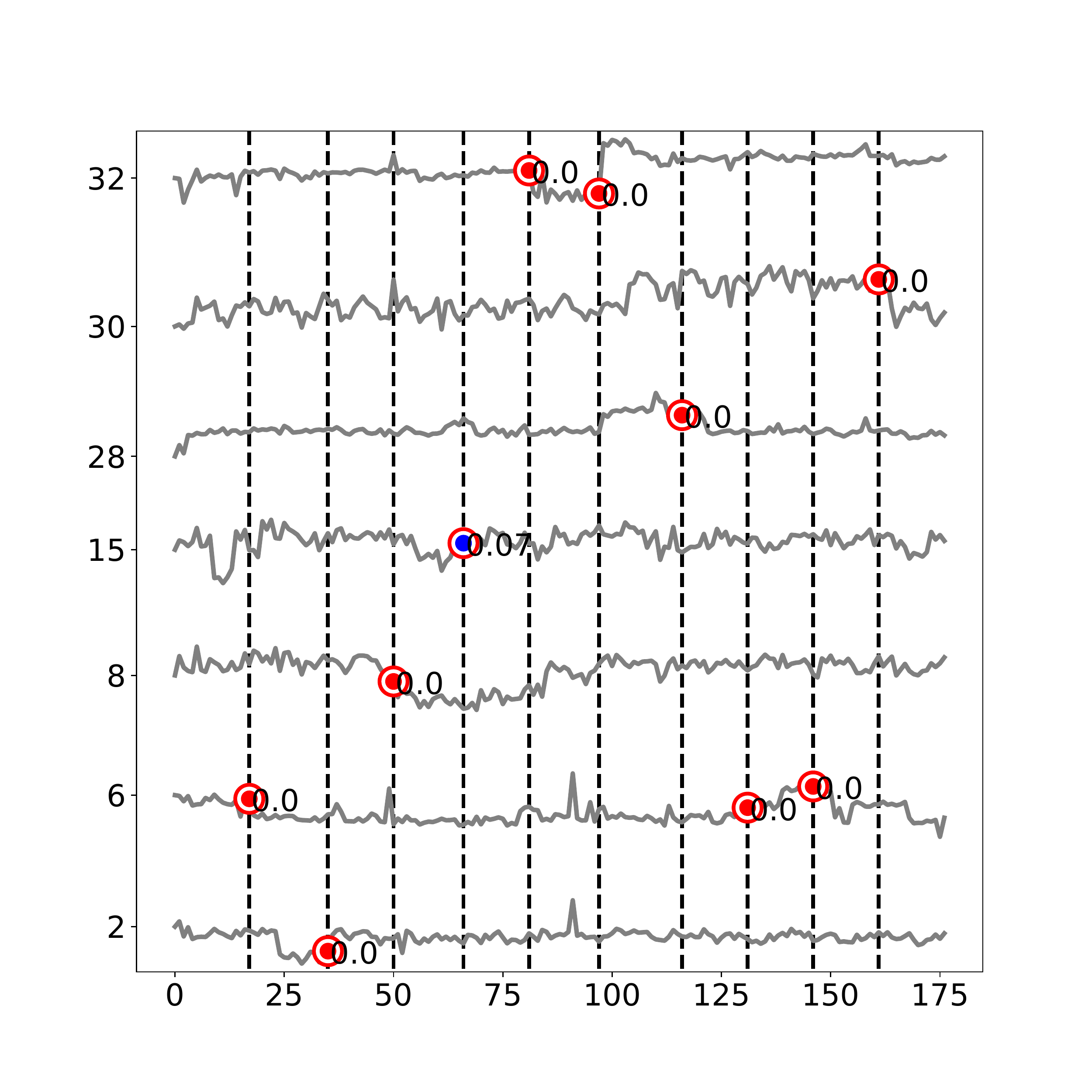}
		\subcaption{AR(1), $W=0$}
	\end{minipage}\\
	\begin{minipage}{.5\textwidth}
		\centering
		\includegraphics[width=.99\linewidth]{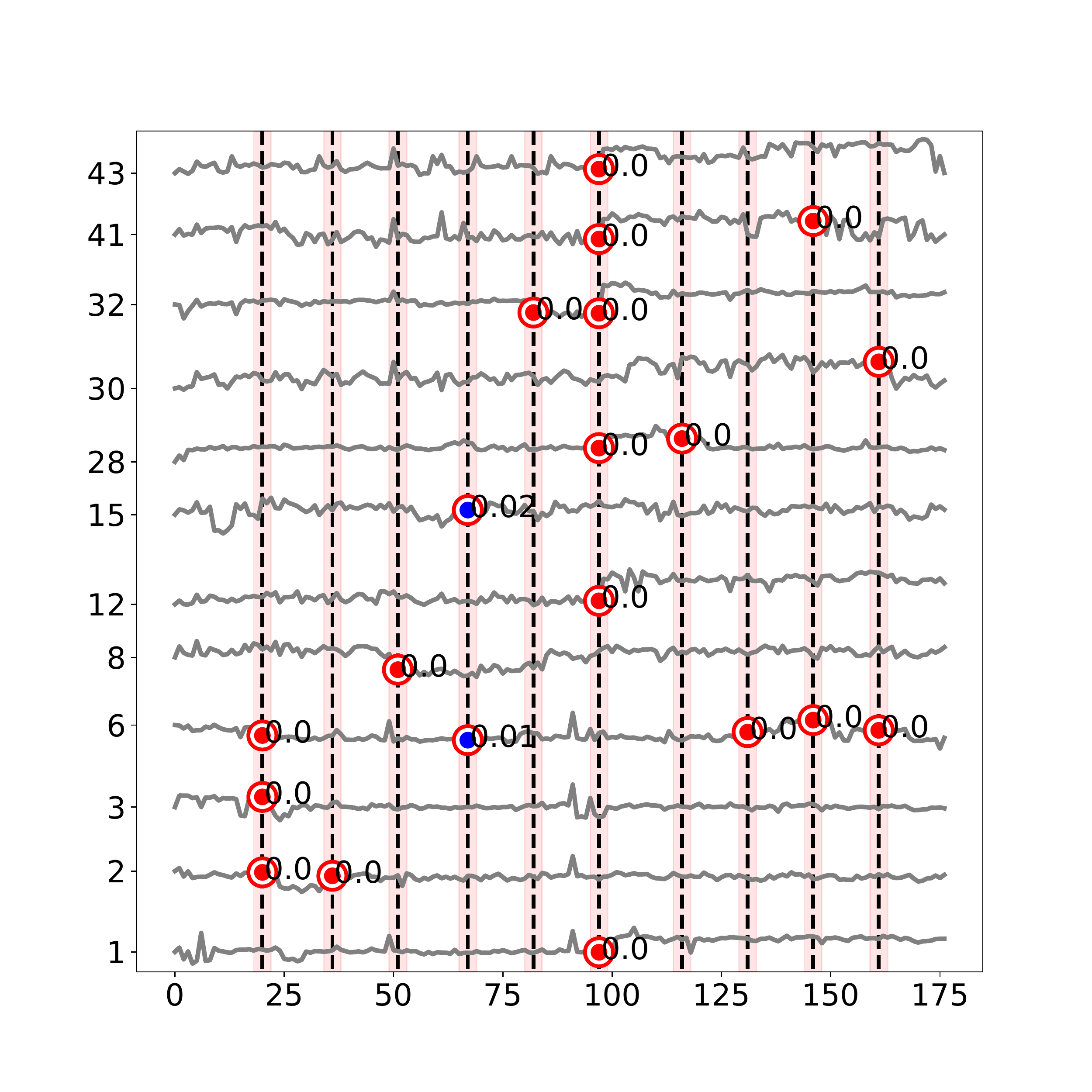}
		\subcaption{Independence, $W=2$}
	\end{minipage}
	\begin{minipage}{.5\textwidth}
		\centering
		\includegraphics[width=.99\linewidth]{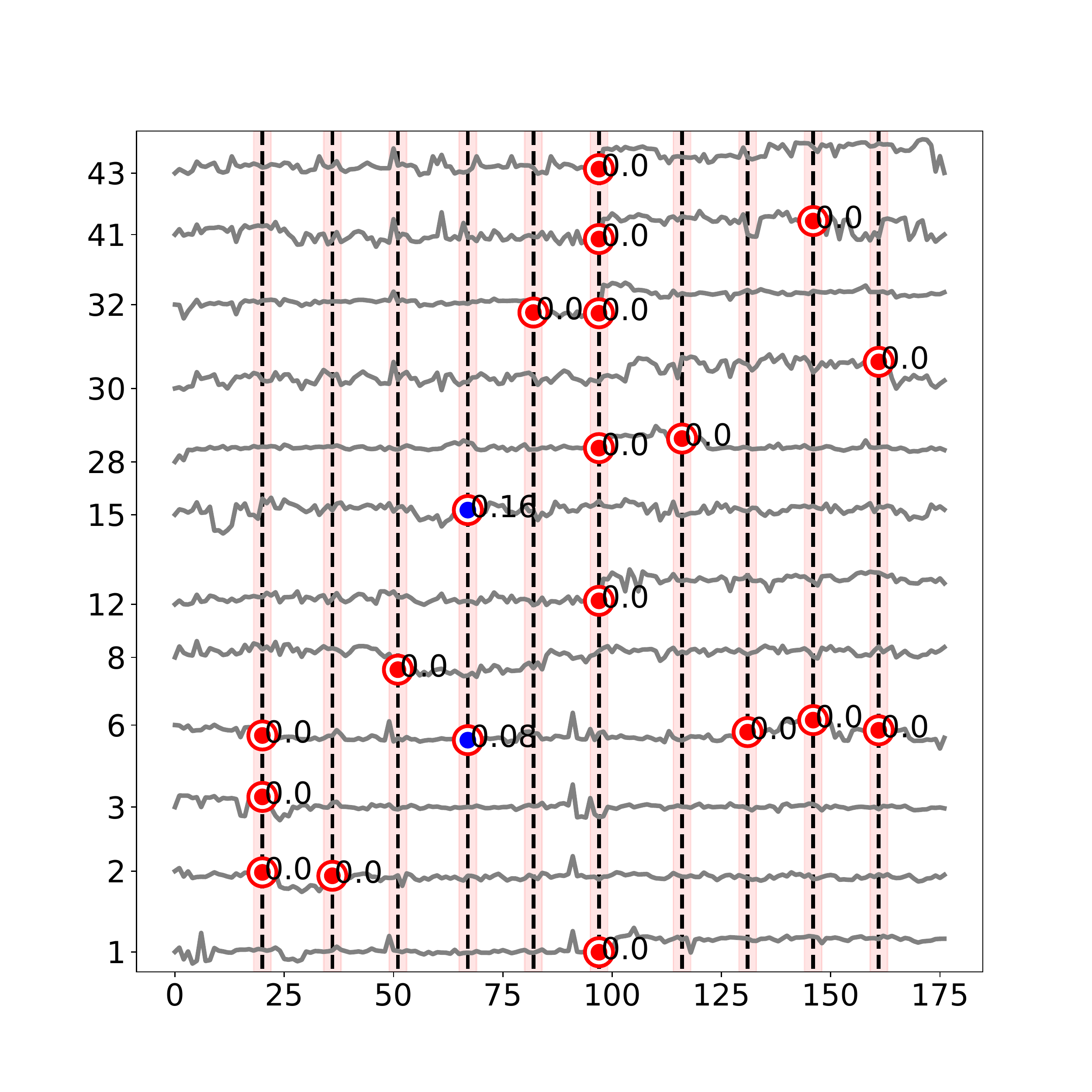}
		\subcaption{AR(1), $W=2$}
	\end{minipage}
	\caption{
 The results of chromosome 1 in Array CGH data analysis with $K=10$ and $L=15$. 
 Only the components (i.e., patients) in which a change is detected are shown. 
 The color of outer circle indicates the result of the naive method (red: statistically significant / blue: not statistically significant) although, in this example, the naive method declared that all the detected CP locations and components are statistically significant.
 The color of inner dot indicates the result of the proposed-MC (red: statistically significant / blue: not statistically significant).
 }
 \label{fig:array1}
\end{figure}

\paragraph{Array CGH}
Array Comparative Genomic Hybridization (Array CGH) is a molecular biology tool for identifying copy number abnormalities in cancer genomes.
We analyzed an Array CGH dataset for $D=46$ malignant lymphoma cancer patients studied in \citep{takeuchi2009potential}.
The covariance matrix $\Sigma$ (both for independent and AR(1) cases) by using Array CGH data for 18 healthy people for which it is reasonable to assume that there are no changes (copy number abnormalities). 
%
%
%
%
%
Figure \ref{fig:array1} shows a part of the experimental results.
In the naive test, all the detected CP locations and components are declared to be significant results. 
On the other hand, the proposed-MC suggests that some of the detected locations and components are not really statistically significant.

\begin{figure}[p]
	\begin{minipage}{.5\textwidth}
		\centering
		\includegraphics[width=.8\textwidth]{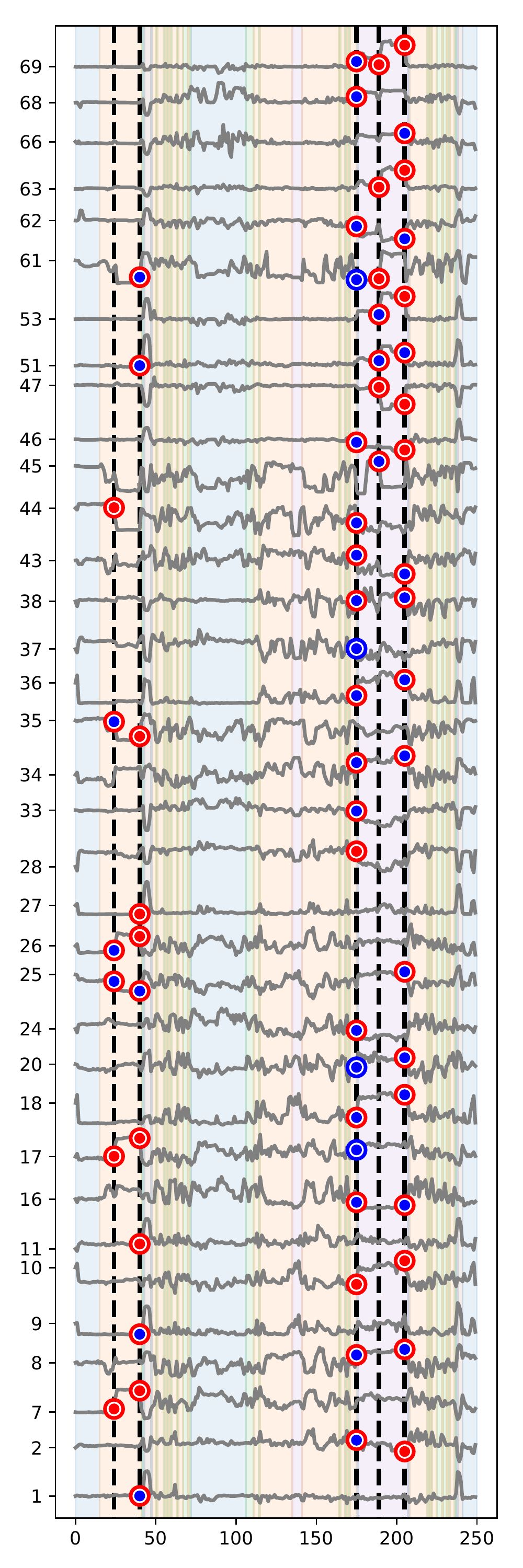}
		\subcaption{Independence}
	\end{minipage}
	\begin{minipage}{.5\textwidth}
		\centering
		\includegraphics[width=.8\textwidth]{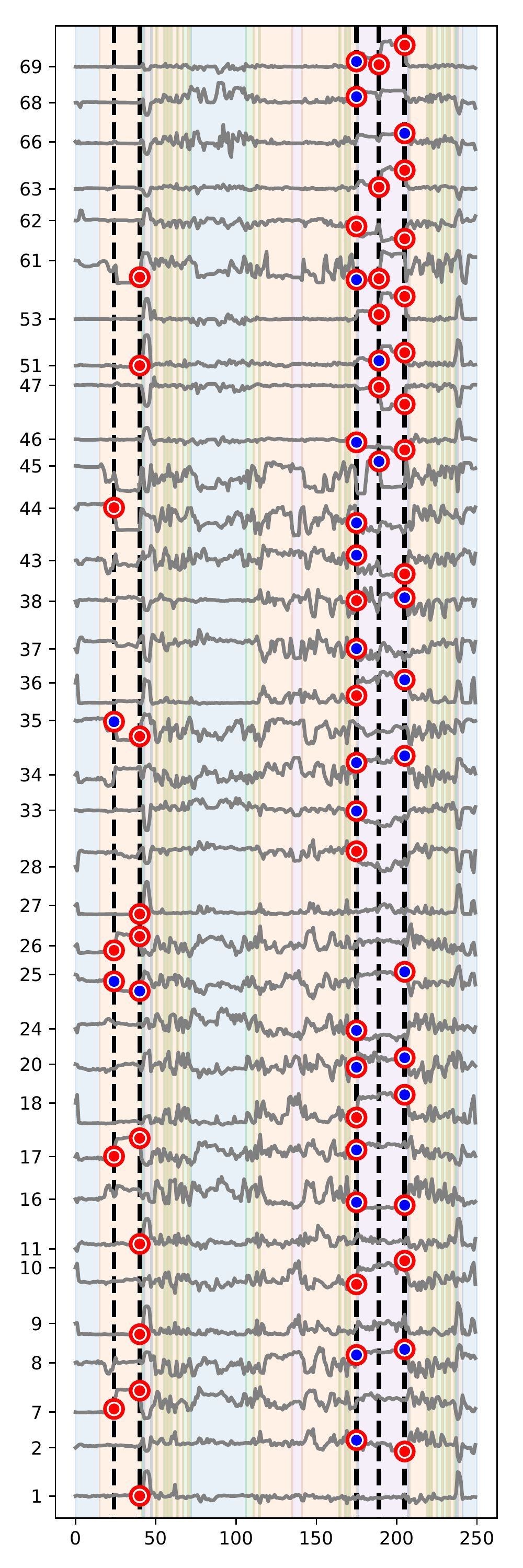}
		\subcaption{AR(1)}
	\end{minipage}
	\caption{S1, ADL1, $K=5, L=10, W=0$ (see the caption in Fig.~\ref{fig:array1} for the details).}
	\label{fig:HAR1}
\end{figure}

\begin{figure}[p]
	\begin{minipage}{.5\textwidth}
		\centering
		\includegraphics[width=.8\textwidth]{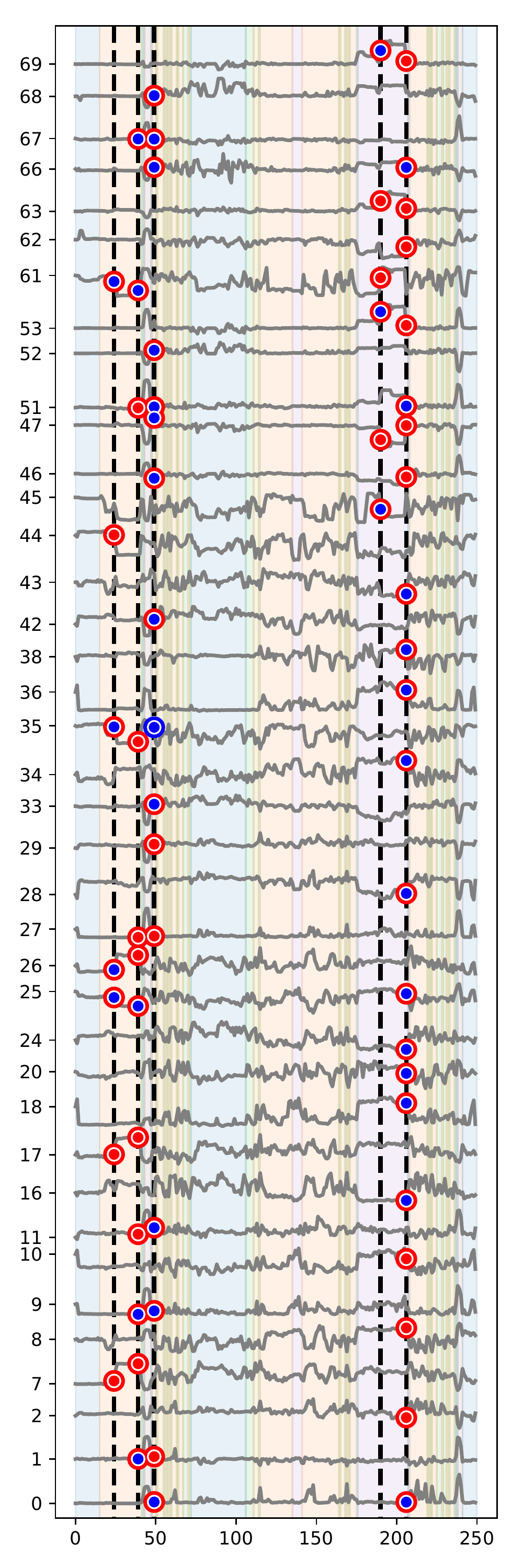}
		\subcaption{Independence}
	\end{minipage}
	\begin{minipage}{.5\textwidth}
		\centering
		\includegraphics[width=.8\textwidth]{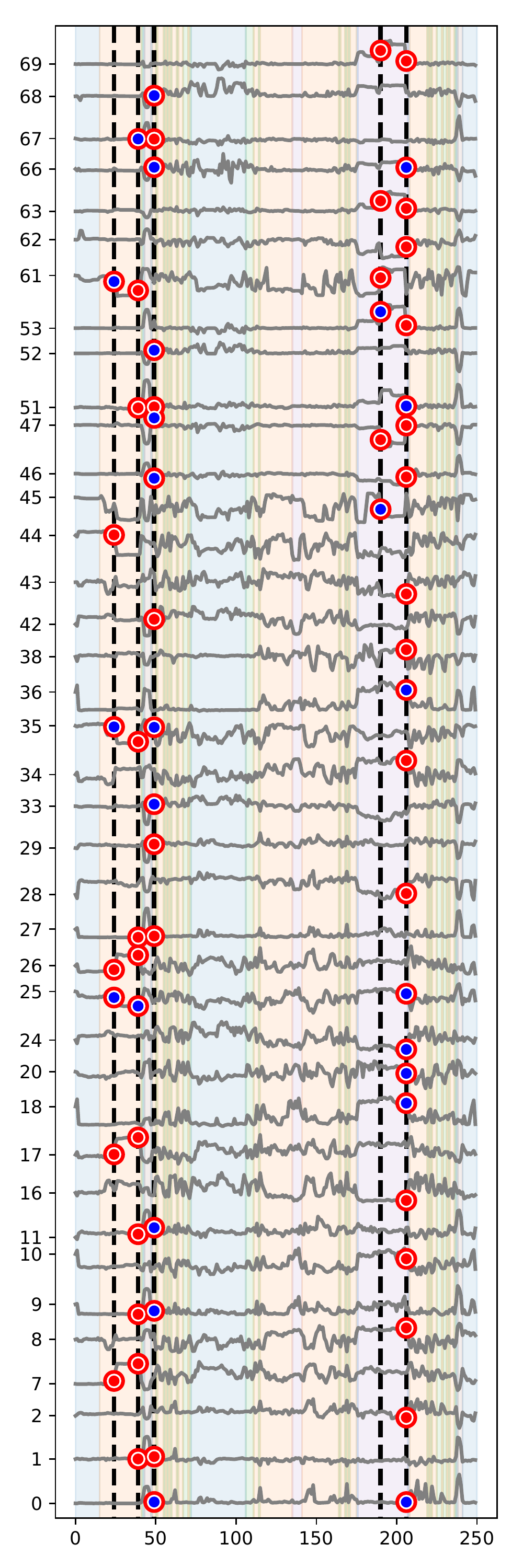}
		\subcaption{AR(1)}
	\end{minipage}
	\caption{S1, ADL1, $K=5, L=10, W=2$.}
	\label{fig:HAR2}
\end{figure}

\paragraph{Human Activity Recognition}
Here, we analyzed human activity recognition dataset studied in \citep{chavarriaga2013opportunity}, in which sensor signals attached in the humber bodies of for subjects were recorded in their daily activities. 
The sensor signals were recorded at $30$Hz, and locomotion annotations (Stand, Walk, Sit, Lie) and gesture annotations (e.g., Open Door and Close Door) were added manually.
%
%
We adopted the same preprocessing as \citep{hammerla2016deep} where we used the accelerometers of both arms and back and the IMU data of both feet and signals containing missing values were removed, which resulted in $D = 77$ sensor signals. 
The covariance structures were estimated by using the signals in the longest continuous locomotion state if ``Lie'' because it indicates the sleeping of the subjects, i.e., no changes exist.
%
%
%
%
In this experiment, the sequence length was compressed to 250 by taking the average so that there is no overlap in the direction of $N$, i.e., the dataset size was $N=250$ and $D=77$.
%
%
We used $8$ different combinations of hyperparameters
$(K, L, W)  \in \{5, 10\} \times \{10, 20\} \times \{0, 2\}$.
Figure \ref{fig:HAR1} and \ref{fig:HAR2} 
show parts of the experimental results.
The background colors of the figures correspond to the locomotion states, such as blue for None, orange for Stand, green for Walk, red for Sit, and purple for Lie.
Most of the detected CP locations and components are declared to be significant by the Naive method. 
On the other hand, the proposed-MC suggests that some of the detected locations and components are not really statistically significant.
%
%

\section{Sec5}
\label{sec:sec5}
In this paper, we present a valid exact (non-asymptotic) statistical inference method for testing the statistical significances of CP locations and components.
In many applied problems, it is important to identify the locations and components where changes occur.
By using the proposed method, it is possible to accurately quantify the reliability of those detected changes.

\newpage
\subsection*{Acknowledgement}
This work was partially supported by Japanese Ministry of Education, Culture, Sports, Science and Technology 16H06538, 17H00758, JST CREST JPMJCR1302, JPMJCR1502, RIKEN Center for Advanced Intelligence Project.

\clearpage

\appendix

\newpage

\section{Algorithm for UpdateDP, SortComp}
\label{app:A}
In this appendix, we present the pseudo-codes for SortComp and UpdateDP in Algorithm \ref{alg:cp_detection}.
\begin{algorithm}
 \caption{SortComp}
 \label{alg:sort_component}
 \begin{algorithmic}
  \REQUIRE $m$
  \STATE Define $\tilde{\bm{\theta}}^{(m)} = (\tilde{\theta}_1^{(m)}, \ldots, \tilde{\theta}_D^{(m)})^{\top} \in \mathbb{R}^D$
  \STATE where $\tilde{\bm{\theta}}^{(m)}$ is the vector of indices of the components sorted to satisfy \\$[Z_{m-L+1:m+L}^{\prime(\tilde{\theta}^{(m)}_d)}(m, W)]^2 > [Z_{m-L+1:m+L}^{\prime(\tilde{\theta}^{(m)}_{d+1})}(m, W)]^2$, for $d \in [D-1]$
  \ENSURE $\tilde{\bm{\theta}}^{(m)}$
 \end{algorithmic}
\end{algorithm}

\begin{algorithm}
 \caption{UpdateDP}
 \label{alg:dp_table_update}
 \begin{algorithmic}
  \REQUIRE $j, k, F_{(k-1, :)}, \cS$
  \STATE $F = \displaystyle\max_{\substack{Lk\leq m \leq j-L, \\ d \in [D]}}\left[F_{(k-1, m)} + C_{m-L+1:m+L}^{(\bm{\theta}^{(m)}_{:d})}(m, W)\right]$
  \STATE $\hat{\tau}, \hat{d} = \displaystyle\argmax_{\substack{Lk\leq m \leq j-L, \\ d \in [D]}}\left[F_{(k-1, m)} + C_{m-L+1:m+L}^{(\bm{\theta}^{(m)}_{:d})}(m, W)\right]$
  \ENSURE $F, \hat{\tau}, \hat{\bm{\theta}} = \tilde{\bm{\theta}}_{:d}^{(\hat{\tau})}$
 \end{algorithmic}
\end{algorithm}

\section{Lemma2}
\label{app:B}
In this appendix, we describe additional details of Lemma 2. 
\paragraph{Definition}
The matrices $A_{(m, d)}$, $B_{(k, j, m, d)}$ and the scalar $c_{(k, j, m, d)}$ in Lemma 2 are defined as follows. 
\begin{align*}
  A_{(m, d)}
  &= 
  \sqrt{2}
  (
  G_{(m, \{\tilde{\theta}^{(m)}_{d+1}\})} 
  - G_{(m, \{\tilde{\theta}^{(m)}_d\})}
  ), \\
  B_{(k, j, m, d)} 
  &= 
  -\sum_{k^{\prime}=1}^k
  G_{(\hat{\tau}_{k^{\prime}}^{(k, j)}, \hat{\bm \theta}_{k^{\prime}}^{(k, j)})} 
  + \sum_{k^{\prime}=1}^{k-1}
  G_{(\hat{\tau}_{k^{\prime}}^{(k-1, m)}, \hat{\bm \theta}_{k^{\prime}}^{(k-1, m)})}
  + G_{(m, \tilde{\bm \theta}_{:d}^{(m)})},\\
  c_{(k, j, m, d)} 
  &= 
  \sum_{k^{\prime}=1}^k\frac{\sqrt{2|\hat{\bm \theta}_{k^{\prime}}^{(k, j)}|}}{2} 
  - \sum_{k^{\prime}=1}^{k-1}\frac{\sqrt{2|\hat{\bm \theta}_{k^{\prime}}^{(k-1, m)}|}}{2} 
  - \frac{\sqrt{2d}}{2}, \\
  G_{(\tau, \bm{\theta})} 
  &= 
  \frac{V^2_{\tau-L+1, \tau, \tau+L}}{\sqrt{2|\bm{\theta}|}}
  \text{diag}(\bm 1_{D, \bm\theta})
  \otimes
  \left(
  \sum_{\Delta=-w}^w\frac{1}{L+\Delta}\bm{1}_{\tau-L+1:\tau+\Delta} 
  - \frac{1}{L - \Delta}\bm{1}_{\tau+\Delta+1:\tau+L}
  \right) \\
  &\hspace{50mm}
  \left(
  \sum_{\Delta=-w}^w\frac{1}{L+\Delta}\bm{1}_{\tau-L+1:\tau+\Delta} 
  - \frac{1}{L - \Delta}\bm{1}_{\tau+\Delta+1:\tau+L}
  \right)^{\top},
\end{align*}
where $\bm 1_{s:e}$ is an $N$-Dimensional binary vector in which the $i^{\text{th}}$ element is $1$ if $s \leq i \leq e$ and $0$ otherwise.

\paragraph{Proof}
The proof of Lemma 2 is as follows. 

\begin{proof}
 To prove Lemma 2,
 it suffices to
 show that 
 the event that the sorting result at location 
 $m$
 is as the same as the observed case 
 and
 the event that the $(k, j)^{\rm th}$ entry of the DP table
 is as the same as the observed case 
 are
 both written
 as intersections of quadratic inequalities of $z$
 as defined
 in 
 $\cE_{\text{sort}}^{(m)}$
 for
 $m \in [L, L-N]$
 and 
 $\cE_{\text{table}}^{(k, j)}$
 for
 $k\in[K]$, $j\in[Lk, N-L(K-k)]$,
 respectively. 

 First, we prove the claim on the sorting event. 
 Using the matrix
 $G_{(\tau, \bm \theta)}$
 defined above,
 we can write
 \begin{align*}
  [Z_{m-L+1:m+L}^{\prime(\tilde{\theta}_d^{(m)})}(m, W)]^2 &= \sqrt{2}\text{vec}(X)^{\top}G_{(m, \{\tilde{\bm \theta}_d^{(m)}\})}\text{vec}(X).
 \end{align*}
 Therefore,
 the sorting event is written as 
 \begin{align}
  \label{eq:quadra_sort}
  [Z_{m-L+1:m+L}^{\prime(\tilde{\theta}_d^{(m)})}(m, W)]^2 \geq [Z_{m-L+1:m+L}^{\prime(\tilde{\theta}_{d+1}^{(m)})}(m, W)]^2
  ~\Leftrightarrow~
  \text{vec}(X)^{\top}A_{(m, d)}\text{vec}(X) \le 0.
 \end{align}
 for $d \in [D-1]$.
 By restricting on a line
 ${\rm vec}(X) = \bm a + \bm b z, z \in \mathbb{R}$,
 the range of
 $z$
 with which the sorting results are as the same as the observed case 
 is written as 
 \begin{align}
  \label{eq:lemma2_z1}
  z \in \cE_{\text{sort}}^{(m)} = \bigcap_{d=1}^{D-1}\{z ~|~(\bm{b}^{\top}A_{(m, d)}\bm{b})z^2 + 2(\bm{b}^{\top}A_{(m, d)}\bm{a})z + \bm{a}^{\top}A_{(m, d)}\bm{a} &\leq 0\}.
 \end{align}

 Next,
 we prove the claim on the DP table. 
 Let 
 \begin{align*}
  C_{s:e}^{(\bm{\theta})}(\tau, W) &= \text{vec}(X)^{\top}G_{(\tau, \bm{\theta})}\text{vec}(X) - \frac{\sqrt{2|\bm{\theta}|}}{2}.
 \end{align*}
 Then,
 the 
 $(k, j)^{\rm th}$
 element of the DP table $F$ is written as 
 \begin{align*}
  F_{(k, j)} &= \sum_{k^{\prime}=1}^kC_{\hat{\tau}_{k^{\prime}}^{(k, j)}-L+1:\hat{\tau}_{k^{\prime}}^{(k, j)}+L}^{(\hat{\bm{\theta}}_{k^{\prime}}^{(k, j)})}(\hat{\tau}_{k^{\prime}}^{(k, j)}, W).
 \end{align*}
 Therefore, 
 the DP table event is written as 
 \begin{align}
  \label{eq:quadra_table}
  F_{(k, j)} \geq F_{(k-1, m)} + C_{m-L+1:m+L}^{(\tilde{\bm{\theta}}_{:d}^{(m)})}(m, W), ~m \in [L(k+1), j-L], d \in [D]. 
 \end{align}
 By restricting on a line
 ${\rm vec}(X) = \bm a + \bm b z, z \in \mathbb{R}$,
 the range of
 $z$
 with which the
 $(k, j)^{\rm th}$
 entry of the 
 DP table is as the same as the observed case 
 is written as 
 \begin{align}
  \label{eq:lemma2_z2}
  z \in 
  \cE_{\text{table}}^{(k, j)} = \bigcap_m\bigcap_d\{Z~|~(\bm{b}^{\top}B_{(k, j, m, d)}\bm{b})z^2 + 2(\bm{b}^{\top}B_{(k, j, m, d)}\bm{a})z + \bm{a}^{\top}B_{(k, j, m, d)}\bm{a} + c_{(k, j, m, d)} \leq 0\}.
 \end{align}

 From 
 \eq{eq:lemma2_z1}
 and
 \eq{eq:lemma2_z2}, 
 the subset
 $\cZ_v \subseteq \RR$
 is
 represented
 as claimed in Lemma 2.
\end{proof}

\section{Additional information about the experiments}
\label{app:C}
In this appendix, we describe the details of the experimental setups and additional experimental results.
\subsection{Additional information about experimental setups}
We executed all the the experiments on Intel(R) Xeon(R) Gold 6230 CPU \@ 2.10GHz, and Python 3.8.8 was used all through the experiments.
The two proposed methods,
Proposed-MC
and
Proposed-OC,
are implemented as conditional SIs with the following conditional distributions: 
\begin{itemize}
 \item Proposed-MC: $\bm \eta_{k, h}^{\top}{\rm vec}(X) ~|~(\tau_k, \theta_h^k) \in \cM(X)$;
 \item Proposed-OC: $\bm \eta_{k, h}^{\top}{\rm vec}(X) ~|~\cA(X^{\rm obs}) = \cA(X)$.
\end{itemize}

\subsection{Additional information about synthetic data experiments}
\begin{figure}[tb]
	\centering
	\begin{minipage}{.40\linewidth}
		\centering
		\includegraphics[width=0.9\textwidth]{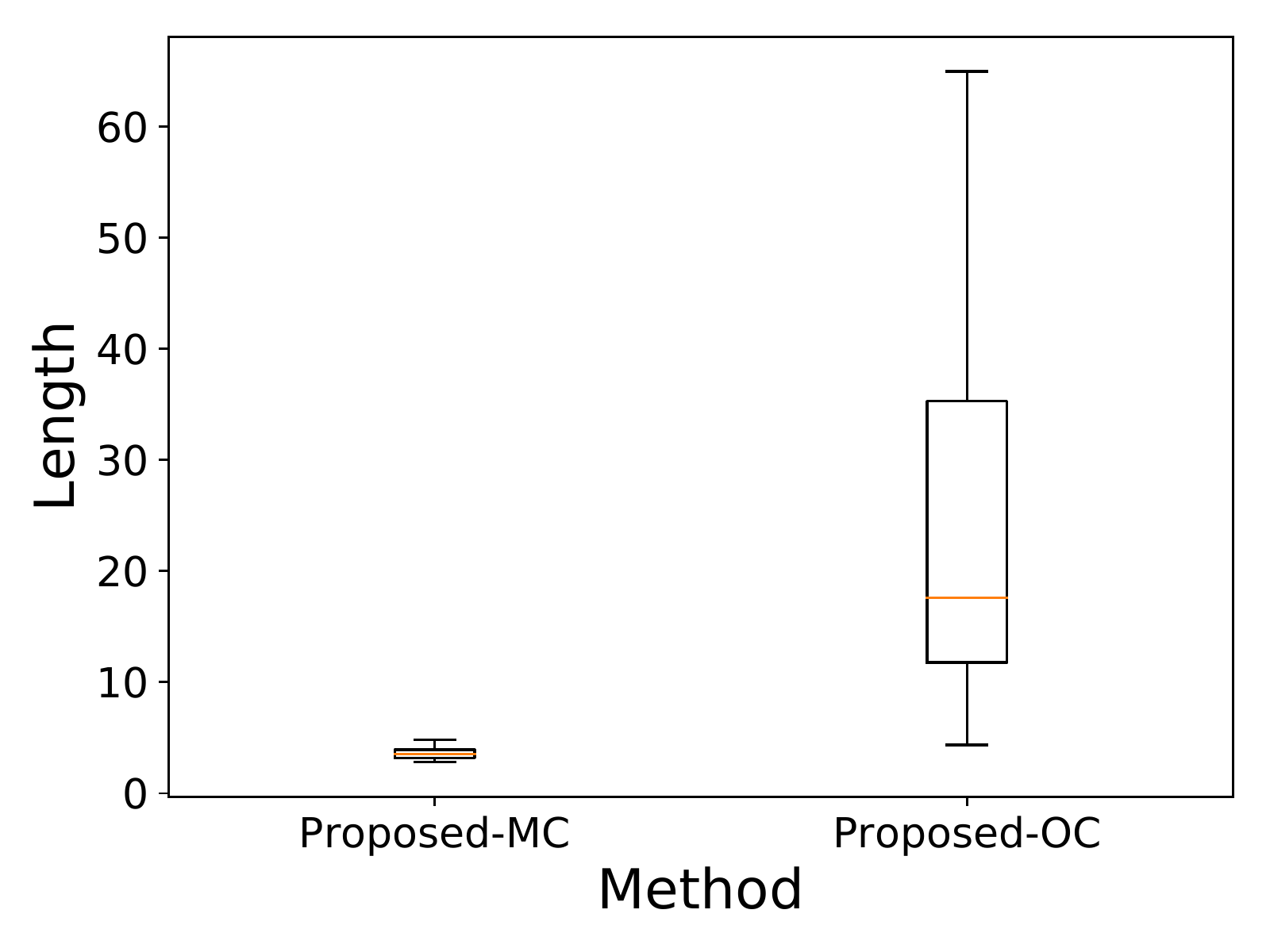}
		\subcaption{Independence}
	\end{minipage}
	\begin{minipage}{.40\linewidth}
		\centering
		\includegraphics[width=0.9\textwidth]{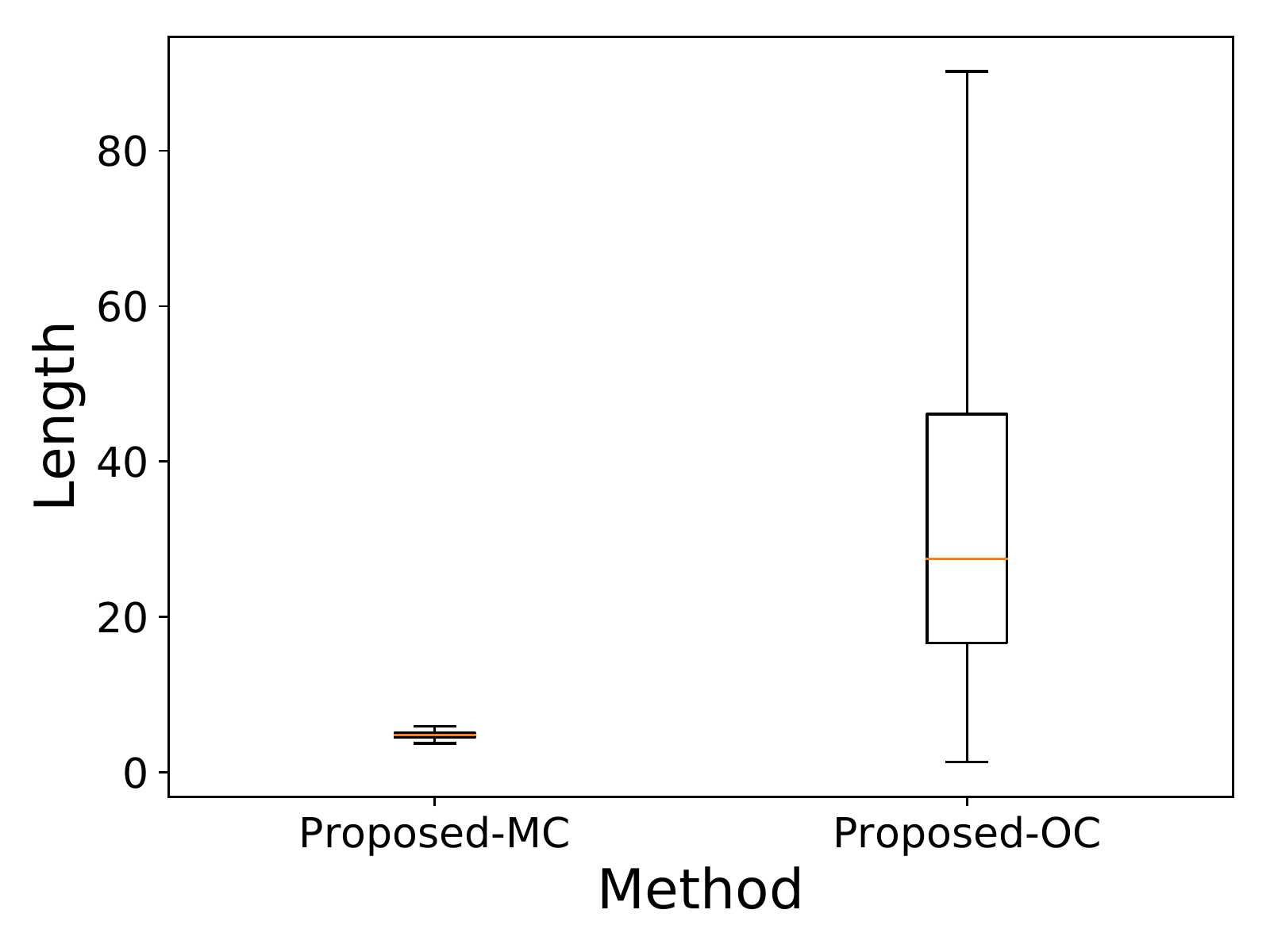}
		\subcaption{AR(1)}
	\end{minipage}
	\caption{Selective CI lengths in synthetic data experiments.}
	\label{fig:CI}
\end{figure}

In the synthetic data experiments,
in addition to FPR and conditional power, 
we also compared selective confidence intervals (CI)
of
Proposed-MC
and
Proposed-OC.
For comparison of selective CIs, 
we set the mean matrix
$M$
of the multi-dimensional sequence
$X$
with 
$N = 30, D = 5$ 
as follows: 
\begin{align*}
M_{1, j} &= \begin{cases}
4 & (1\leq j \leq 10) \\
2 & (11 \leq j \leq 20) \\
0 & (21 \leq j \leq 30) \end{cases}, \\
M_{2, j}, M{3, j} &= \begin{cases}
2 & (1\leq j \leq 10) \\
0 & (11 \leq j \leq 30)\end{cases} ,\\
M_{4, j}, M_{5, j} &= \begin{cases}
0 & (1\leq j \leq 30) \end{cases}. \\
\end{align*}
We considered the same covariance structures as in the experiments for conditional power. 
For each setting, we generated 100  multi-dimensional sequence samples.
Then, the lengths of the 95\% CIs in proposed-MC and proposed-OC were compared.
Figure~\ref{fig:CI} shows the result, which are consistent with the results of conditional power\footnote{Here, a few outliers are removed for visibility.}.

\subsection{Additional information about real data experiments}
Here,
we present the detail experimental setups and additional experimental results
of
the two real data experiments. 

\begin{figure}[tbp]
	\begin{minipage}{.5\textwidth}
		\centering
		\includegraphics[width=.9\linewidth]{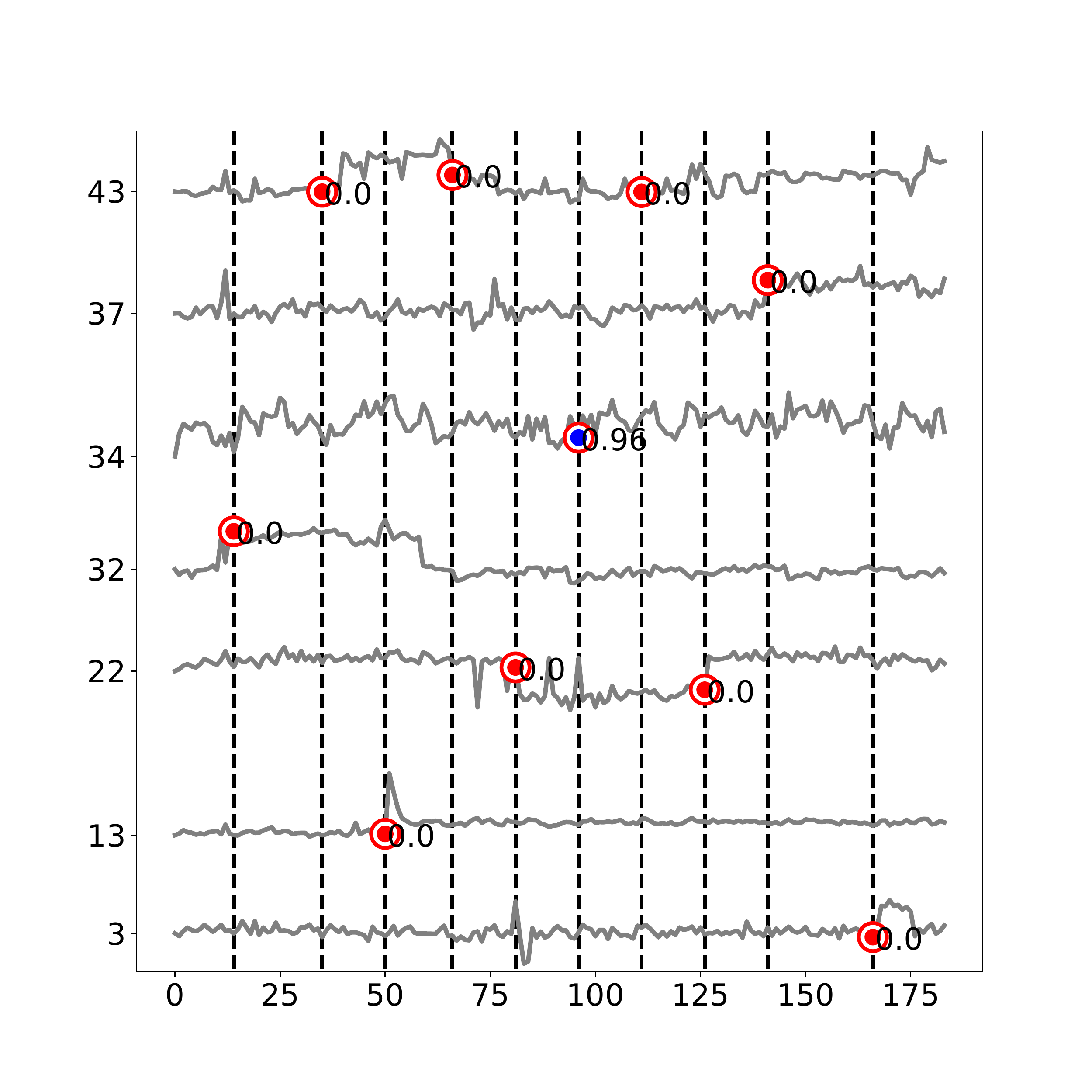}
		\subcaption{Independence, $W=0$}
	\end{minipage}
	\begin{minipage}{.5\linewidth}
		\centering
		\includegraphics[width=.9\linewidth]{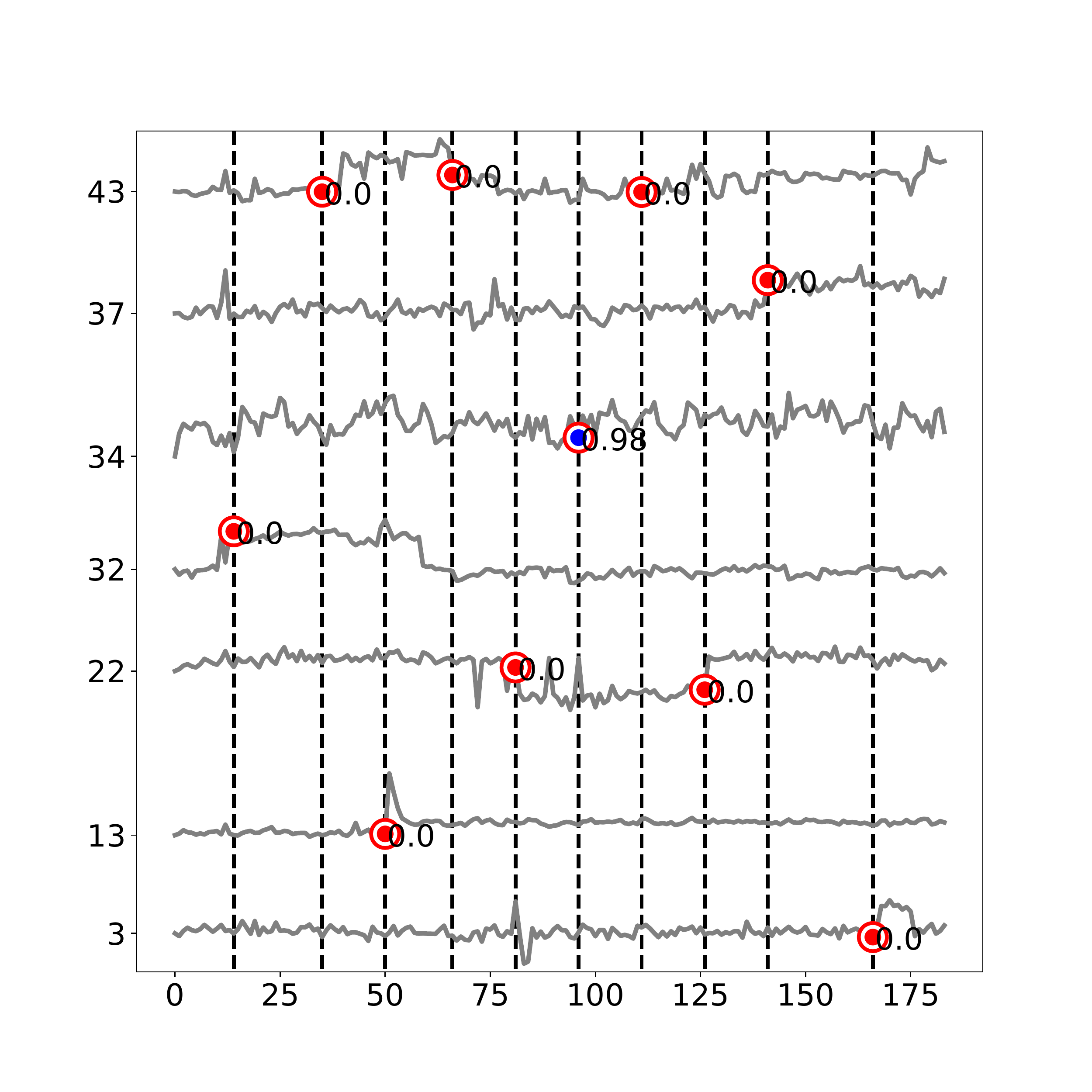}
		\subcaption{AR(1), $W=0$}
	\end{minipage}
	\begin{minipage}{.5\textwidth}
		\centering
		\includegraphics[width=.9\linewidth]{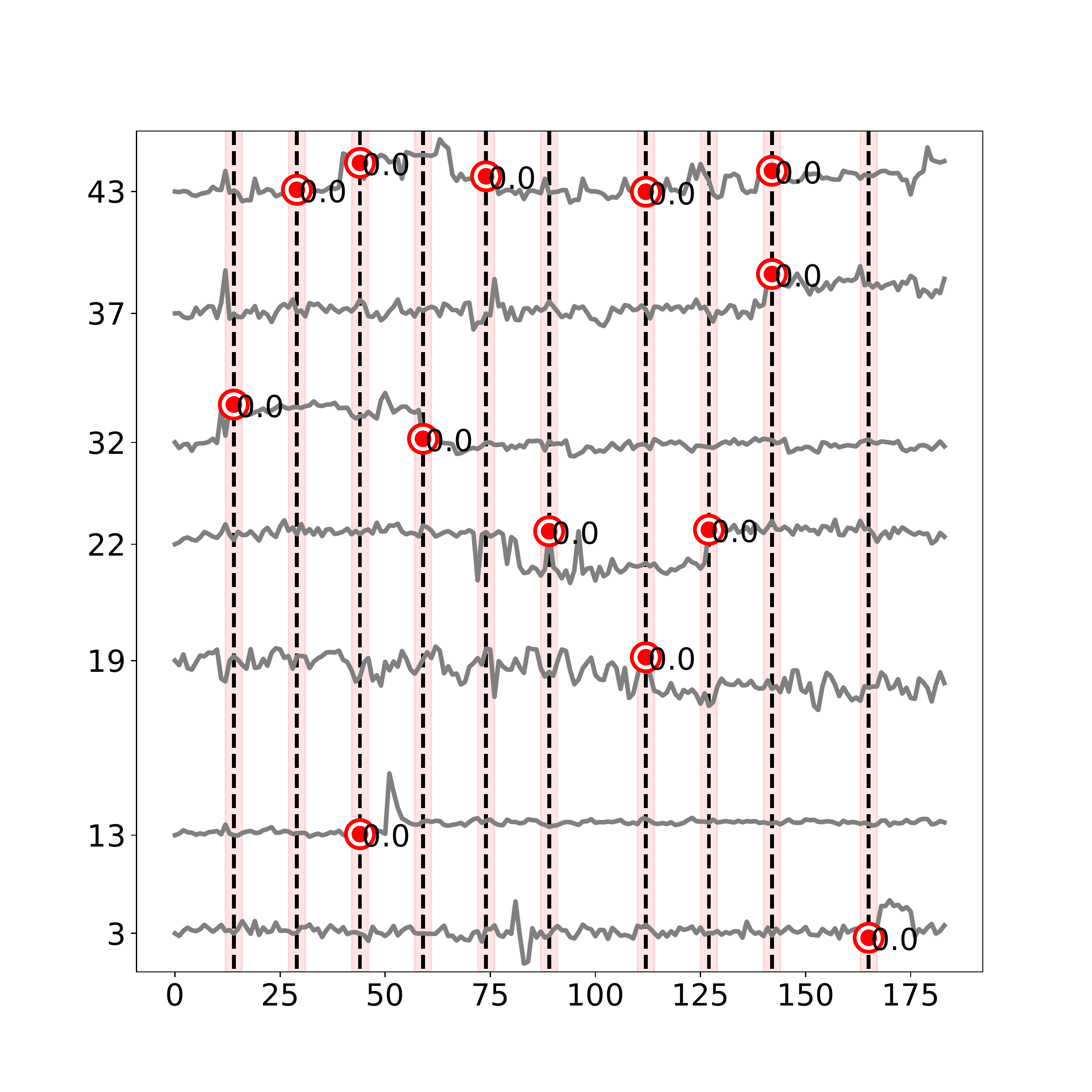}
		\subcaption{Independence, $W=2$}
	\end{minipage}
	\begin{minipage}{.5\linewidth}
		\centering
		\includegraphics[width=.9\linewidth]{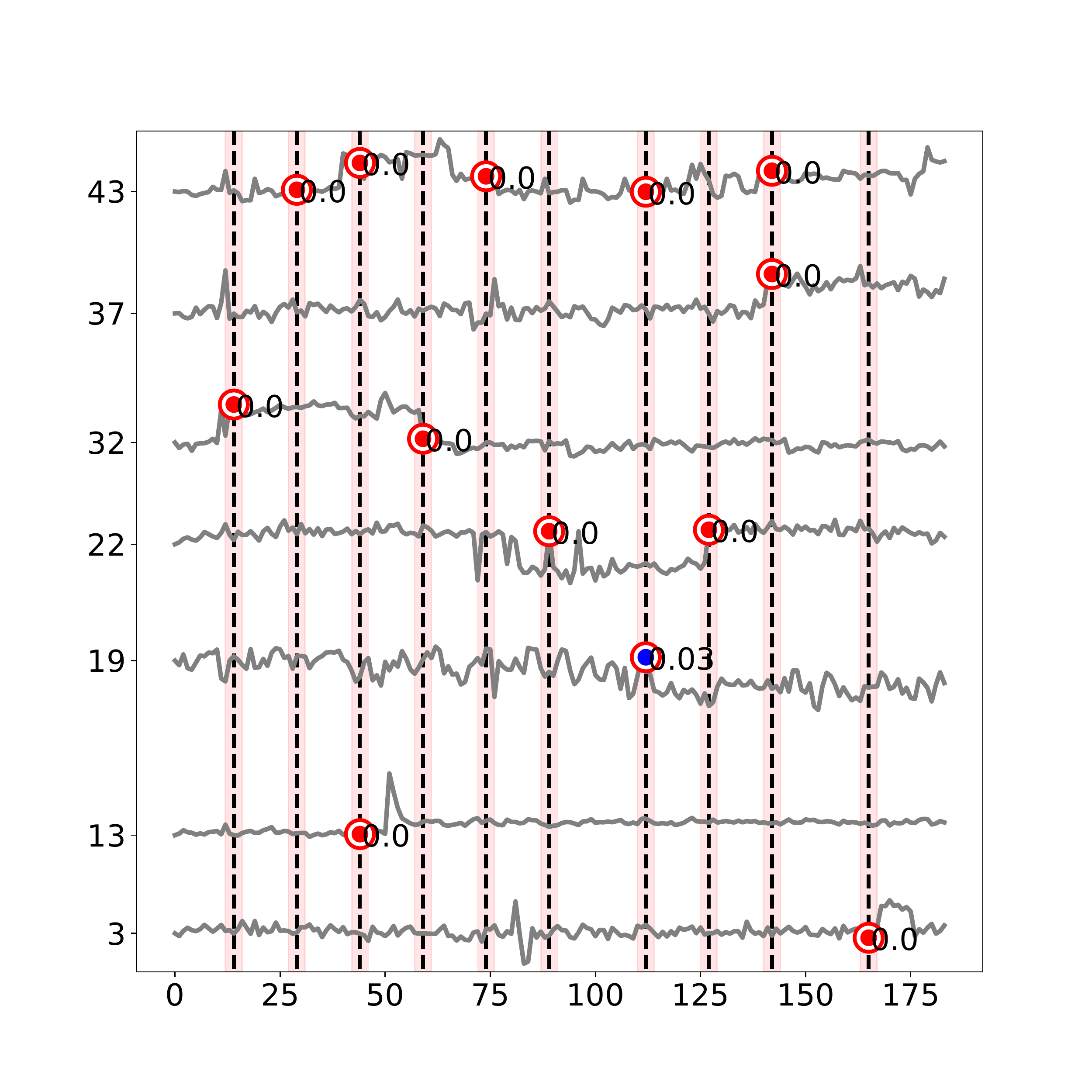}
		\subcaption{AR(1), $W=2$}
	\end{minipage}
	\caption{The results on Array CGH for chromosome 2 with $K=10$ and $L=15$.}
	\label{fig:array_chrom2}
\end{figure}

\begin{figure}[tbp]
	\begin{minipage}{.5\textwidth}
		\centering
		\includegraphics[width=.9\linewidth]{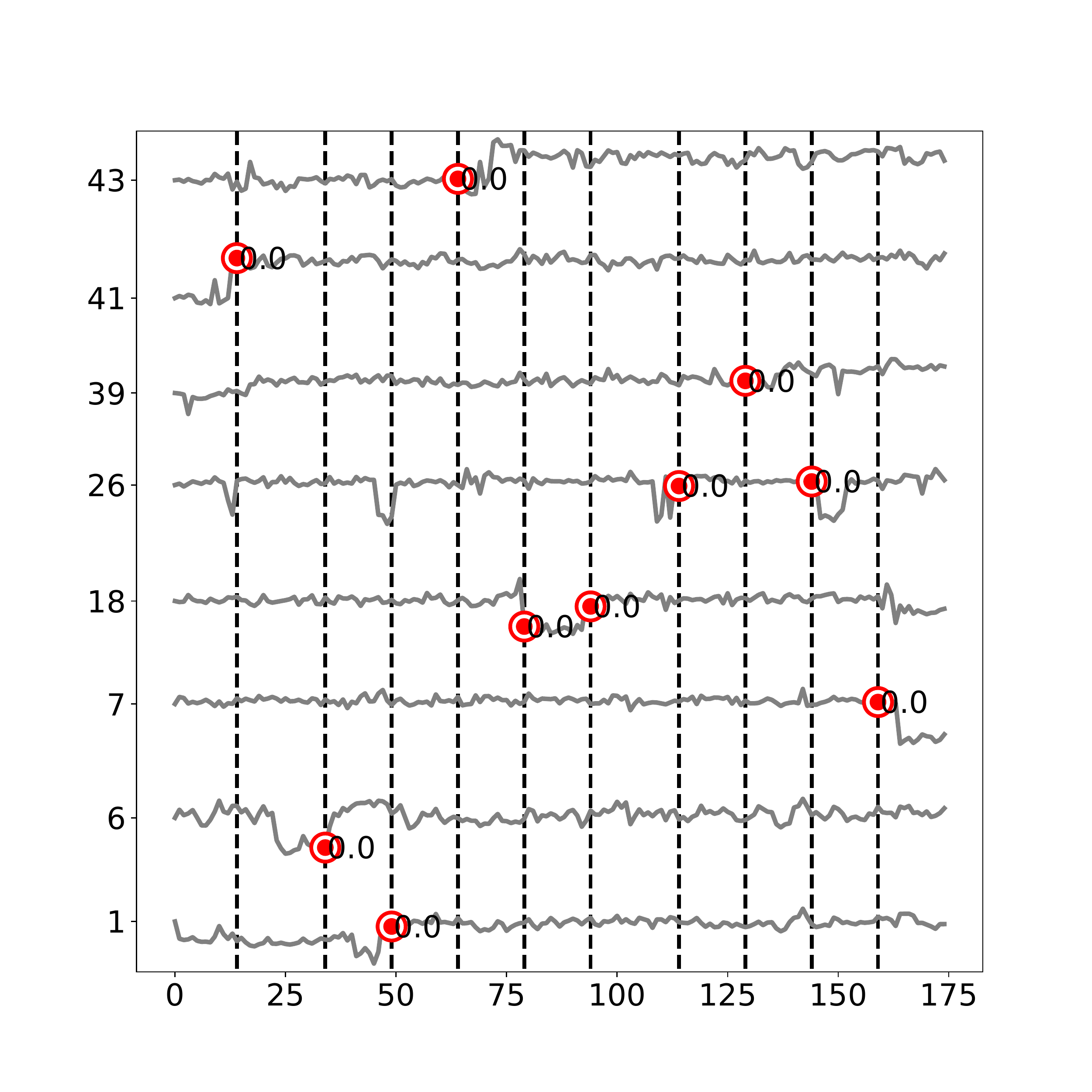}
		\subcaption{Independence, $W=0$}
	\end{minipage}
	\begin{minipage}{.5\linewidth}
		\centering
		\includegraphics[width=.9\linewidth]{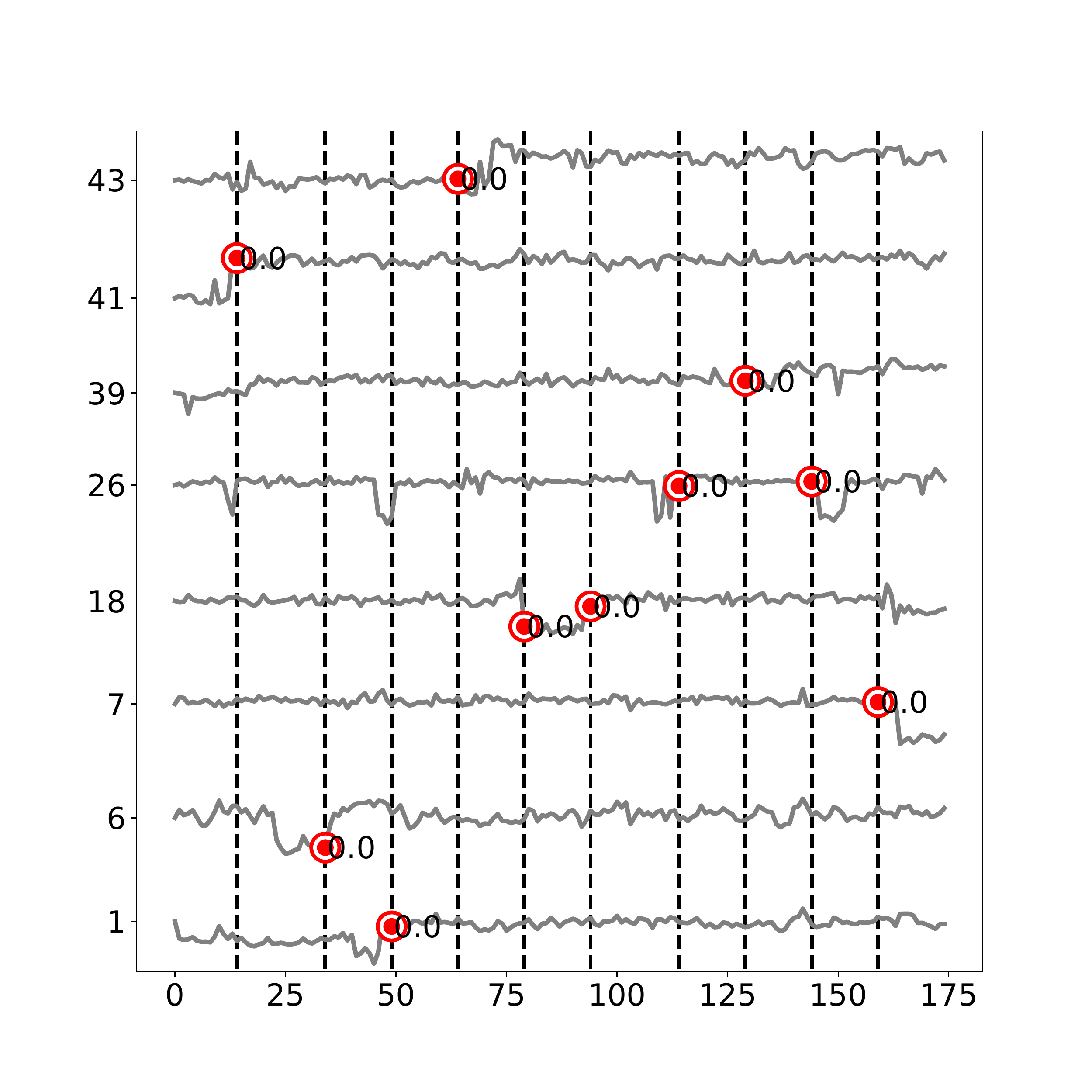}
		\subcaption{AR(1), $W=0$}
	\end{minipage}
	\begin{minipage}{.5\textwidth}
		\centering
		\includegraphics[width=.9\linewidth]{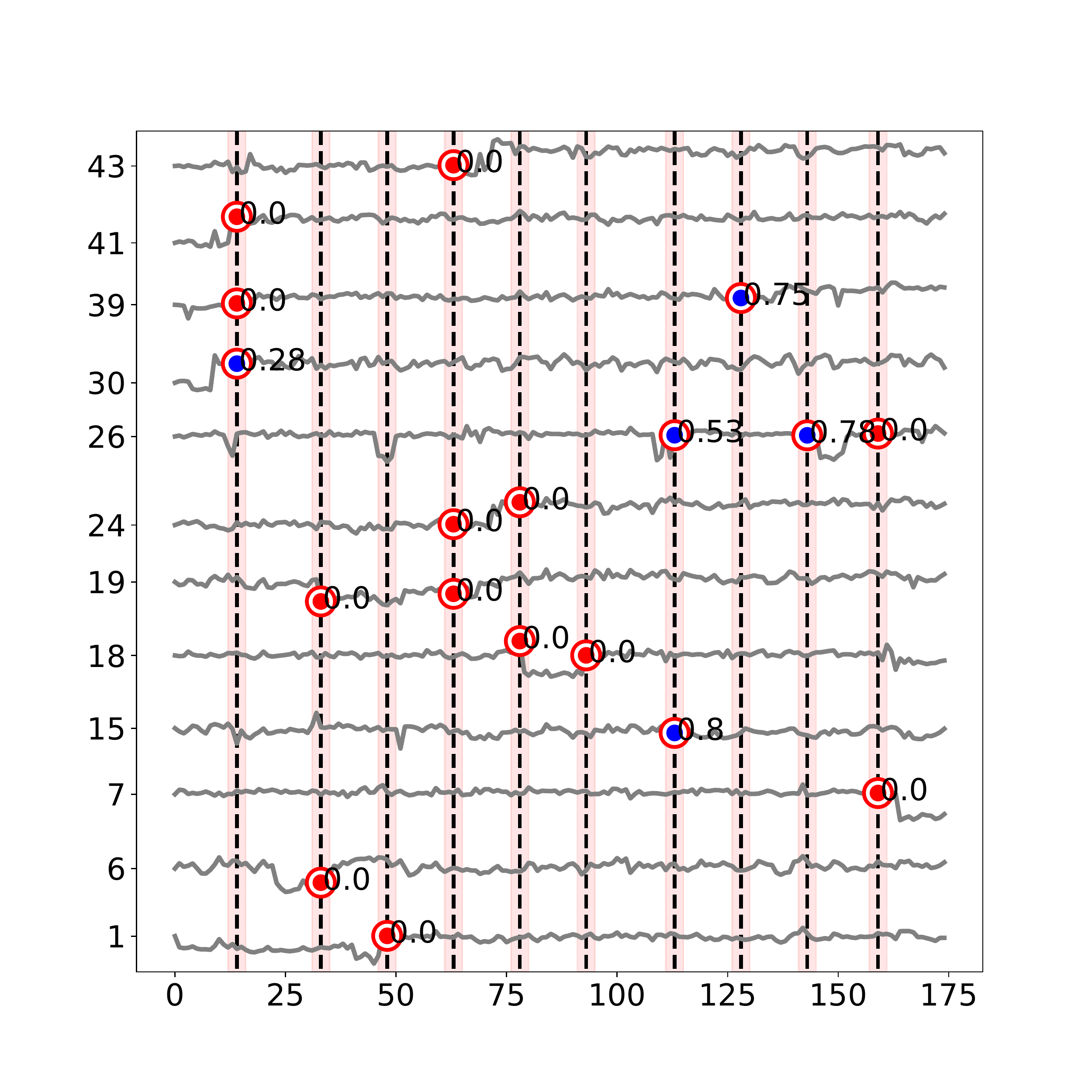}
		\subcaption{Independence, $W=2$}
	\end{minipage}
	\begin{minipage}{.5\linewidth}
		\centering
		\includegraphics[width=.9\linewidth]{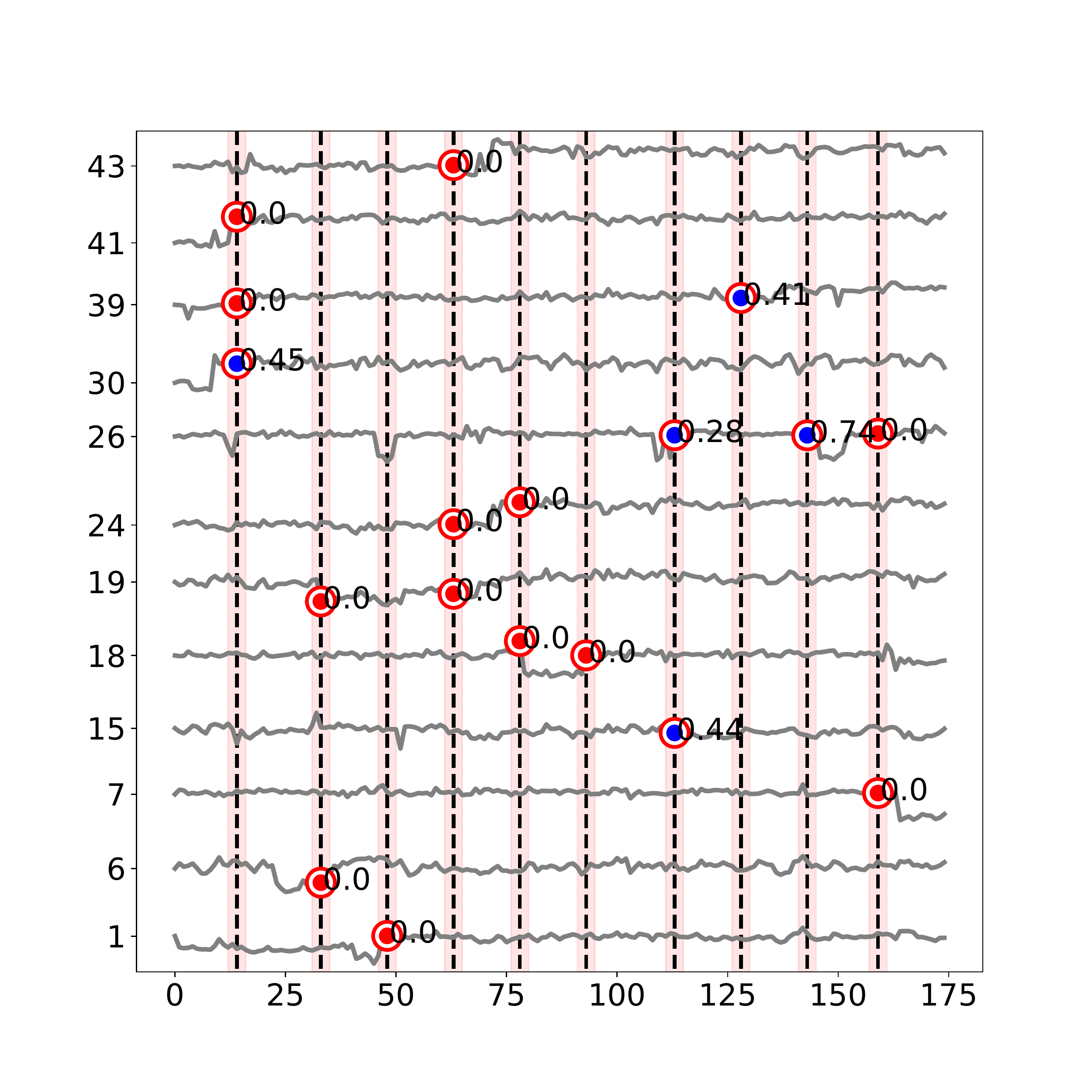}
		\subcaption{AR(1), $W=2$}
	\end{minipage}
	\caption{The results on Array CGH for chromosome 3 with $K=10$ and $L=15$.}
	\label{fig:array_chrom3}
\end{figure}

\begin{figure}[tbp]
	\begin{minipage}{.5\textwidth}
		\centering
		\includegraphics[width=.9\linewidth]{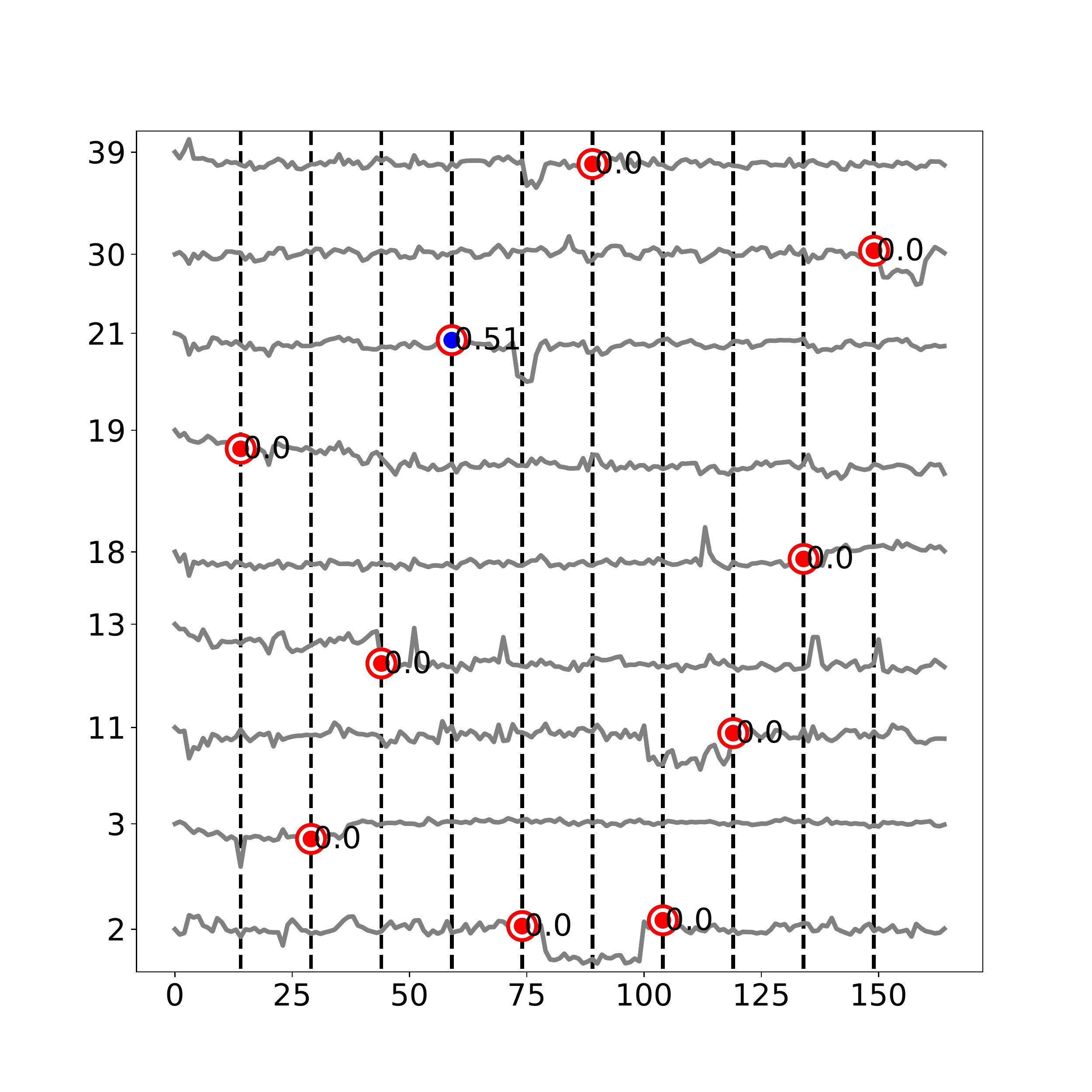}
		\subcaption{Independence, $W=0$}
	\end{minipage}
	\begin{minipage}{.5\linewidth}
		\centering
		\includegraphics[width=.9\linewidth]{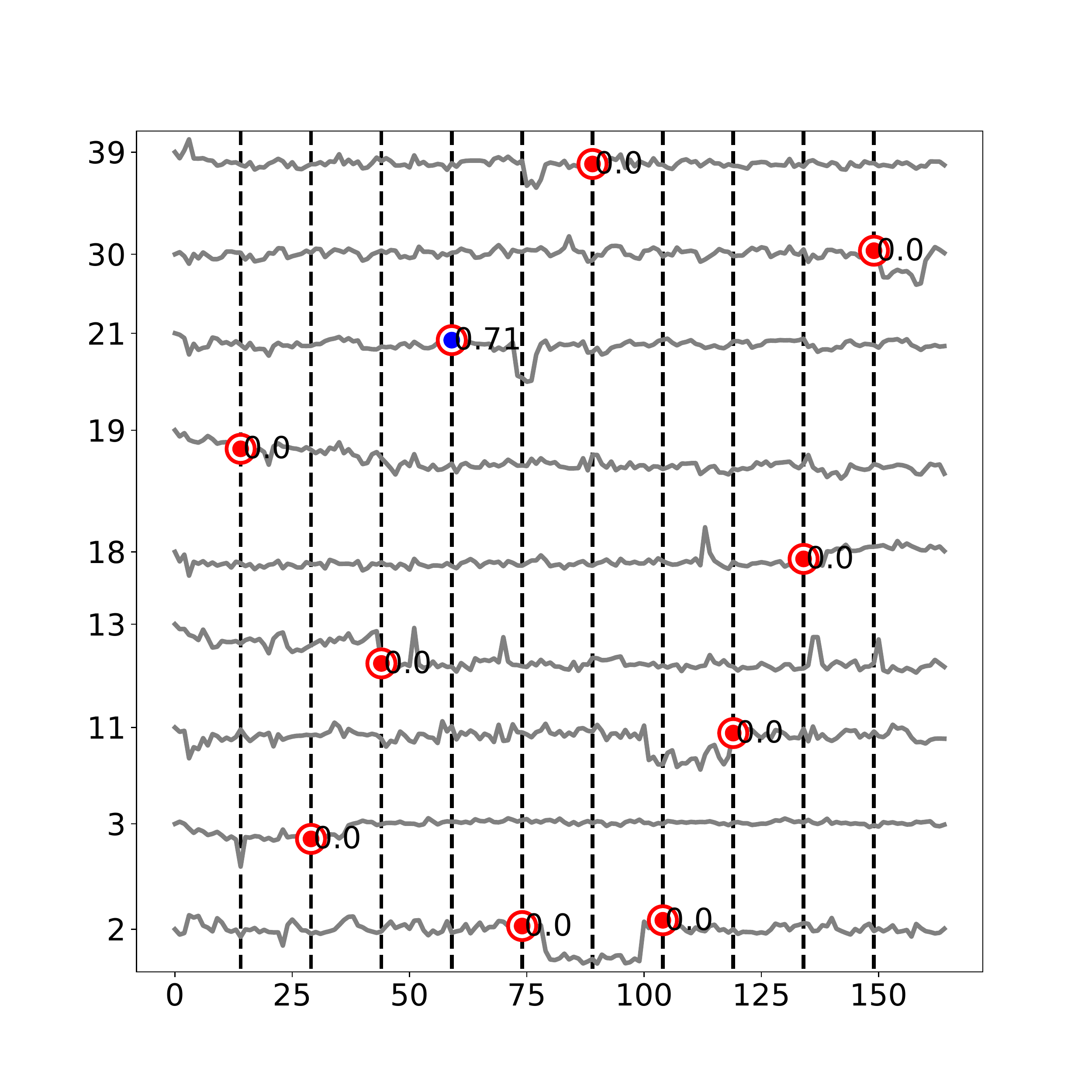}
		\subcaption{AR(1), $W=0$}
	\end{minipage}
	\begin{minipage}{.5\textwidth}
		\centering
		\includegraphics[width=.9\linewidth]{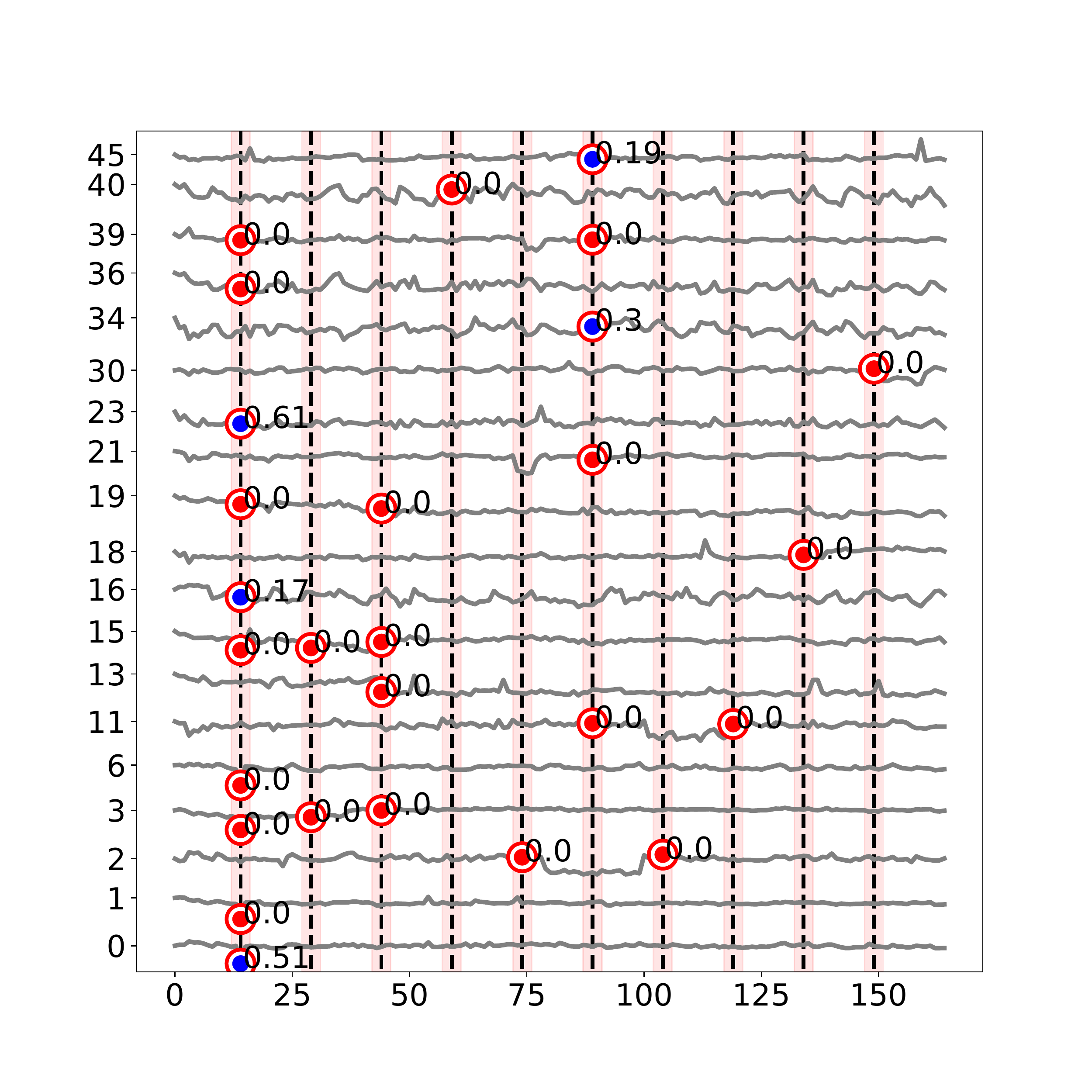}
		\subcaption{Independence, $W=2$}
	\end{minipage}
	\begin{minipage}{.5\linewidth}
		\centering
		\includegraphics[width=.9\linewidth]{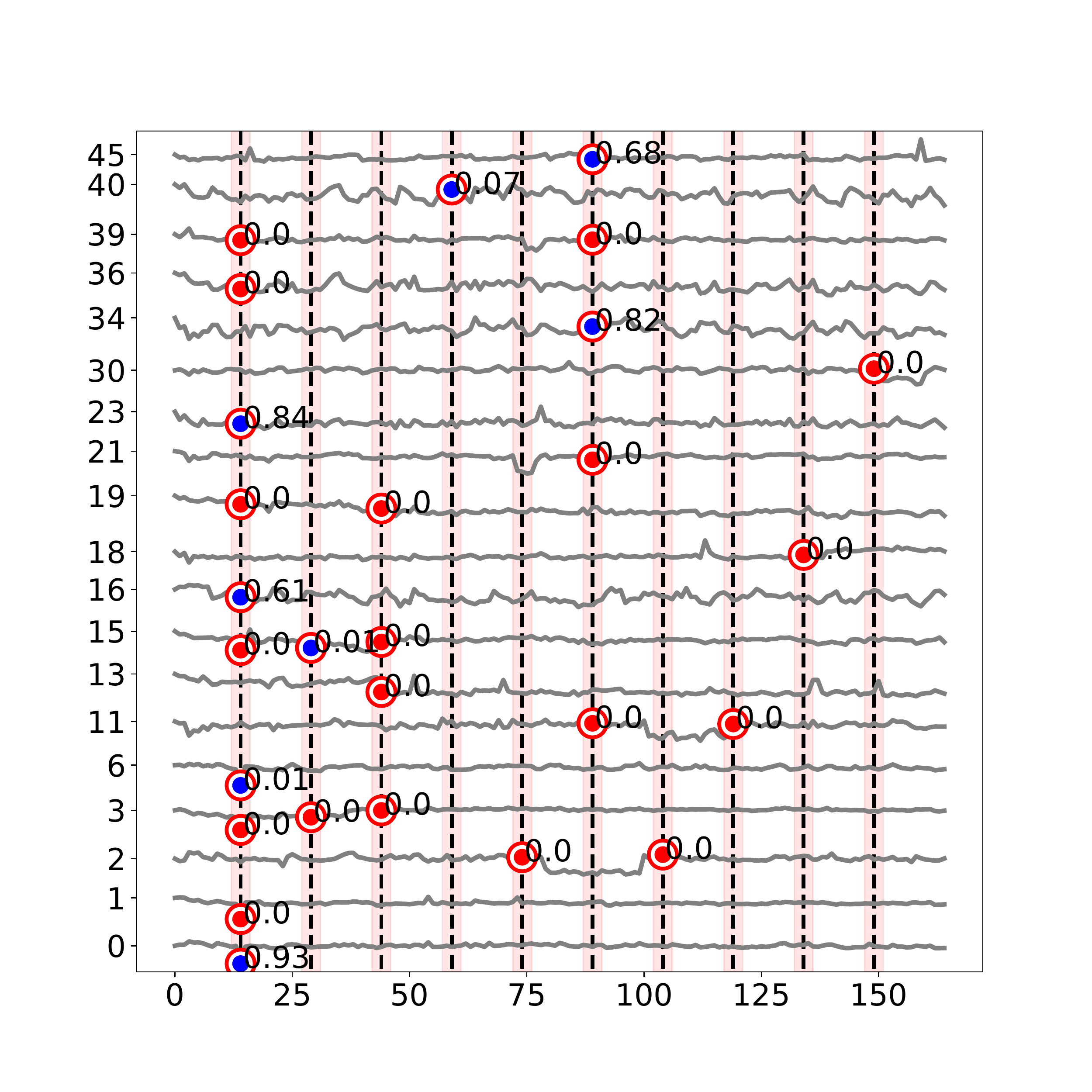}
		\subcaption{AR(1), $W=2$}
	\end{minipage}
	\caption{The results on Array CGH for chromosome 4 with $K=10$ and $L=15$.}
	\label{fig:array_chrom4}
\end{figure}

\begin{figure}[tbp]
	\begin{minipage}{.5\textwidth}
		\centering
		\includegraphics[width=.9\linewidth]{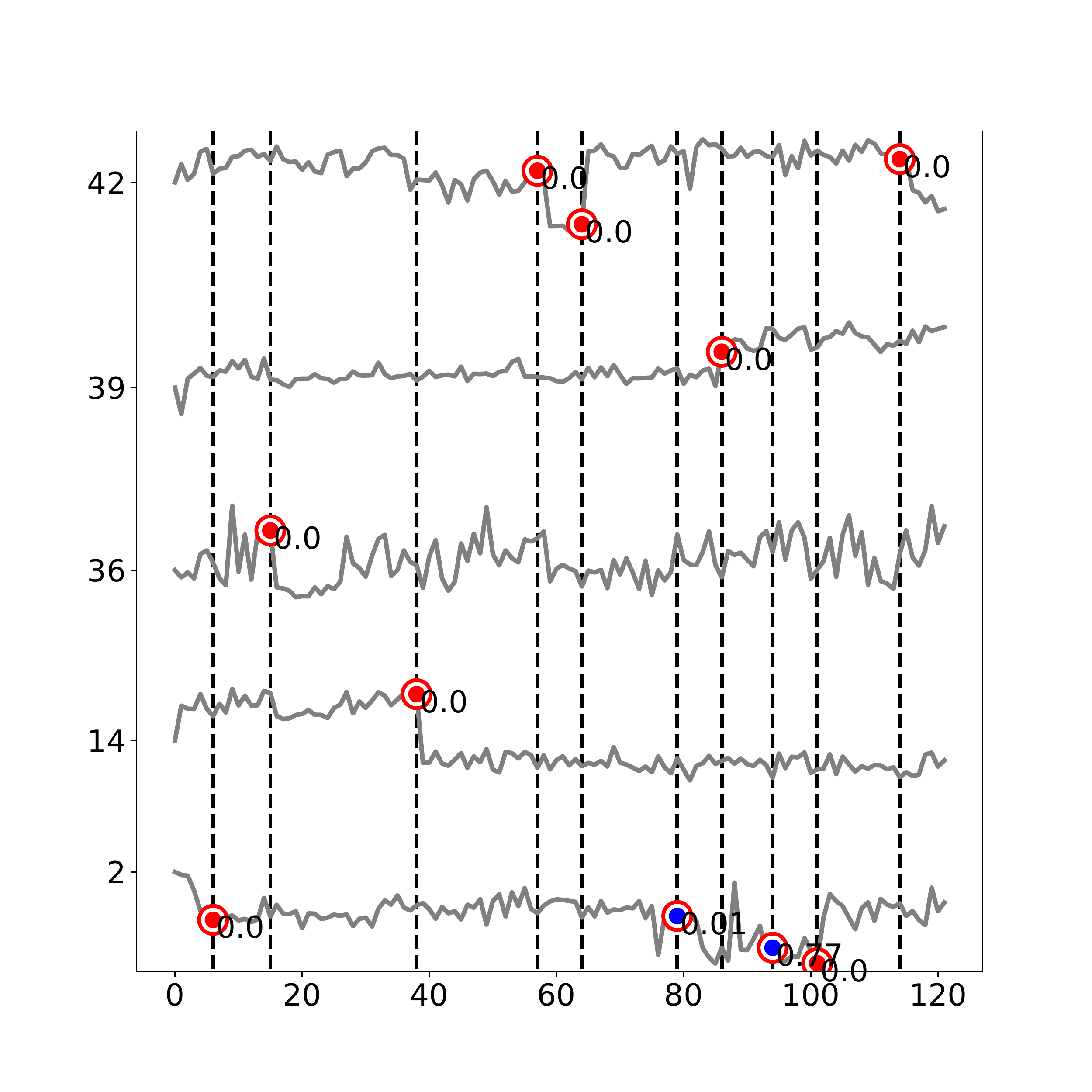}
		\subcaption{Independence, $W=0$}
	\end{minipage}
	\begin{minipage}{.5\linewidth}
		\centering
		\includegraphics[width=.9\linewidth]{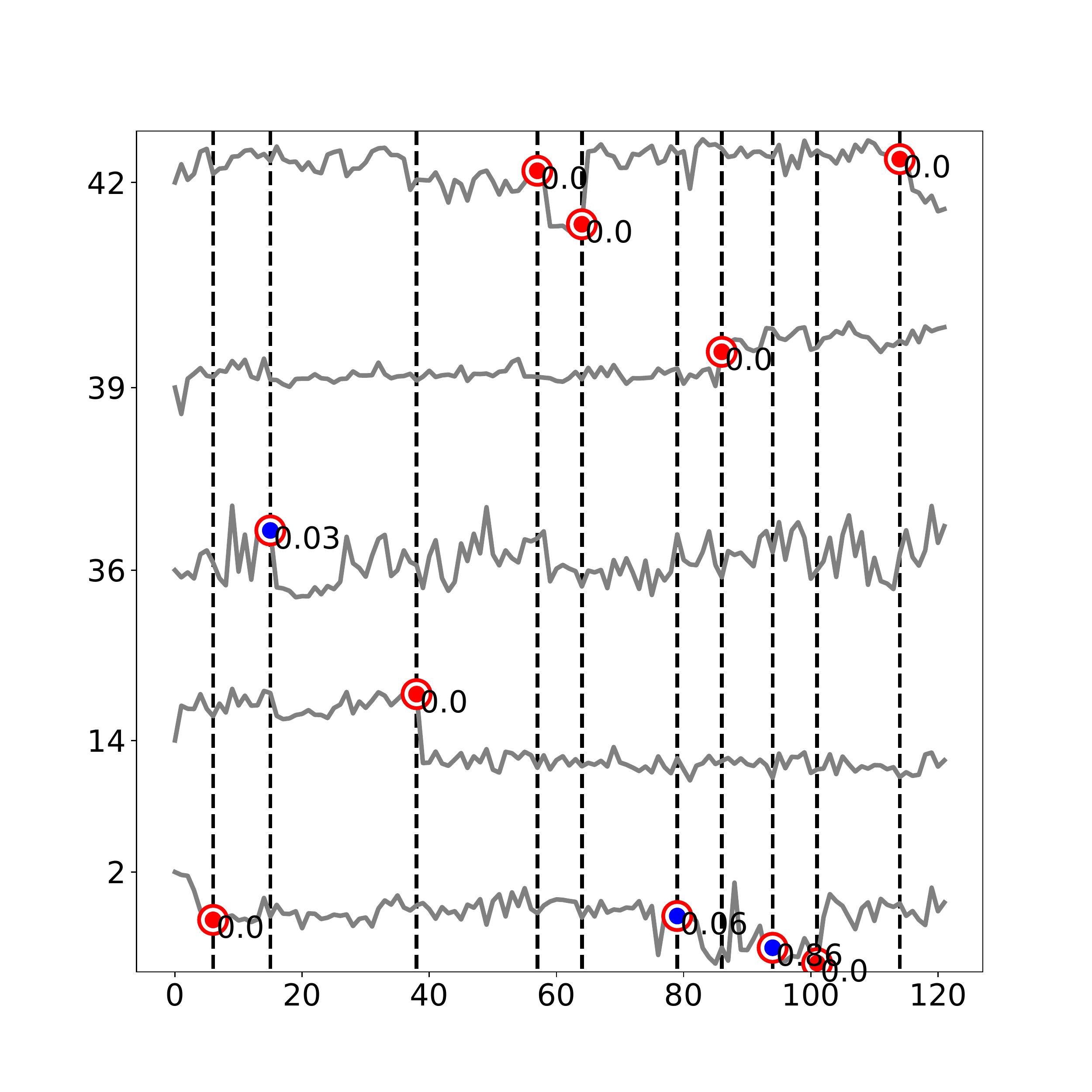}
		\subcaption{AR(1), $W=0$}
	\end{minipage}
	\begin{minipage}{.5\textwidth}
		\centering
		\includegraphics[width=.9\linewidth]{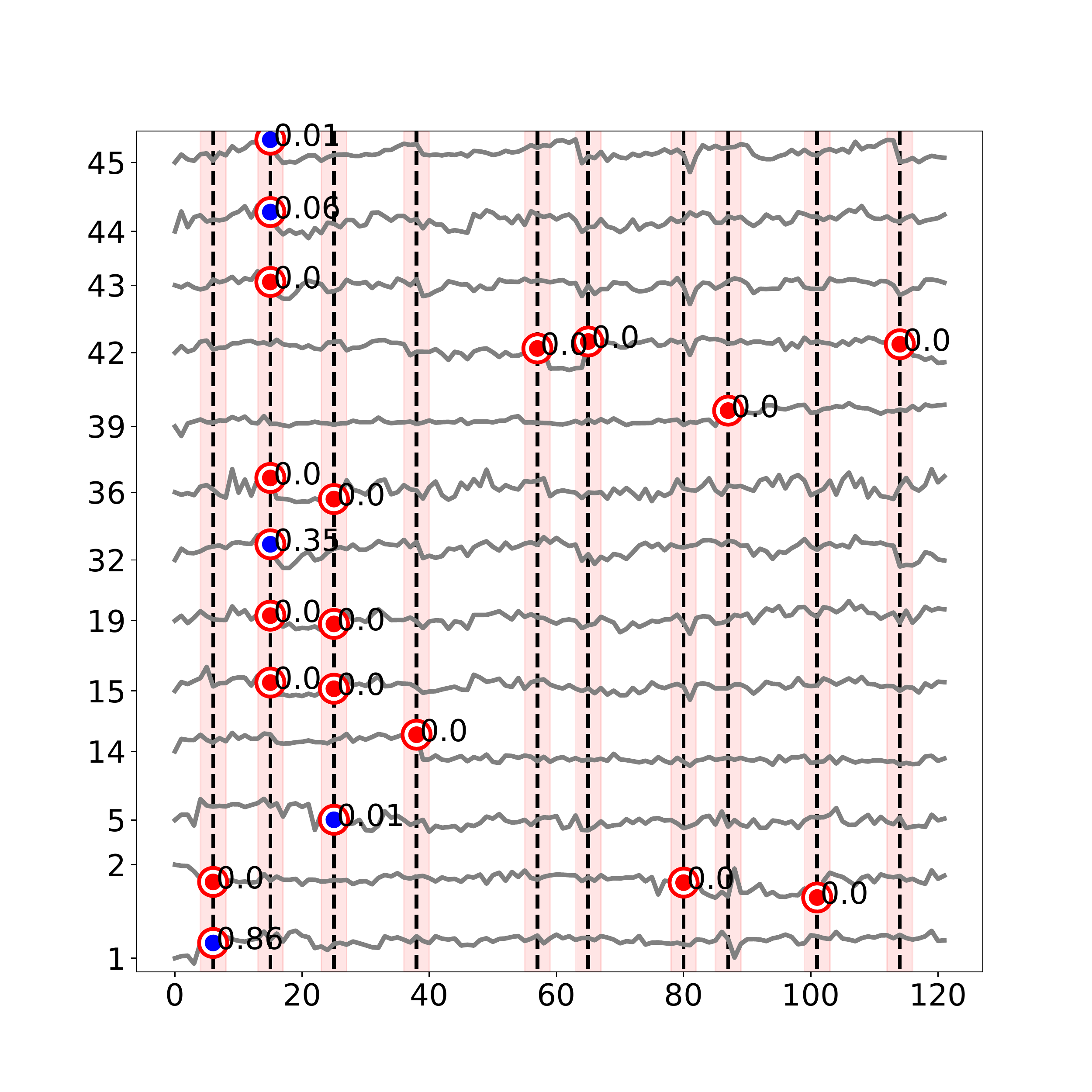}
		\subcaption{Independence, $W=2$}
	\end{minipage}
	\begin{minipage}{.5\linewidth}
		\centering
		\includegraphics[width=.9\linewidth]{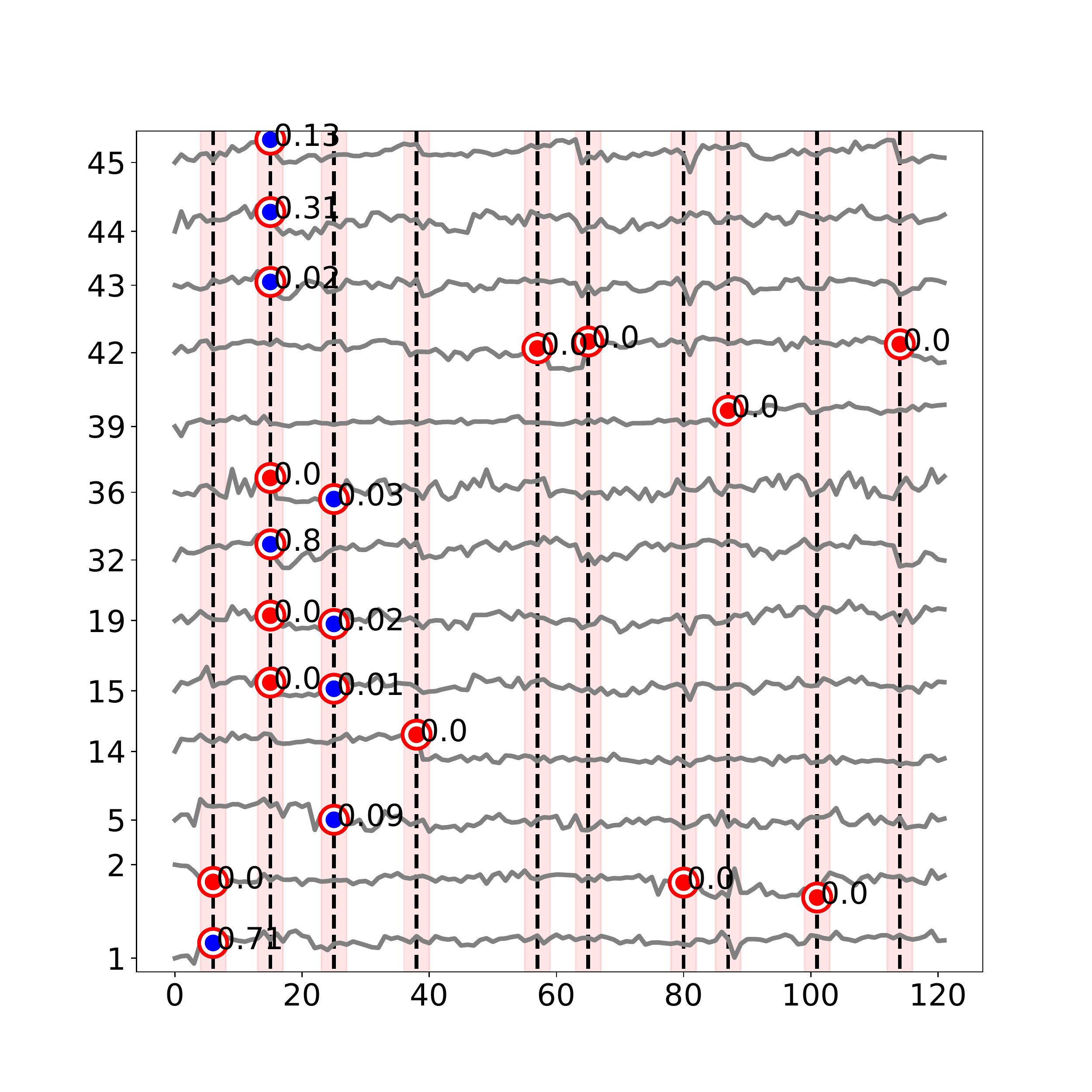}
		\subcaption{AR(1), $W=2$}
	\end{minipage}
	\caption{The results on Array CGH for chromosome 5 with $K=10$ and $L=7$.}
	\label{fig:array_chrom5}
\end{figure}

\paragraph{Array CGH}
We applied the naive method and the proposed-MC method to each of the 22 chromosomes, where $N$ and $D$ correspond to the length of each chromosome and the number of subjects, respectively. 
Here, in addition to the result of chromosome 1 in the main paper, we show the results of chromosome 2 -- 5 to avoid too large a file size.
For the covariance matrix $\Sigma$, we considered two alternatives based on the Array CGH data set of 18 healthy subjects, where it is reasonable to assume that CP is not present.
The first option is to assume the independence structure and the second option is to assume AR(1) structure. 
The parameters $\sigma^2$ (for both of independence and AR(1) options), $\gamma$ and $\rho$ (for AR(1) option) were estimated as 
\begin{align*}
  \hat{\sigma}^2 &= \frac{1}{18N}\sum_{i=1}^{18}\sum_{j=1}^N(X_{i, j} - \mu_i)^2 \\
  \hat{\gamma} &= \frac{1}{18(N-1)}\sum_{i=1}^{18}\sum_{j=2}^N(X_{i, j} - \mu_i)(X_{i, j-1} - \mu_i) \\
  \hat{\rho} &= \frac{\hat{\gamma}}{\hat{\sigma}^2}, 
\end{align*}
where $\mu_i = \frac{1}{N}\sum_{j=1}^NX_{i, j}$.
Here, we assumed that correlations between components are uncorrelated because each component represents individual subjects. 
Figures~\ref{fig:array_chrom2}--\ref{fig:array_chrom5} show results of chromosomes 2--5.
As we discussed in \S\ref{sec:sec4}, the naive test declared that most of the detected CP locations and components are statistically significant, whereas the proposed-MC method suggested that some of them are not really statistically significant.


\begin{figure}[tbp]
	\begin{minipage}{.5\linewidth}
		\centering
		\includegraphics[width=.85\textwidth]{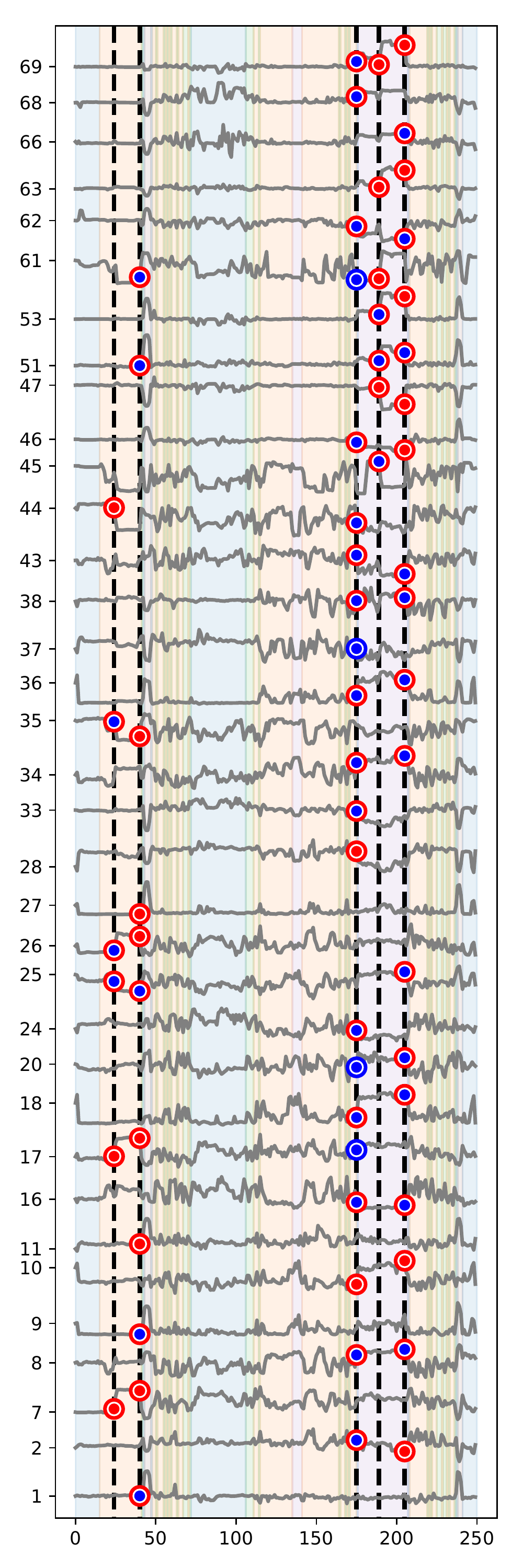}
		\subcaption{Independence}
	\end{minipage}
	\begin{minipage}{.5\linewidth}
		\centering
		\includegraphics[width=.85\textwidth]{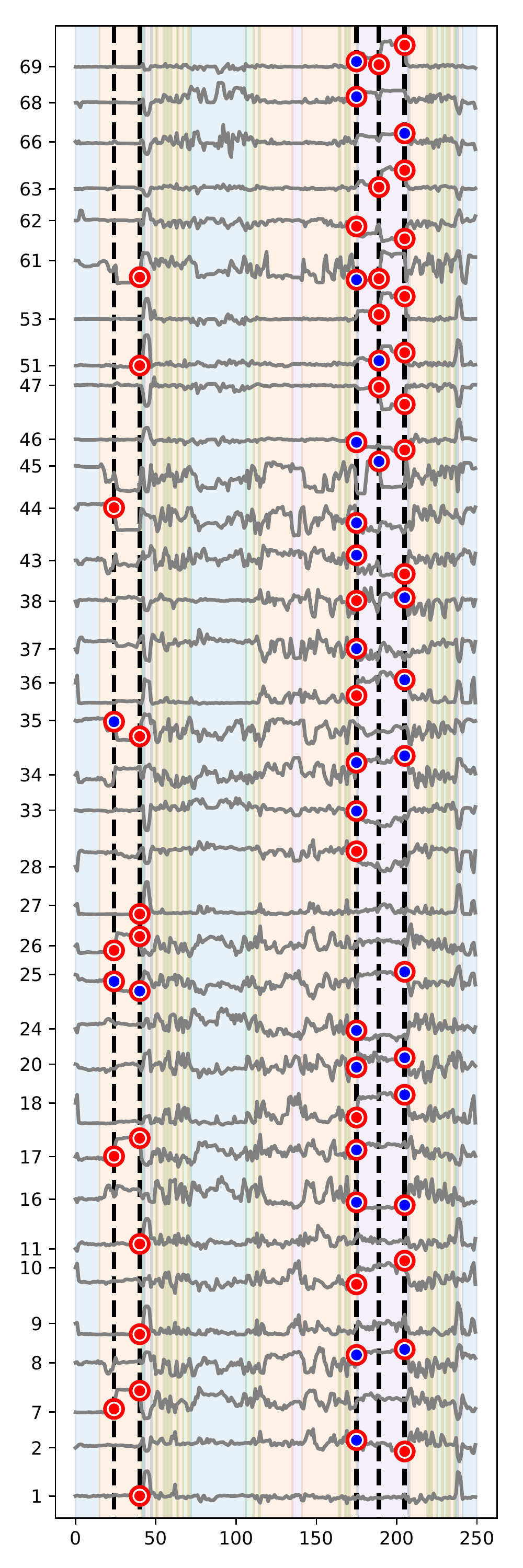}
		\subcaption{AR(1)}
	\end{minipage}
	\caption{The results on Human Activity Recognition for the setting ADL1 of subject S1 with $K=5, L=10, W=0$.}
	\label{fig:HAR_all1}
\end{figure}

\begin{figure}[tbp]
	\begin{minipage}{.5\linewidth}
		\centering
		\includegraphics[width=.85\textwidth]{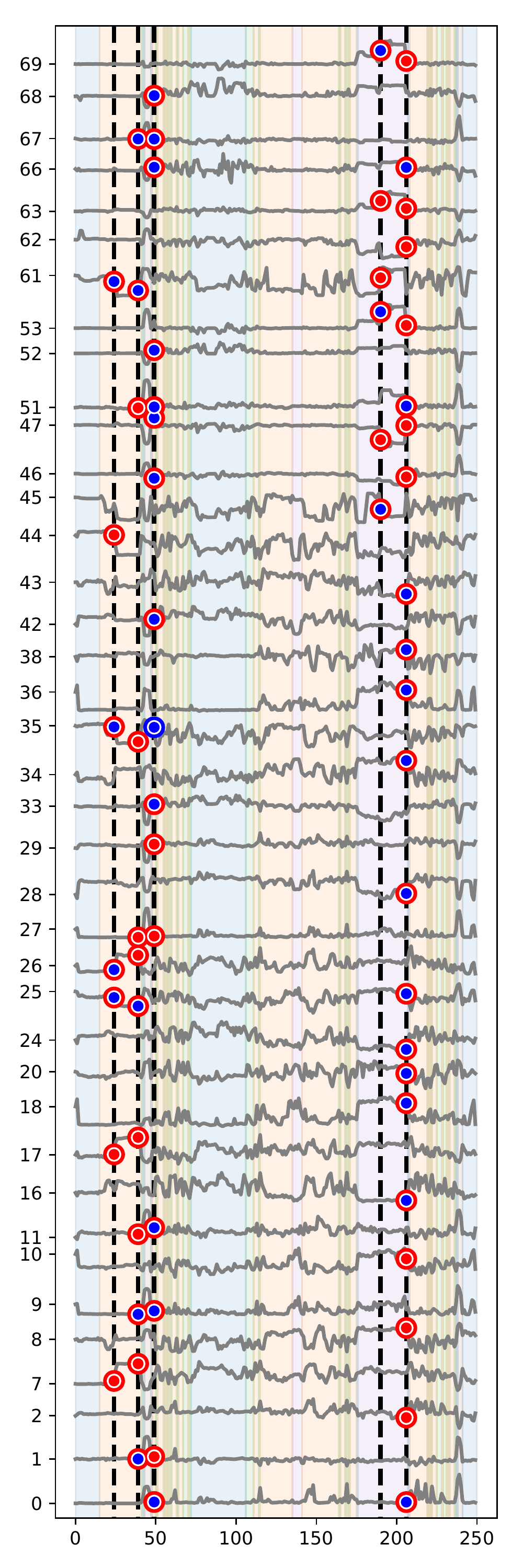}
		\subcaption{Independence}
	\end{minipage}
	\begin{minipage}{.5\linewidth}
		\centering
		\includegraphics[width=.85\textwidth]{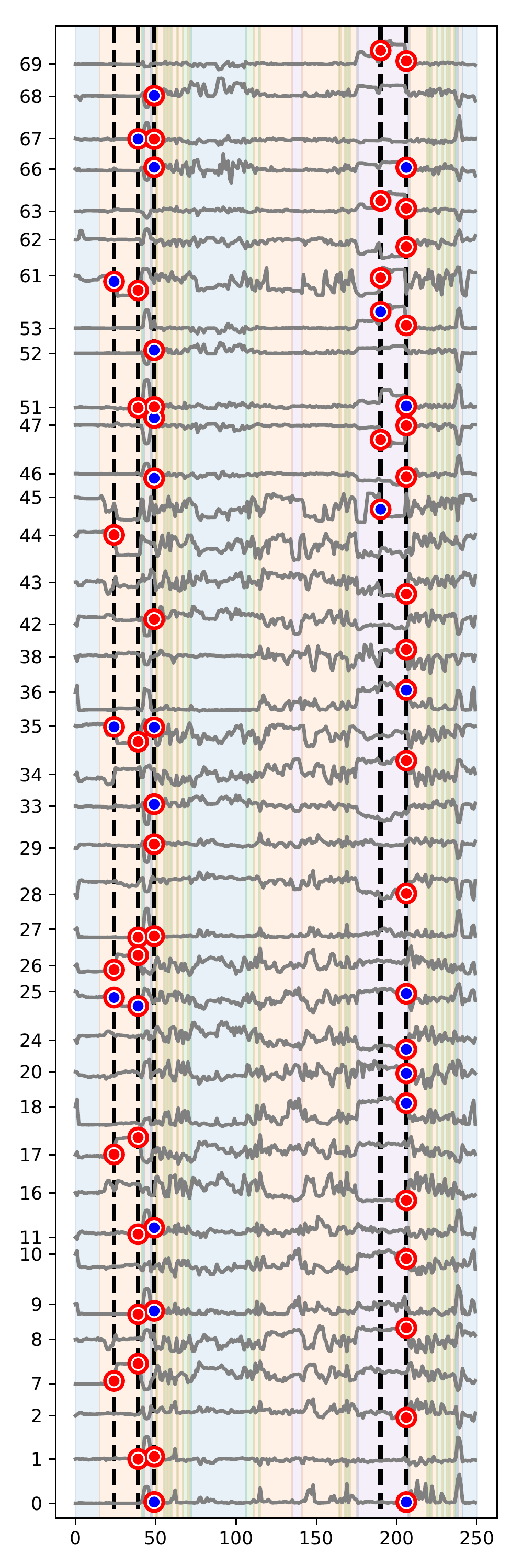}
		\subcaption{AR(1)}
	\end{minipage}
	\caption{The results on Human Activity Recognition for the setting ADL1 of subject S1 with $K=5, L=10, W=2$.}
	\label{fig:HAR_all2}
\end{figure}

\begin{figure}[tbp]
	\begin{minipage}{.5\linewidth}
		\centering
		\includegraphics[width=.85\textwidth]{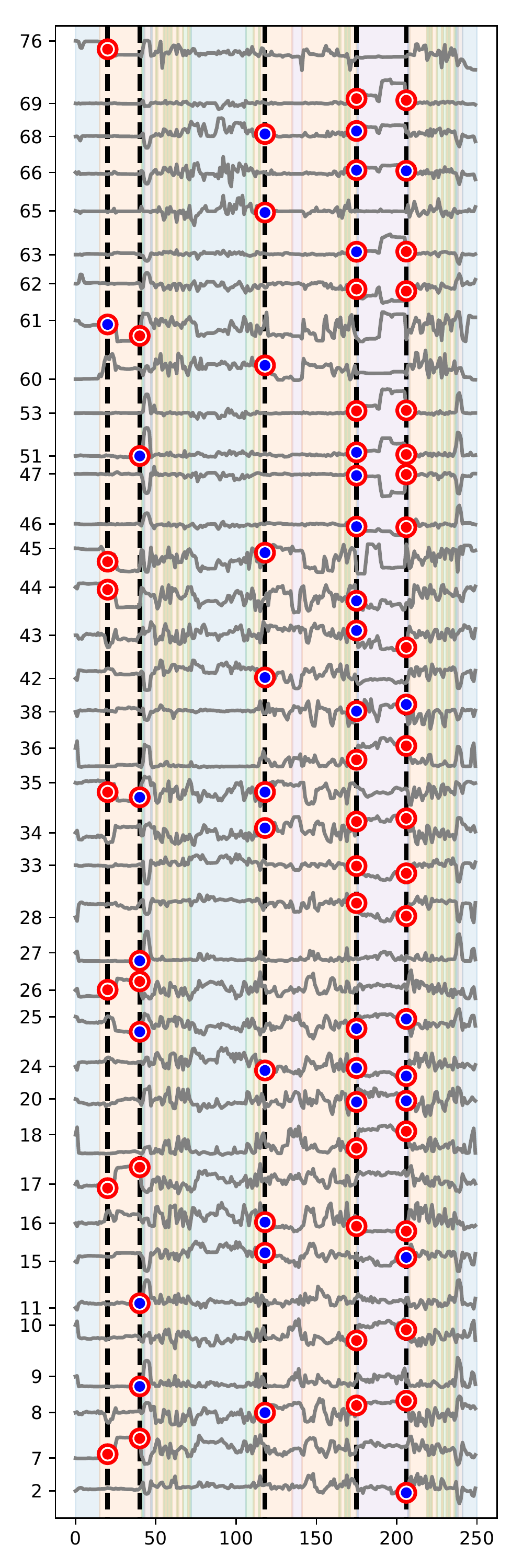}
		\subcaption{Independence}
	\end{minipage}
	\begin{minipage}{.5\linewidth}
		\centering
		\includegraphics[width=.85\textwidth]{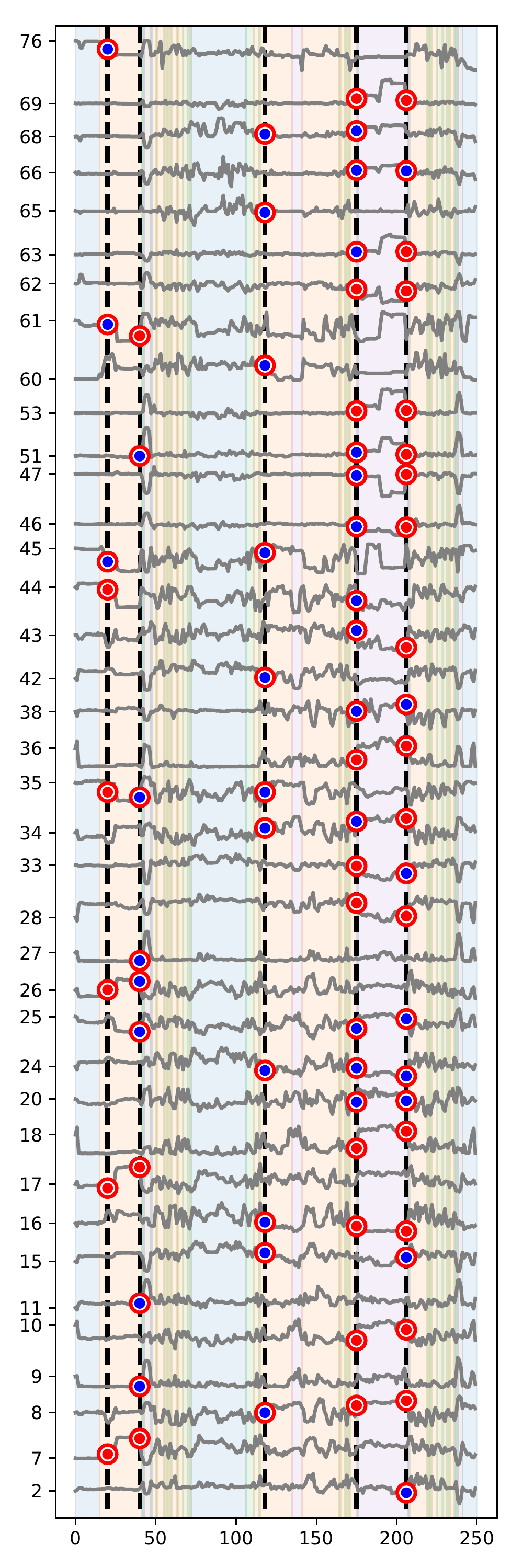}
		\subcaption{AR(1)}
	\end{minipage}
	\caption{The results on Human Activity Recognition for the setting ADL1 of subject S1 with $K=5, L=20, W=0$.}
	\label{fig:HAR_all3}
\end{figure}

\begin{figure}[tbp]
	\begin{minipage}{.5\linewidth}
		\centering
		\includegraphics[width=.85\textwidth]{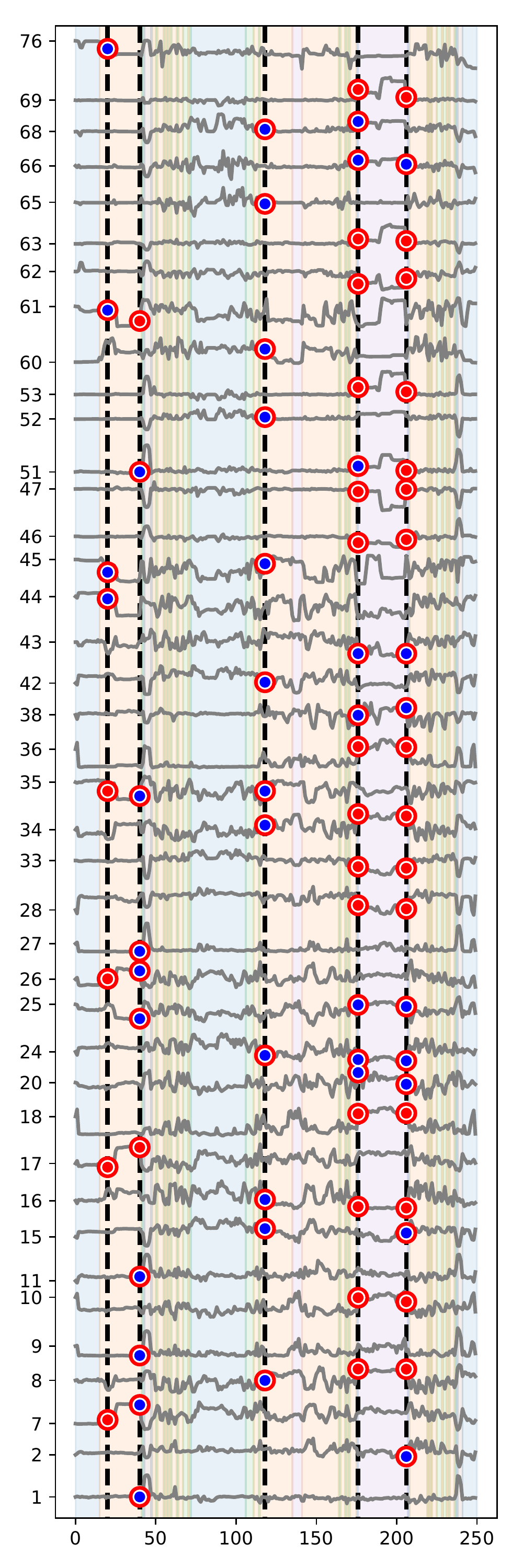}
		\subcaption{Independence}
	\end{minipage}
	\begin{minipage}{.5\linewidth}
		\centering
		\includegraphics[width=.85\textwidth]{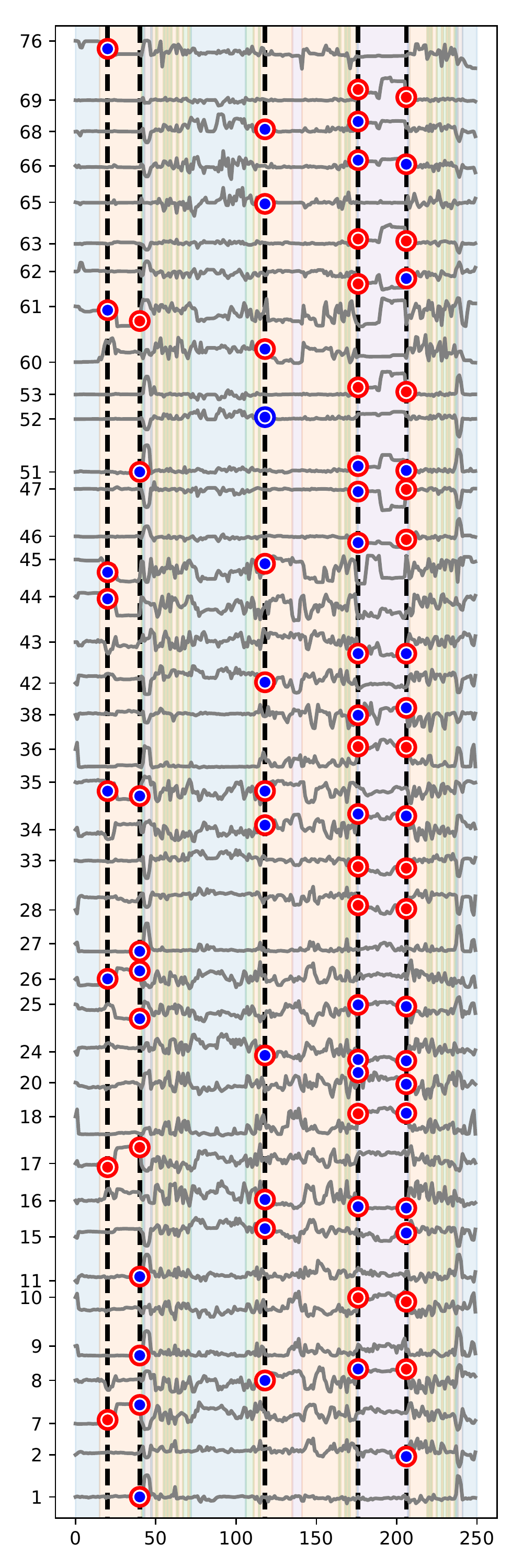}
		\subcaption{AR(1)}
	\end{minipage}
	\caption{The results on Human Activity Recognition for the setting ADL1 of subject S1 with $K=5, L=20, W=2$.}
	\label{fig:HAR_all4}
\end{figure}

\begin{figure}[tbp]
	\begin{minipage}{.5\linewidth}
		\centering
		\includegraphics[width=.85\textwidth]{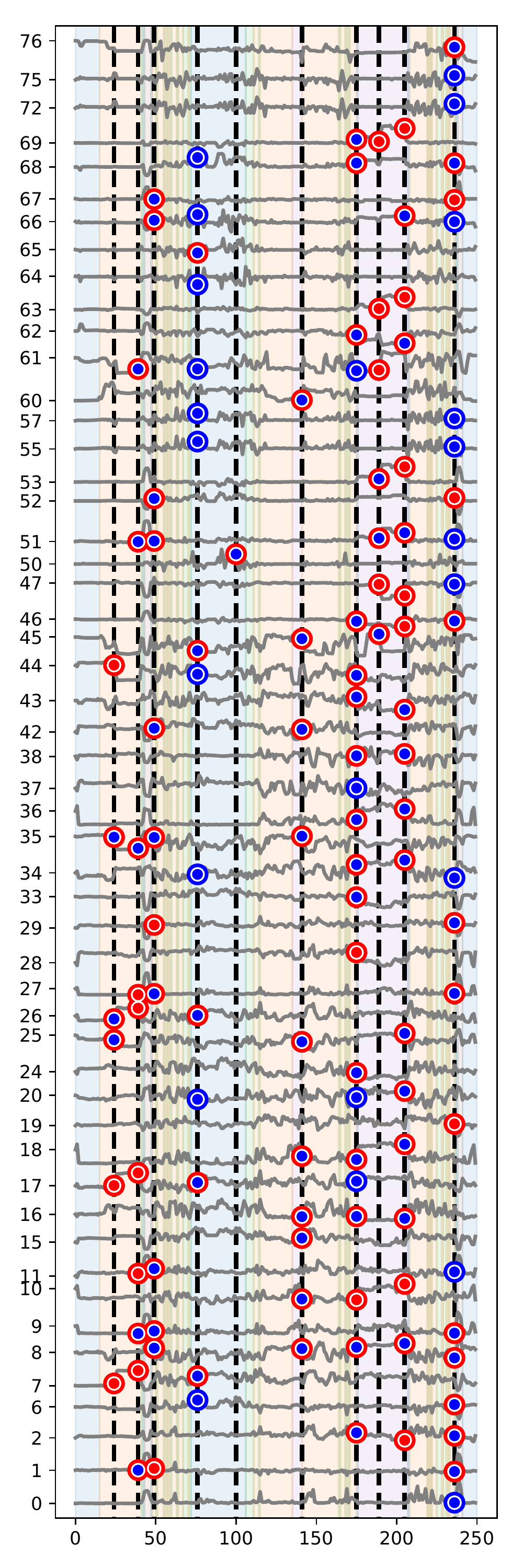}
		\subcaption{Independence}
	\end{minipage}
	\begin{minipage}{.5\linewidth}
		\centering
		\includegraphics[width=.85\textwidth]{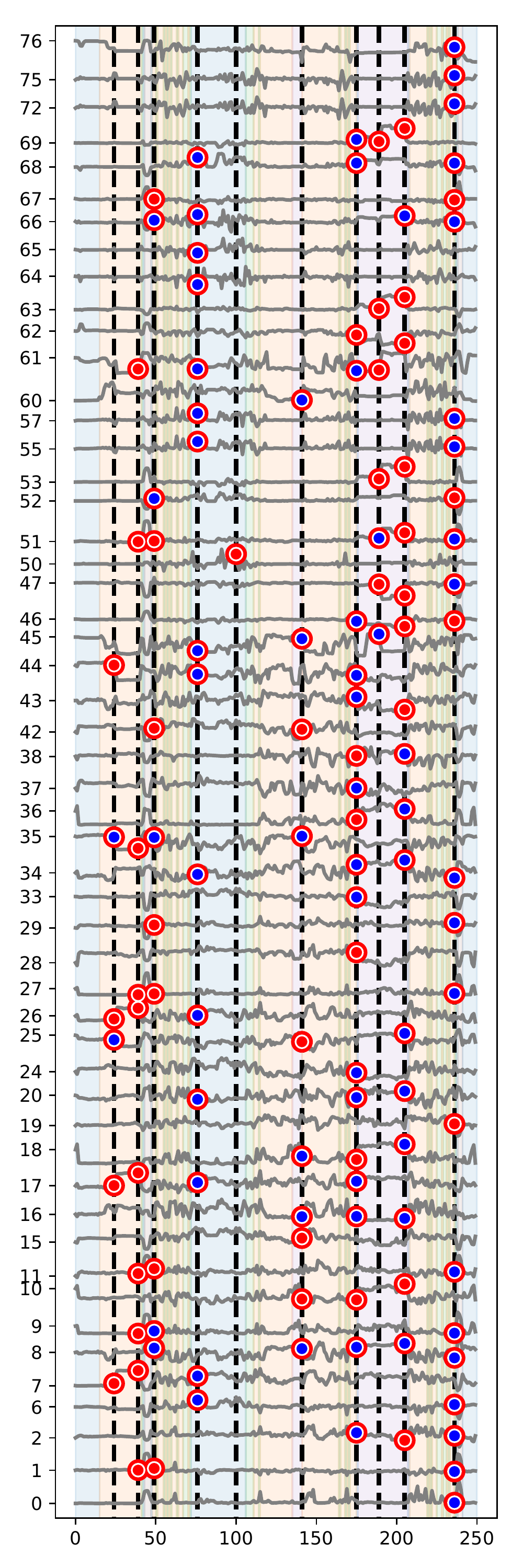}
		\subcaption{AR(1)}
	\end{minipage}
	\caption{The results on Human Activity Recognition for the setting ADL1 of subject S1 with $K=10, L=10, W=0$.}
	\label{fig:HAR_all5}
\end{figure}

\begin{figure}[tbp]
	\begin{minipage}{.5\linewidth}
		\centering
		\includegraphics[width=.85\textwidth]{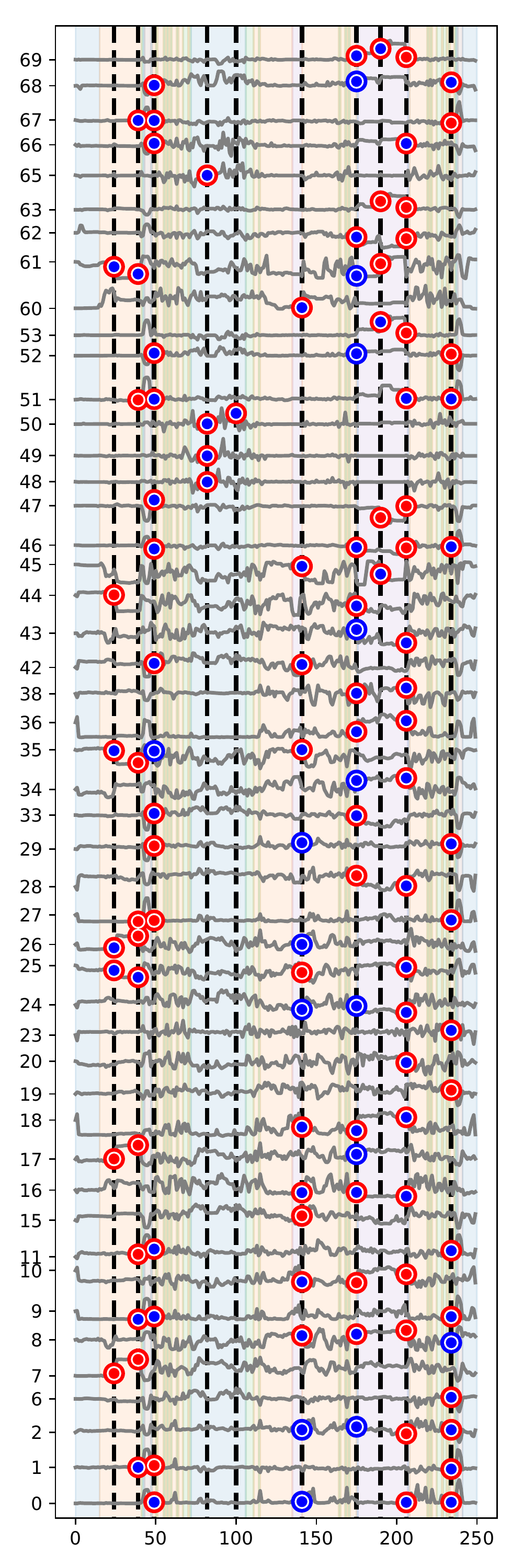}
		\subcaption{Independence}
	\end{minipage}
	\begin{minipage}{.5\linewidth}
		\centering
		\includegraphics[width=.85\textwidth]{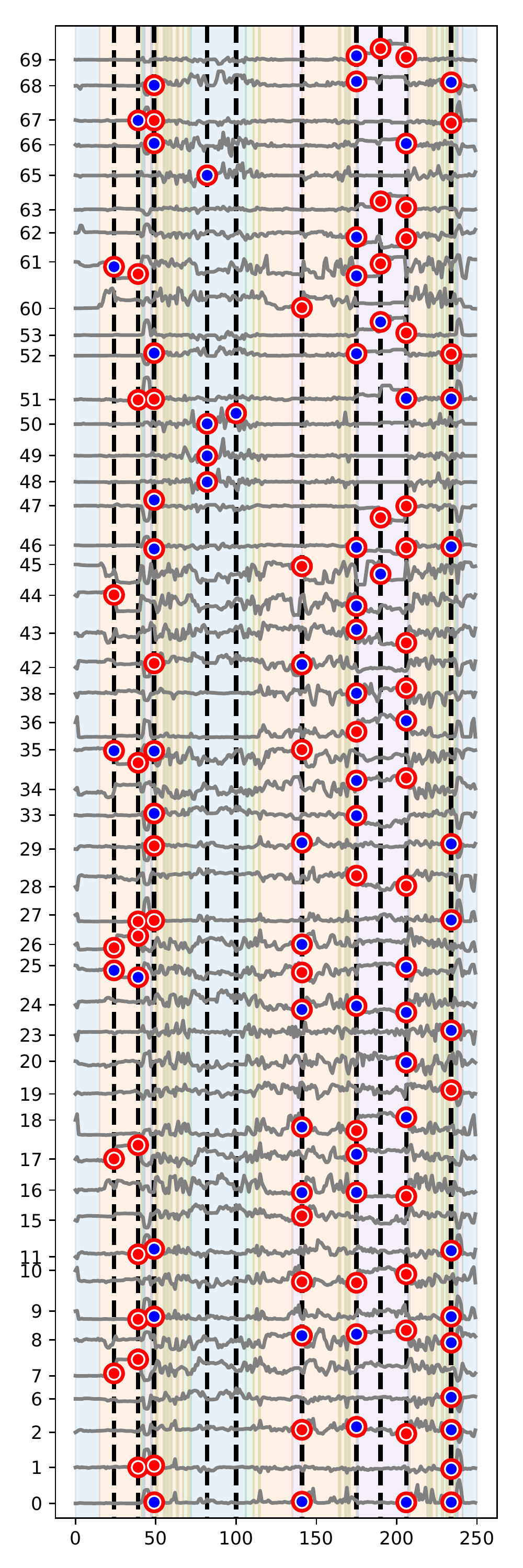}
		\subcaption{AR(1)}
	\end{minipage}
	\caption{The results on Human Activity Recognition for the setting ADL1 of subject S1 with $K=10, L=10, W=2$.}
	\label{fig:HAR_all6}
\end{figure}

\begin{figure}[tbp]
	\begin{minipage}{.5\linewidth}
		\centering
		\includegraphics[width=.85\textwidth]{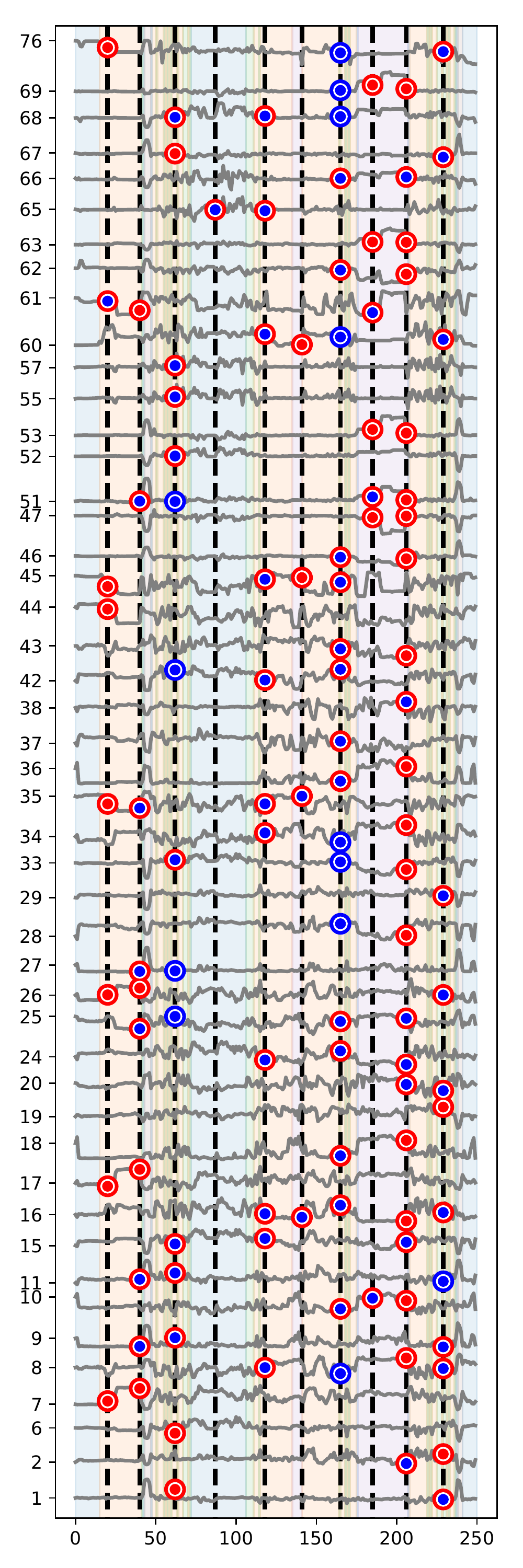}
		\subcaption{Independence}
	\end{minipage}
	\begin{minipage}{.5\linewidth}
		\centering
		\includegraphics[width=.85\textwidth]{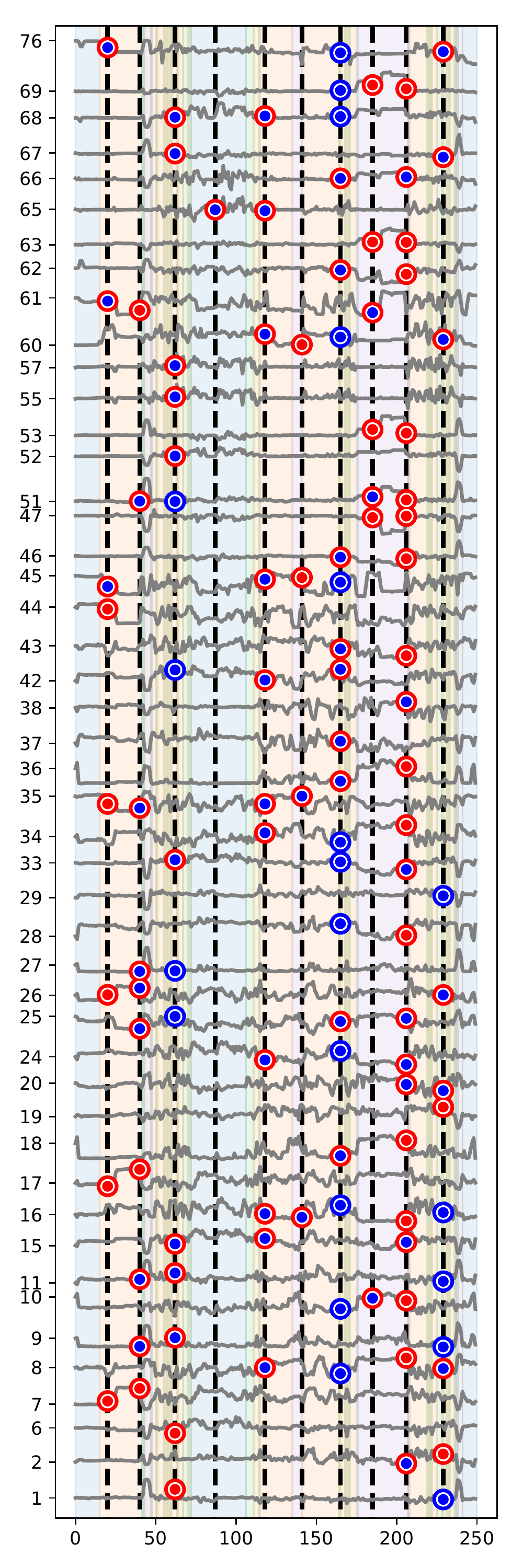}
		\subcaption{AR(1)}
	\end{minipage}
	\caption{The results on Human Activity Recognition for the setting ADL1 of subject S1 with $K=10, L=20, W=0$.}
	\label{fig:HAR_all7}
\end{figure}

\begin{figure}[tbp]
	\begin{minipage}{.5\linewidth}
		\centering
		\includegraphics[width=.85\textwidth]{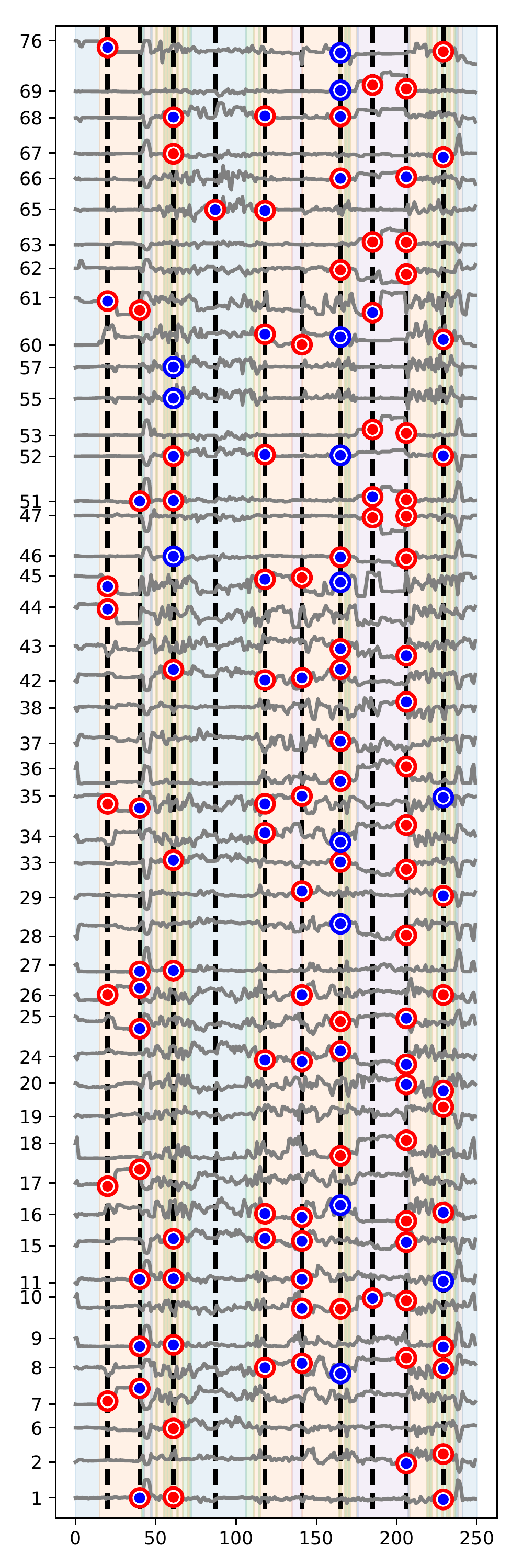}
		\subcaption{Independence}
	\end{minipage}
	\begin{minipage}{.5\linewidth}
		\centering
		\includegraphics[width=.85\textwidth]{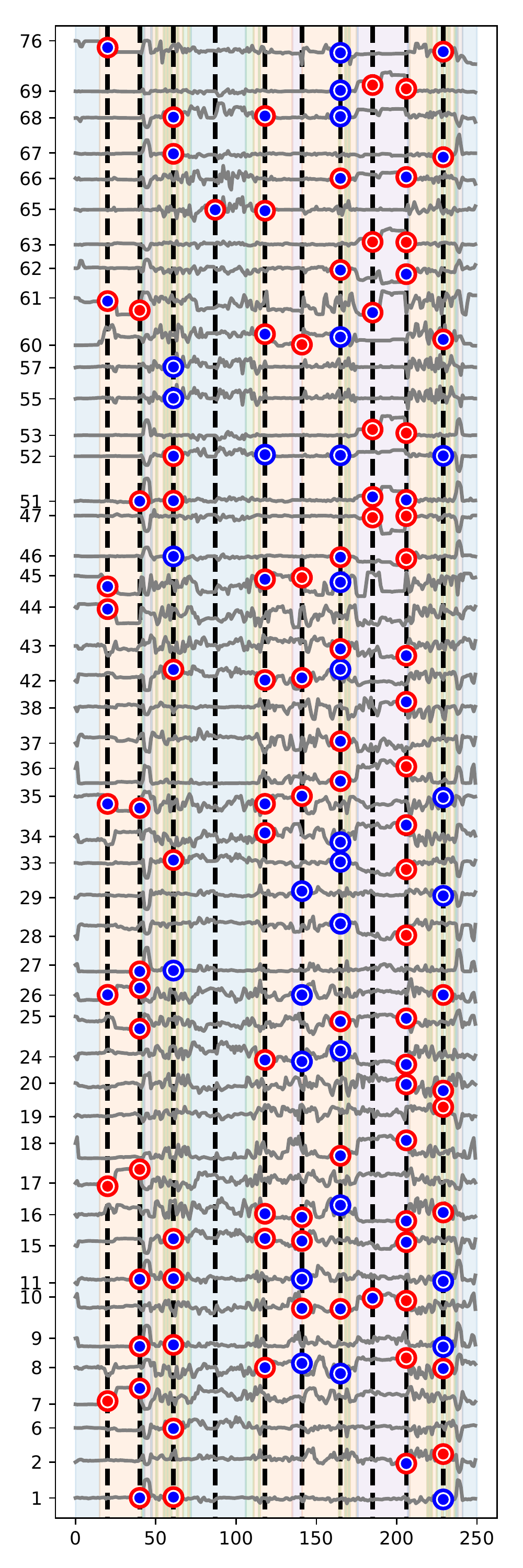}
		\subcaption{AR(1)}
	\end{minipage}
	\caption{The results on Human Activity Recognition for the setting ADL1 of subject S1 with $K=10, L=20, W=2$.}
	\label{fig:HAR_all8}
\end{figure}

\paragraph{Human Activity Recognition}
The original dataset in
\citet{chavarriaga2013opportunity}
consists of
multi-dimensional sensor signals
taken from four subjects (S1-S4) in five settings (ADL1-ADL5).
In this paper, 
we applied the naive method and the proposed-MC method
only to the first setting (ADL1) of first subject (S1). 
We considered the same covariance structure options as in the Array CGH data.
The covariance structures were estimated by using the signals in the longest continuous locomotion state ``Lie'' because it indicates the sleeping of the subjects, i.e., no changes exist, as follows:
\begin{align*}
\hat{\sigma}^2 &= \frac{1}{77n}\sum_{i=1}^{77}\sum_{j=1}^n(X_{i, j} - \mu_i)^2, \\
\hat{\gamma} &= \frac{1}{77(n-1)}\sum_{i=1}^{77}\sum_{j=2}^n(X_{i, j} - \mu_i)(X_{i, j-1} - \mu_i), \\
\hat{\rho} &= \frac{\hat{\gamma}}{\hat{\sigma}^2}.
\end{align*}
Here, we used $8$ different combinations of hyperparameters
$(K, L, W)  \in \{5, 10\} \times \{10, 20\} \times \{0, 2\}$.
Figures
\ref{fig:HAR_all1}--\ref{fig:HAR_all8}
show the results of all the parameter settings. 
As discussed in
\S\ref{sec:sec4},
the naive test declared that most detected CP locations and components are statistically significant, 
but the proposed-MC method suggested that some of them are not really statistically significant.
%

\newpage

\bibliographystyle{plainnat}
\bibliography{draft}

\end{document}